\documentclass[final, 3p, 12pt]{elsarticle}
\usepackage[utf8]{inputenc}             % allow utf-8 input
\usepackage[T1]{fontenc}                % use 8-bit T1 fonts
\usepackage{hyperref}                   % hyperlinks
\usepackage{url}                        % simple URL typesetting
\usepackage{booktabs}                   % professional-quality tables
\usepackage{amsfonts}                   % blackboard math symbols
\usepackage{nicefrac}                   % compact symbols for 1/2, etc.
\usepackage{amsmath}
\usepackage{microtype}                  % microtypography
\usepackage{amssymb}                    % symbols
\usepackage{graphicx,dblfloatfix}       % graphics
\usepackage{lipsum}
\usepackage{comment}
\usepackage{tabu}
\usepackage{float}
\usepackage{xcolor}
\usepackage[linesnumbered,ruled]{algorithm2e}
\usepackage{footnote}
\usepackage[]{caption}
\usepackage{subcaption}
\usepackage{makecell}
\usepackage{float}
\usepackage{adjustbox}
\usepackage{fullpage}
\usepackage{setspace}
\usepackage{soul}

\DeclareMathOperator*{\argmax}{argmax} % no space, limits underneath in displays

\DeclareMathOperator*{\E}{\mathbb{E}}

\journal{Computer Networks}

\begin{document}
\begin{frontmatter}

\title{Concurrent Decentralized Channel Allocation and Access Point Selection using Multi-Armed Bandits in multi BSS WLANs}

\author{\'Alvaro L\'opez-Ravent\'os}\corref{cor1}
\author{Boris Bellalta}
\cortext[cor1]{Corresponding author. E-mail address: \url{alvaro.lopez@upf.edu}}
\address{Wireless Networking (WN) Research Group \\
                Universitat Pompeu Fabra (UPF), Barcelona, Spain}

\begin{abstract}
Enterprise Wireless Local Area Networks (WLANs) consist of multiple Access Points (APs) covering a given area. In these networks, interference is mitigated by allocating different channels to neighboring APs. Besides, stations are allowed to associate to any AP in the network, selecting by default the one from which receive higher power, even if it is not the best option in terms of the network performance. 

Finding a suitable network configuration able to maximize the performance of enterprise WLANs is a challenging task given the complex dependencies between APs and stations. Recently, in wireless networking, the use of reinforcement learning techniques has emerged as an effective solution to efficiently explore the impact of different network configurations in the system performance, identifying those that provide better performance. 

In this paper, we study if Multi-Armed Bandits (MABs) are able to offer a feasible solution to the decentralized channel allocation and AP selection problems in Enterprise WLAN scenarios. To do so, we empower APs and stations with agents that, by means of implementing the Thompson sampling algorithm, explore and learn which is the best channel to use, and which is the best AP to associate, respectively. Our evaluation is performed over randomly generated scenarios, which enclose different network topologies and traffic loads. The presented results show that the proposed adaptive framework using MABs outperform the static approach (i.e., using always the initial default configuration, usually random) regardless of the network density and the traffic requirements. Moreover, we show that the use of the proposed framework reduces the performance variability between different scenarios. Results also show that we achieve the same performance (or better) than static strategies with less APs for the same number of stations. Finally, special attention is placed on how the agents interact. Even if the agents operate in a completely independent manner, their decisions have interrelated effects, as they take actions over the same set of channel resources.

\end{abstract}

\end{frontmatter}
\newpage
%%%%%%%%%%%%%%%%%%%%%%%%% Section 1: Introduction %%%%%%%%%%%%%%%%%%%%%%%%%
\section{Introduction}\label{sec:introduction}

% Intro sobre la necessitat de tenir xarxes denses
In the past few years, multimedia contents, such as social networks, virtual reality, on-demand video platforms and video-streamed gaming, have witnessed a remarkable growth in terms of bandwidth consumption. The widespread use of IEEE 802.11 based wireless local area networks (WLANs), commonly known as WiFi, has helped network providers to cope with the increasing demands of wireless communications \cite{cisco2018cisco}.

%Enterprise WLAN
Since the growing demand of data services, we can find enterprise WLANs in a wide range of private and public spaces. Enterprise WLANs are composed by several APs, also called basic service sets (BSSs), that coexist under the same extended service set (ESS), allowing different APs to keep the same service set identifier (SSID). Then, end-users, i.e., smartphones, laptops, or tablets perceive the whole set of deployed APs as a unique WLAN, allowing them to roam from one AP to another keeping the connectivity. Normally, such WLANs are deployed in places like airports, shopping malls, or university campuses, resulting in a very attractive solution to provide Internet access.

% Spectrum Scarcity & MAC Limitations
Although being a cost-effective solution, it is well-known that WiFi networks may suffer from severe performance degradation in dense deployments. First, a loss in performance can be related to the limited number of channels that are available in the Industrial Scientific and Medical (ISM) band, which make WLANs to highly suffer from co-channel and adjacent-channel interference (CCI and ACI) issues. For example, the need to avoid interference has lead to only use three of the different available channels in the 2.4~GHz band. During the years, the spectrum scarcity has been a severe problem. In order to prevent this situation, new IEEE 802.11 WLAN amendments, such as the IEEE 802.11ac \cite{ong2011ieee}, and the upcoming IEEE 802.11ax \cite{bellalta2016ieee,afaqui2016ieee,khorov2018tutorial} and 802.11be \cite{lopez2019ieee}, promote the use of 5~GHz and 6~GHz bands. 

On the other hand, the use of a random channel access mechanism may arise an unsatisfactory user experience in dense areas. In WiFi networks, the medium access control (MAC) is performed through the Distributed Coordination Function (DCF), which implements a carrier sense multiple access with collision avoidance (CSMA/CA) mechanism. The DCF operation is simple but functional, and in general, it performs well as long as the user density or throughput requirements are kept low. However, the higher the number of stations in a given area, the higher the probability of having unsuccessful data exchanges. In addition, this effect may be boosted by the strongest signal first (SSF) AP selection mechanism, which may create an unbalanced use of the different APs. The main reason is that the SSF is based purely on a physical (PHY) metric, such as the received signal strength indicator (RSSI), without taking into consideration how much traffic is already being handled by the destination AP. Although this approach may work under low load conditions, it is unfeasible to apply when considering dense scenarios. Then, this situation evokes to think in a different AP selection strategy based on a more complex metric capable to capture the network conditions (i.e. users' positions, current channel load, nodes' configuration, etc.)

% Solucions que presentem
As envisioned, resource management strategies must consider the instantaneous users' activity requirements as part of network configuration procedures. To proceed with this shift, we address the channel allocation (CA) and AP selection (APS) problems in enterprise WLANs through machine learning (ML) mechanisms, and in particular with the reinforcement learning (RL) branch. We adopt the well-known multi armed bandits (MABs) framework to implement a dynamic CA (DCA), as well as a dynamic APS (DAPS). To do so, we provide stations and APs with agents that are expected to autonomously learn the best performing action based on the environment. Besides, we address the learning problem in a decentralized adversarial multi-player context, in which APs and stations' agents, compete for a common set of resources using different action sets. Setting up this kind of adversarial multi-player environment is very challenging due to the fact that the reward experienced by an agent is conditioned not only by its own actions, but also by the actions performed by others\footnote{Note that we extend the definition of adversarial setting to the case where actions taken by other players change the distribution of our rewards. This situation could also be seen as a non-stationary setting, where environment conditions change along with the actions taken by the other players.}.

%Goal & contribution
In this work we want to assess the feasibility of a decentralized solution using learning MABs to concurrently perform channel allocation and AP selection. Rather than proposing an actual solution, the relevance of our work remains in the study of MABs algorithms in adversarial multi-player environments. Hence, the goal of this paper is to provide different insights on how nodes intelligently modify their configuration, by learning from past experiences to improve future performance. Then, efficient learning and intelligent decision-making algorithms are key to achieve such objective. In this paper, the lack of information, and the high level of uncertainty among the different actions are two conditions that have been addressed through the use of the MAB framework, as the decision-making process is carried out through a learning-by-interaction approach.

% Our contributions
The contributions of this paper are:
\begin{itemize}
	
	% A practical case
	\item We consider the use of MABs for network optimization in an adversarial multi-player setup. To do so, we take into account aspects such as user and network dynamics, asynchronous and continuous time operation of the agents, realistic reward computation, age of information, and the interaction between agents. 
	% The case
	\item We study the case of concurrent decentralized channel allocation and AP selection problems in large enterprise WLANs scenarios. We use the Thompson sampling (TS) algorithm as an action-selection strategy, showing the effectiveness of such a technique to adapt to changing conditions, as well as to adversarial scenarios where actions from one agent affect to the distribution of the system rewards observed by the other agents. 
	% Results
	\item We evaluate the system performance for different network setups (i.e., different traffic patterns, different number of stations and APs). We show that using the developed framework, the system is able to convergence to a better solution. Finally, we consider a non-stationary case beyond the default adversarial setting, in order to validate the ability of the TS algorithm to adapt to changes. 
	
\end{itemize}

% Paper structure

The remainder of this article is organized as follows. In Section \ref{sec:related_work} we describe the related work. Then, Section \ref{sec:sys_model} describes the system model under consideration as well as all the considerations to formulate the joint DCA and DAPS. The problem statement is assessed in Section \ref{sec:proposed_solution} jointly with the MABs framework, whereas the simulation results are presented in Section \ref{sec:system_eval}. Finally, we provide some conclusions in Section \ref{sec:conclus}.
%%%%%%%%%%%%%%%%%%%%%%%%%%%%%%%%%%%%%%%%%%%%%%%%%%%%%%%%%%%%%%%%%%%%%%%%%%%%

%%%%%%%%%%%%%%%%%%%%%%%%%% Section 2: Related work %%%%%%%%%%%%%%%%%%%%%%%%%
\section{Related work}\label{sec:related_work}

Channel allocation and AP selection in WiFi networks have been widely studied by the research community. In this section we overview some relevant works that can be found in the literature, with the aim to properly frame the contributions of this paper. Note that, since we focus in WiFi networks, we have only considered  works related to this technology.

\subsection{Channel Allocation}

Two main approaches regarding channel allocation have been discussed over the years. We refer to \textit{opportunistic channel allocation} \cite{niyato2009cognitive,xu2013decision,mishra2006distributed}, which is intended to overcome frequency holes, and promote higher frequency utilization by accessing them in a short-interval basis, and \textit{dynamic channel allocation}, which is intended to be responsive against changing network conditions in a long-term basis. 

In this paper, we are only interested in the study of DCA. In this regard, recent approaches have incorporated the use of ML to improve the channel allocation process. In \cite{hoyhtya2010classification}, authors propose an approach that exploits an smart channel selection strategy by classifying the traffic pattern on primary channels, and choosing the channel with the longest idle time. It is shown that using their approach the amount of collisions can be reduced drastically, increasing the system performance. Besides, \cite{lingzhi2015online} propose an online CA by adopting the MABs framework. On top of the MABs framework, they implement a weighted algorithm in order to carry the action selection process, in which the probability of selecting a certain action is adjusted according to the regret observed. More recently, authors in \cite{jeunen2018machine} proposed a channel allocation method based on graph analysis, linear programming and regression to minimize the overlap among APs. An study about the exploration-exploitation trade-off for different learning algorithms with the objective to achieve the best pair of channel and power allocation is presented in \cite{wilhelmi2019collaborative}. In addition, authors compare the performance of the different considered action selection strategies, while studying the implications of applying them under an adversarial setting. In \cite{barrachina2019online}, authors propose a dynamic-wise, light-weight and decentralized, online primary channel selection algorithm for performing dynamic channel bonding, which considers the activity on both target primary and secondary channels in order to maximize the expected throughput.

Finally, we can find other works that are based on a centralized architecture. Nowadays, this type of architectures are taken a lot of interest as they can get a global picture of the network state. Under this approach, a dynamic channel selection for sectorized WiFi cells is implemented in \cite{mack2014dynamic}. Here, the network is continuously monitored in order to identify the WiFi interference sources, so can be mitigated through establishing a configuration. In addition, other common framework used is the software defined network (SDN) paradigm, which is employed in \cite{seyedebrahimi2016sdn} to address the spectrum congestion in dense deployments. Then, the objective is to capture the network state, so an optimized channel assignment can be executed to minimize the interference. Although solutions based on centralized architectures are very powerful as the central controller has an overall picture of the network, these solutions may not be appropriate for high dynamic scenarios where the network state changes fast.

\subsection{Access Point Selection}

In regards of access point selection by the stations, we can identify different solutions depending on the optimization target. For instance, we can find mechanisms that try to minimize the number of stations per AP, whereas other schemes try to maximize the RSSI, or the throughput achieved. In this context, authors in \cite{tang2016throughput} evaluate an association control algorithm to optimize the throughput in WLANs. Bandwidth demands of users are considered as constraints to carry the association process, which is performed through estimating the AP utilization. In \cite{athanasiou2008cross} authors present two different association schemes. The first one is based on channel quality in both uplink and downlink, whereas the second one uses the airtime metric of each cell. In \cite{jabri2008ieee} the average workload of the network is used to redistribute the traffic when a new station joins the network or when the signal quality of a client deteriorates. The proposed approach however requires changes to the standard beacon frames. A similar scheme is presented in \cite{bejerano2004fairness} where the stations are migrated to the least loaded AP in order to balance the traffic load. However, since the channel quality is not considered, this approach may significantly reduce the aggregate throughput, as no consideration regarding the performance anomaly effect is done.

As well as in the case of CA, some articles explored the station association in centralized environments. In \cite{gong2013line}, authors explore an online AP selection process for 802.11n with heterogeneous clients (802.11a/b/g/n), with the objective to evaluate the impact of legacy clients. Moreover, authors in \cite{lin2017adaptive} use an SDN based solution to solve an unbalanced distribution of the stations among APs. The APs that are congested due to a high number of connected stations are requested to reconfigure their transmission power in order to force a hand-off process in some stations. However, this kind of approaches may not work properly since the number of attached clients is not an accurate estimator of the load. In this context, authors in \cite{coronado2017wi} propose, over a software defined WiFi network, an association scheme capable to detect situations in which the traffic is not efficiently distributed and so, reschedule to other APs the clients whose transmissions are causing performance issues.

In regards of works using ML techniques, we can already find some papers. Authors in~\cite{lingzhi2015online} also tackled the association process. They use the same weighted algorithm to perform both channel allocation and AP selection. In \cite{carrascosa2019decentralized} is proposed a decentralized approach to perform the AP selection through the MABs framework. In their solution, authors propose an extension of the epsilon greedy algorithm that includes stickiness to perform the AP selection, which results into a notable improvement in the system performance. It is important to mention that to the best of our knowledge and up to date, we have not found any other works that adopt ML for improving the association process.

\begin{table}[t]
	\small % text size of table content
	\centering % center the table
	\caption{Notation used in the system model}%
	\begin{tabular}{cl} % alignment of each column data
		\toprule
		\textbf{Notation} & \textbf{Description}\\ 
		\midrule
		$n$ & Number of APs in the network\\
		$m$ & Number of stations in the network\\
		$k$ & Number of available channels\\
		${\mathcal{A} = \{A_1,\dots, A_{n}\}}$ & Set of deployed APs\\
		${\mathcal{S} = \{S_1,\dots, S_{m}\}}$ & Set of deployed stations\\
		${\mathcal{C} = \{c_{1}},\dots, c_{k}\}$ & Set of available radio channels\\
		$\mathcal{S}_{j}$ & Stations attached to AP $j$\\
		$\mathcal{N}_{j}$ & Neighboring APs detected by AP $j$\\
		$\mathcal{N}_{j}^{c}$ & Neighboring APs using the same channel $c$ that AP $j$\\
		$\mathcal{A}_{i}$ & APs detected by station $i$ above the RSSI$_{\rm{th}}$\\
		$B_i$ & Bandwidth requested by station $i$\\
		$L_{\rm{d}}$ & Data packet size\\
		$r_{i,j}$ & Bit rate for station $i$, when associated to AP $j$\\
		$u_{i,j}(B_{j},L_{\rm{d}},r_{i,j})$ & Airtime required by station $i$ when associated to AP $j$\\
		$t_{s}(r_{i,j})$ & Duration of successful packet transmission between station $i$ and AP $j$\\
		$t_{e}$ & Duration of an empty slot\\
		$E[\psi]$ & Expected backoff duration\\
		$p_{e}$ & Packet error probability\\ 
		$\Psi_{j}^{c}$ & Average channel reward for an AP $j$ using channel $c$\\
		$\Omega_{j,i}$ & Average satisfaction for a station $i$ attached to an AP $j$\\\vspace{1pt}
		$\ell_{j}^{c}$ & Channel load experienced at AP $j$ using channel $c$\\
		\bottomrule
	\end{tabular}
	\label{tab:notation}
\end{table}

%%%%%%%%%%%%%%%%%%%%%%%%%%%%%%%%%%%%%%%%%%%%%%%%%%%%%%%%%%%%%%%%%%%%%%%%%%%%

\section{System model}\label{sec:sys_model}

In this section, we introduce the enterprise WLAN scenario considered in this paper. We expose the main assumptions that have been done, as well as presenting  the CSMA/CA abstraction used to model the WiFi operation. Finally, we introduce the performance metrics we will use in Section~\ref{sec:system_eval}. Table~\ref{tab:notation} summarizes the notation used through this paper.

% -------------------------------------------------------_
% -------------------------------------------------------_

\subsection{WiFi Network description}\label{subsubsec:network}

We consider an enterprise WLAN composed by a set of APs ${\mathcal{A} = \{A_1,\dots, A_{n}\}}$, and a set of stations ${\mathcal{S} = \{S_1,\dots, S_{m}\}}$, where $n$ and $m$ are the total number of APs and stations respectively. Over a given area, both types of devices are randomly placed following a uniform distribution. In addition, we consider that they all implement 802.11k and 802.11r amendments \cite{sanchez2016ieee}. The 802.11k amendment introduces new functionalities to support resource management, whereas the 802.11r amendment addresses the transition from one AP to another within the same WLAN aiming to minimize the interruption of connectivity.

In regards of APs, let us define their action set as ${\mathcal{C} = \{c_{1},\dots, c_{k}\}}$, which is composed by the different available radio channels. Thus, an AP~$j$ will select a channel ${c\hspace{0.5pt}\in\hspace{0.5pt}\mathcal{C}}$. It is worth mention that all the APs' share the same action space~${\mathcal{C}}$ and so, different APs may select the same channel~$c$. The set of stations within the Clear Channel Assessment (CCA) area of an AP~$j$ is denoted as $\mathcal{S}_{j}$, whereas the neighboring APs is expressed as~$\mathcal{N}_{j}$. Additionally, $\mathcal{N}_{j}^{c}$ will refer to the neighboring APs using the same radio channel~$c$ of AP~$j$. Note that since positions are assigned randomly, each AP may have different entries for $\mathcal{S}$ and $\mathcal{N}$.

On the other hand, from the stations' perspective, we define as $\mathcal{A}_{i}$ the action set for a station~$i$, which is composed by all the APs seen by station $i$ that are above a certain RSSI$_{\rm{th}}$ threshold, which is a system parameter added to improve the AP selection process\footnote{If an station~$i$ only detects one AP over the CCA threshold, whose RSSI is lower than the RSSI$_{\rm{th}}$, the action set~$\mathcal{A}_{i}$ will be composed only by this entry, as connectivity is ensured for all stations. Therefore, stations with a unique entry in their action set will not be able to perform any learning}. By not including APs below the RSSI$_{\rm{th}}$ threshold, we avoid exploring APs placed too far, in which the probability of being unsatisfied is significantly high. In addition, properly configuring this threshold allows to minimize the performance anomaly \cite{heusse2003performance}, which is produced when one station occupies the channel for a long time due to its low transmission rate, penalizing other stations that use higher rates. Notice that the size of the stations' action set may be different for each station, as it only depends on the number of detected APs a station $i$ detects above the mentioned threshold.

Only downlink traffic is considered. We model stations' activity using an on/off Markovian model, where both active periods (the station requires a certain downlink throughput) and inactive periods (the station is idle) are exponentially distributed with mean T$_{\text{on}}$ and T$_{\text{off}}$, respectively. Every time a station activates (moves to the 'on' state), we will say a new downlink traffic flow, or simply a flow, starts.

% -------------------------------------------------------_

\subsection{CSMA/CA model abstraction}\label{subsec:CSMA_abs}

In order to evaluate the DCA and DAPS over large-scale WLAN Networks for large periods of time (several hours), we abstract the CSMA/CA operation. While the considered abstraction does not capture low-level details of the PHY and MAC layers operation, it maintains the essence of the CSMA/CA: the 'fair' share of the spectrum resources among contending APs and stations. Basically, the considered abstraction takes into account the aggregate channel load at each AP to calculate the airtime that can be allocated to each station.

\begin{figure}[t]
	\includegraphics[width=\textwidth]{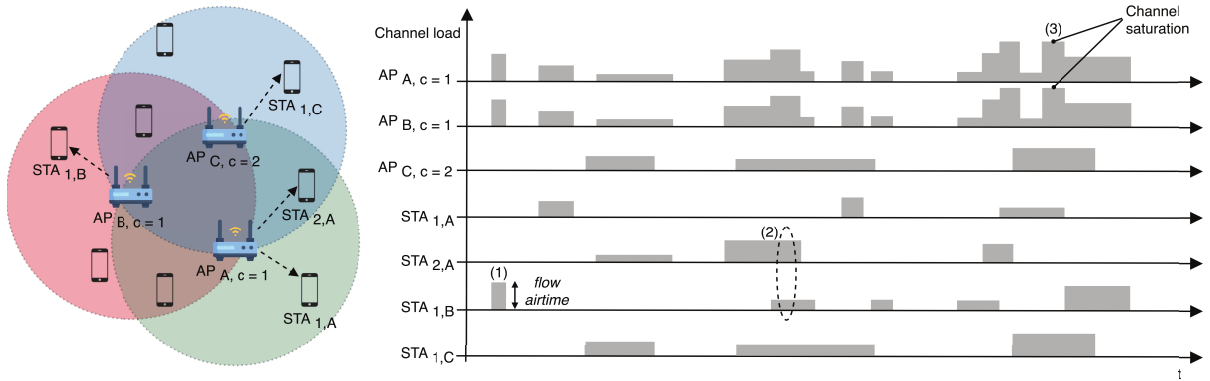}
	\caption{\small{Example of the CSMA/CA abstraction considered in this work. Horizontal axis represents time, whereas vertical axis represents the channel load. In the APs' axes it is represented the aggregate traffic, whereas the individual traffic is represented in the stations' axes. At (1), it is represented a single flow transfer. At (2), there are represented two concurrent traffic flows, and their repercussion on the neighboring APs's channel load using the same channel. At (3), AP$_{\rm{A}}$ and AP$_{\rm{B}}$ reach saturation conditions due to the higher traffic demands.}}
	\label{fig:Example}
\end{figure}

To explain how the proposed CSMA/CA abstraction works, refer to Figure~\ref{fig:Example}, where an illustrative example of an enterprise WLAN (3 APs, under the same ESS) is depicted. In this example, we consider that two APs use the same channel, whereas the remaining AP uses a different one. Some stations are deployed within the coverage area of each AP, which will act as receivers of the data flows. Depending on the throughput requirement $B_i$ of each flow, the length of a data packet $L_{\rm{d}}$, and the bit rate $r_{i,j}$, the total airtime that an station $i$ will require from its serving AP $j$ is given by
\begin{equation}\label{eqn:Airtime}
\centering
\begin{split}
u_{i,j}(B_{i},L_{\rm{d}},r_{i,j}) = \frac{1}{(1-p_{e})} \left\lceil{\frac{B_{i}}{L_{\rm{d}}}}\right\rceil(E[\psi]t_{e}+t_{s}(r_{i,j},L_{\rm d})),
\end{split}
\end{equation}
where $p_{e}$ is the packet error probability, the term $\frac{1}{(1-p_{e})}$ represents the average number of transmissions per packet, $E[\psi]$ is the average backoff duration, $t_{e}$ is the duration of an empty slot, and $t_{s}(r_{i,j},L_{\rm d})$ is the duration of a successful packet transmission, which is given by:
\begin{equation}\label{eqn:Ts}
\centering
\begin{split}
t_{s}(r_{i,j},L_{\rm d}) =  t_{\rm{RTS}} + 3t_{\rm{SIFS}} + t_{\rm{CTS}} +  t_{\rm{DATA}}(r_{i,j},L_{\rm d}) + t_{\rm{ACK}} + t_{\rm{DIFS}} + t_{e},
\end{split}
\end{equation}
where $t_{\rm{DIFS}}$ and $t_{\rm{SIFS}}$ are the distributed inter-frame space (DIFS) and the short inter-frame apace (SIFS), $t_{\rm{DATA}}(r_{i,j},L_{\rm d})$ is the duration of a data packet at the transmission rate used between station $i$ and AP $j$, and $t_{\rm{ACK}}$ correspond to the time that an acknowledgment (ACK) packet lasts. In \ref{appendix:A}, more details about how these parameters are computed can be found, as well as the description of the 11ax path-loss model considered to obtain the transmission rates.

As Figure~\ref{fig:Example} shows, at point (1), AP$_{\rm{A}}$ initiates a flow transfer to its attached station STA$_{\rm{1,A}}$. Due to this event, AP$_{\rm{B}}$ starts sensing the medium busy, registering an increment of the channel load. Note that this effect is a consequence of AP$_{\rm{B}}$ using the same channel as AP$_{\rm{A}}$. Moving to point (2), there are depicted two concurrent flows from AP$_{\rm{A}}$ and AP$_{\rm{B}}$ to their corresponding stations. Here, we observe that the effective channel load experienced by both APs corresponds to the aggregate airtime for all active flow transfers, either if they are due to their own flows or from neighboring APs. It is important to highlight that this effect will not happen if the APs were not allocated to the same channel. In addition, it is interesting to note that new incoming flows either from AP$_{\rm{A}}$ or AP$_{\rm{B}}$ to their stations may be successfully served, since the aggregate airtime of the active flows does not exceed the maximum allocable time\footnote{Since throughput is measured in Mbit/s, the maximum allocable time for transmissions is 1 second. Thus, if the effective channel occupancy experienced, and consequently the effective channel load, surpasses this value, the traffic requirements will not be met and so, we will talk about a saturated channel.}. Finally, at point (3), several concurrent flows lead AP$_{\rm{A}}$ and AP$_{\rm{B}}$ to experience channel saturation, as the maximum allocable airtime is achieved. Under this conditions, the airtime of each flow will be reduced in order to remain proportional to the maximum allocable value, and so, remaining at the saturation point. 

It is worth mention that AP$_{\rm{C}}$ does not perceive any change since it uses a different channel and so, its load is not affected by activity registered in the other channel. In  Section~\ref{subsubsec:satisfaction} a numerical example is presented to further illustrate these interactions.

% -------------------------------------------------------
% -------------------------------------------------------

\subsection{Performance metrics}\label{subsec:metrics}

To evaluate the performance of the system, we define two different metrics that will be used when carrying out the decision-making process. First, we define the channel reward metric for the DCA, which will be based on the channel occupancy. Then, we define the stations' satisfaction, which is related to the airtime, and it will be employed to evaluate the performance of the DAPS scheme. Both metrics will be used by the MAB framework introduced in Section \ref{sec:proposed_solution}.

% -------------------------------------------------------

\subsubsection{Channel reward}\label{subsubsec:ch_reward}

We define the channel reward as a metric to characterize the occupancy level experienced when using a certain channel. Since the purpose of this metric is to evaluate how channels are performing, it is only intended for APs to use it. To compute the channel reward, APs will register new instances of this metric every time they detect a change on the channel load in the channel in which they are operating. Then, every time the DCA agent activates, it will average all the obtained values, within a time interval specified by a temporal window, in order to get a quantitative representation of the channel occupancy level (i.e., the average channel load). In further Section~\ref{subsub:Activation_timers} the temporal window concept is presented. By using this metric, we consider that APs will have a detailed vision of the spectrum usage, accurately capturing all the intrinsic dynamics of a wireless environment. We define the channel reward for an AP~$j$ that uses a channel~$c$ at time~$t$ as
\begin{equation}\label{eqn:chreward}
\centering
\begin{split}
\Psi_{j}^{c} (t) = \text{max}(0, 1-\ell_{j}^{c}(t)) \leq 1,
\end{split}
\end{equation}
where $\ell_{j}^{c}(t)$ is the effective channel load at the AP $j$ when it uses the channel $c$ expressed as
\begin{equation}\label{eqn:chload}
\centering
\begin{split}
\ell_{j}^{c} (t) = \sum\limits_{\forall n\in \mathcal{N}_{j}^{c}}{\ell_{n}^{c
	}(t)} + \sum\limits_{\forall i\in \mathcal{S}_{j}}u_{i,j}(B_{i},L_{\rm{d}},r_{i,j}).
\end{split}
\end{equation}
Indeed, as stated in Section \ref{subsec:CSMA_abs}, $\ell_{j}^{c}(t)$ will depend on the channel load due to the own flows of AP $j$, and $\ell_{n}^{c}(t)$, which is the channel load registered due to flows from neighboring APs (using the same channel). Note that the effective channel load experienced can be higher than 1.

\subsubsection{Station's satisfaction}\label{subsubsec:satisfaction}

We adopt the concept of station's satisfaction to assess whether an association pair between an station and its serving AP is performing well or not, and therefore, it will be only used by stations. Conceptually, we define the satisfaction as the ratio between the required airtime by an station, and the actual amount that can be allocated by its serving AP. Then, we will refer to satisfied stations if the resulting value of the metric is one (traffic requirements fulfilled), whereas we will refer to unsatisfied stations if the resulting value is lower than one (traffic requirements can not be fulfilled).

To compute the metric, stations are intended to ask for the total amount of channel load that APs have experienced. In order to use this capability, stations are considered to be compliant with the 802.11k amendment, which defines the \textit{channel load request} messages. As defined in the standard, this type of frame is composed by a request-response sequence in which stations can ask for the amount of time in which the channel has been measured as busy (either through physical or the virtual carrier sense mechanism). As well as the channel reward, when the DAPS agent is activated, this metric averages all the tracked measures within a time interval specified by a temporal window, in order to get a quantification of the performance. Note that the satisfaction may change during the lifetime of a flow, and so we track all those changes, to average them at the end. We define the satisfaction for a station~$i$ associated to AP~$j$ operating in channel~$c$ at time~$t$ as:
\begin{equation}\label{eqn:satisfaction}
\centering
\begin{split}
\Omega_{i,j}(t) = \frac{\text{min}(1,\ell_{j}^{c}(t))}{\ell_{j}^{c}(t)} \leq 1
\end{split}
\end{equation}
where $\ell_{j}^{c}(t)$ is the channel load as defined in (\ref{eqn:chload}). Since we consider that all resources are proportionally allocated in our CSMA/CA abstraction, the satisfaction value obtained by stations under the same AP will be the same.

To clearly understand how this metric works, refer to the point marked as (2) in Figure~\ref{fig:Example}. Here, we assume that STA$_{\rm{1,A}}$ and STA$_{\rm{1,B}}$ require a traffic load of 40\% and 30\% respectively. As both APs share the same channel, the channel load perceived adds up to 70\%, lower than the maximum 100\%, and therefore, making stations to be satisfied as they receive the airtime allocation that they need. On the contrary, in point (3), we consider that STA$_{\rm{1,A}}$ still requires 40\%, but STA$_{\rm{1,B}}$ increases its traffic needs to 90\%. This higher requirement of STA$_{\rm{1,B}}$ makes APs to enter in saturation, since the total channel load raises up to 130\%. As a result, the satisfaction experienced by STA$_{\rm{1,A}}$ and STA$_{\rm{1,B}}$ scores a value of 76,9\%. Essentially, this value indicates that only the 76,9\% of the required airtime of both stations (i.e., 30.76\% for STA$_{\rm{1,A}}$, and 69.21\% for STA$_{\rm{1,B}}$) will be allocated. Again, note that, all active stations will receive the same satisfaction as we consider that resources are proportionally distributed.

Once we have the satisfaction, the throughput achieved by station~$i$, associated to AP~$j$ at time~$t$ is given by
\begin{equation}
\label{eq:throughput}
\Gamma_{i, j}(t)= B_{i} \Omega_{i,j}(t)
\end{equation}

% --------------------Problem formulation-----------------------------------

\subsection{Problem formulation}

By using airtime related metrics, we intend to make both APs and stations capable to keep track of the network congestion, and how it affects the stations' satisfaction. Here, we define the target objectives for DCA and DAPS, respectively. 

First, from the APs' perspective, the strategy to ensure a good network performance is to select the less congested channel. Therefore, the optimizations problem is reduced to maximize the channel reward and so, it can be expressed as
\begin{equation}\label{eqn:CH_Max}
\centering
\begin{split}
    c^* = \argmax_{\forall c \hspace{0.5pt}\in\hspace{0.5pt} \mathcal{C}} \hspace{2pt} \Psi_{j}^{c}
\end{split}
\end{equation}

On the contrary, from an station perspective, the strategy to enhance its own satisfaction is to minimize the congestion observed at its serving AP. Therefore, we have designed the DAPS relying on the satisfaction metric in order to decide whether an AP can be considered as a potential serving AP or not. Then, we formulate this problem as
\begin{equation}\label{eqn:SatMax}
\centering
\begin{split}
    a^* = \argmax_{\forall a \in \mathcal{A}_i}\hspace{2pt} \Omega_{i,a}
\end{split}
\end{equation}

We can observe that the optimization problem has been formulated as a maximization for both channel reward and station satisfaction. Then, using the proposed decentralized framework, APs and stations will take decisions autonomously to try to accomplish their target. Moreover, since decisions are taken asynchronously, we will see high variations when analyzing the behavior of the system as the reward observed by each agent will depend on the  action selection of all other agents. However, in the long term, we expect a reduction of the variance as the system will enter in a steady state regime, in which APs will have a fair distribution of the network load, and so of the stations associated to them.

% -------------------------------------------------------

%%%%%%%%%%%%%%%%%%%%%%%%%%%%%%%%%%%%%%%%%%%%%%%%%%%%%%%%%%%%%%%%%%%%%%%%%%%%%%%%%%
%%%%%%%%%%%%%%%%%%%%%%%%%% Section 4: Problem statement %%%%%%%%%%%%%%%%%%%%%%%%%%

\section{DCA and DAPS through the MABs framework}\label{sec:proposed_solution}

In this section we introduce the operation of the agent-based framework for decentralized channel allocation and access point selection, the MABs framework in which it is based, and the action-selection strategy that will be used following the performance metrics presented in Section \ref{sec:sys_model}.

% -------------------------------------------------------
% -------------------------------------------------------

\subsection{Multi-Armed Bandits}\label{subsec:MAB}

The multi-armed bandit problem models an interaction between a learning agent, often called player, and an environment. Traditionally, the agent decides on a number of alternative arms or actions, which iteratively pulls, one at a time, during a number of rounds ($t = 1,2,3,...,T$). From each action played, the learning agent receives a reward from the environment, which is used to evaluate the performance of the action, as well as to select subsequent actions. Then, the goal of the learning agent is to maximize the long-term reward to reach an optimal result. In addition, this strategy typically involves an exploration/exploitation trade-off, in which the agent must deal between learning at a faster or slower pace. To manage this trade-off, the learning rate parameter is used to balance both exploration and exploitation tasks in order to acquire enough knowledge to maximize the payoff. Note that a faster learning rate may lead to not exploring enough, ending into a suboptimal solution, whereas an slower learning rate may waste too much time on bad decisions. Therefore, tuning and selecting the appropriate learning rate is fundamental in order to achieve good results.

We can find different types of MABs depending on the characteristics of the reward. Typically, they are classified into stochastic, bayesian, contextual and adversarial bandits. For instance, in stochastic bandits actions have and independent and identical reward distribution, whereas in bayesian bandits, an arm is selected following a probability distribution that is proportional to the historic of the rewards experienced by that arm. Works such as \cite{yang2018multi}, \cite{sriyananda2016multi} \cite{xin2018active}, \cite{moura2019automatic}, \cite{al2012multi} show the wide variety of applications in wireless communications. In addition, a further and extensive introduction to the MABs framework can be found in \cite{slivkins2019introduction}. 

Independently of the type, the main objective of the MABs framework is to find the arm or action that maximizes the obtained reward. To do so, a common way of measuring the performance of MABs algorithms is by means of the regret function. The regret for a player $i$ at time $t$, after $T$ rounds as stated in \cite{vermorel2005multi}, is
\begin{equation}\label{eqn:regret}
\centering
\begin{split}
    R_{i,t} = Tr_{i,t}^{*} - \sum\limits_{t = 1}^{T}{r_{i,t}},
\end{split}
\end{equation}
where $r_{i,t}^{*}$ is defined as the reward given by the optimal action at time $t$, and $r_{i,t}$ is the reward obtained by the current action selected. From the regret definition, learn is said to happen if the cumulative regret function grows sublinearly, and therefore, the algorithm is able to identify the action with the highest reward. In this case, the expected regret, $\E[R_{i,t}]$, will decrease over time, converging to zero.

% -------------------------------------------------------
% -------------------------------------------------------

\subsubsection{Thompson sampling}\label{subsubsec:TS}

We use the Thompson Sampling (TS) \cite{thompson1933likelihood} algorithm to carry on the decision-making process.
The TS algorithm is a Bayesian algorithm that selects a given action based on its past noticed performance. To do so, during the learning stage, TS observes the reward, and updates its prior belief in a way that the probability of a particular arm being optimal matches with the probability of each arm being selected. In practice, this is done by sampling each arm from its posterior distribution, and selecting the one that returns the maximum expected reward. This property will result very useful, allowing us to tackle the intrinsic non-stationarity of our environment. Hence, arms that were chosen initially because of their good rewards, can be discarded over time if they start to perform badly. Section \ref{subsec:nonstationary_scenario} tackles this feature in-depth.

In our study, the prior belief on the rewards is assumed to be Gaussian distributed, as performed in \cite{wilhelmi2019collaborative}. Further details on the application of TS using Gaussian priors can be found in \cite{agrawal2013further}. Under this model, TS takes a sample for each action ($\theta_{x}$) according to a Gaussian distribution, which is provided by $\mathcal{N}(\hat{\mu}_{x}(t), \sigma_{x}^2(t))$, and so, selecting the action returning the maximum value of $\theta_{x}$.

For the considered distribution, the mean and variance are calculated as
\begin{equation}\label{eq:MeanVar}
\centering
\hat{\mu}_{i}(t) = \frac{\sum_{w = 1:i}^{t-1}{r_{i}(t)}}{n_i(t) + 1}\nonumber
\hspace{0.5cm}, \quad
\sigma_{i}^{2}(t) = \frac{1}{n_i(t) + 1}
\end{equation}
where $r_{i}(t)$ is the reward experienced for the action $i$ until round $t$, and $n_i(t)$ is the number of times that the action $i$ has been selected until round $t$. It is important to note that at the first TS iteration for each action will be given by a $\mathcal{N}(0, 1)$. The implementation of the TS algorithm can be found in Algorithm \ref{alg:thompsonS}.

Recently, authors in \cite{wilhelmi2019collaborative} have proved that TS performs better than $\epsilon$-greedy, upper confidence bound (UCB) and EXP3 in complex WiFi scenarios, as it results in faster convergence rates. In addition, empirical results from \cite{chapelle2011empirical} demonstrate as well that TS outperforms UCB despite its simplicity.

\begin{algorithm}[b]
	\vspace{2pt}
    \SetKwInOut{Input}{Input}	
	\Input{set of possible actions, $\mathcal{X}$ = \{$x_{1},..., x_{N}$\}}
	\vspace{2pt}
	\textbf{Initialize:} for each arm $x_i \in \mathcal{X}$, set $\hat{\mu}_{i} = 0$ and $n_{i} = 0$ \\
	\vspace{2pt}
	\While{active}
	{
    For each arm $x_i \in \mathcal{X}$, sample $\theta_{i}(t)$ from $\mathcal{N}(\hat{\mu}_{i}(t), \sigma_{i}^2(t))$\\
	\vspace{4pt}
	Select arm ${x_i}$ = $\argmax\limits_{i = 1,...,N}$ $\theta_{i}(t)$\\
    \vspace{2pt}
	Observe and compute the reward experienced $r_{i}(t)$\\
	\vspace{2pt}
	$\hat{\mu}_{i}(t) \leftarrow \frac{\hat{\mu}_{i}(t) \cdot n_{i}(t) + r_{i}(t)}{n_{i}(t) + 2}$\\
    \vspace{2pt}
	$n_{i}(t) \leftarrow n_{i}(t) + 1$\\
	}
	\caption{Implementation of Thompson sampling.}
	\label{alg:thompsonS}
\end{algorithm}

% -------------------------------------------------------

\subsection{Decentralized DCA and DAPS: an adversarial MAB approach}\label{subsubsec:AMABs}

In our study case, we consider that both DCA and DAPS problems fit into the adversarial MAB framework. The adversarial environment is developed through the fact that rewards experienced by actions depend on how the other players behave. So, from a point of view of single player, its opponents will have control over its rewards. Players are naturally classified into APs and stations, which rewards are defined accordingly to the metrics presented in Section~\ref{subsec:metrics}. It is important to note that players belonging to different groups may interact during the decision-making process. Indeed, an interesting contribution of this work is to consider these interactions between two different types of players.

\begin{figure}[t!]
\hspace*{\fill}%
 \begin{subfigure}[]{0.45\textwidth}
    \centering
     \includegraphics[ width=0.85\textwidth]{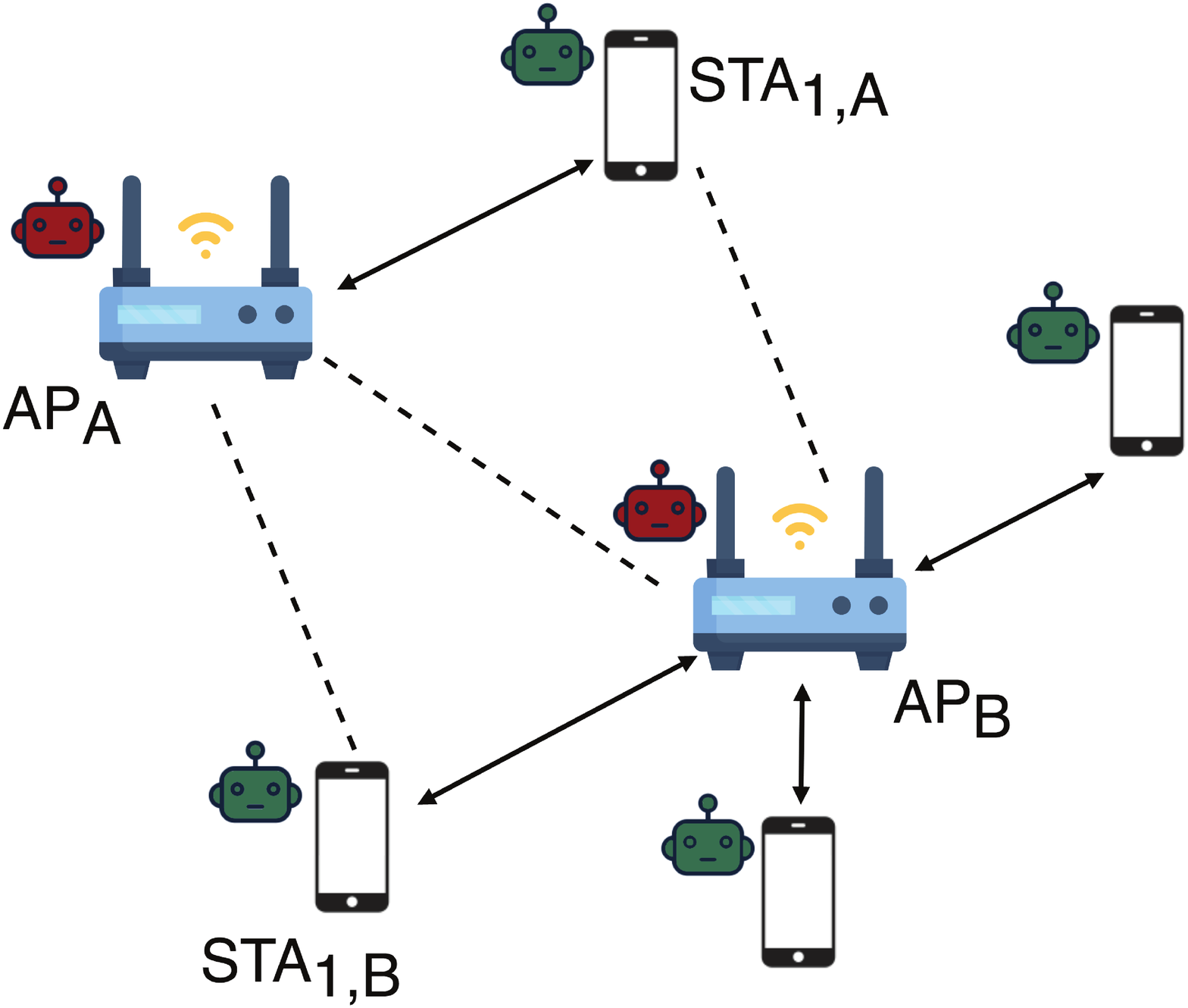}
     \caption{Topology example}
     \label{fig:Topology}
 \end{subfigure}
 \hspace*{\fill}%
 \begin{subfigure}[]{0.45\textwidth}
 \centering
     \includegraphics[ width=0.7\textwidth]{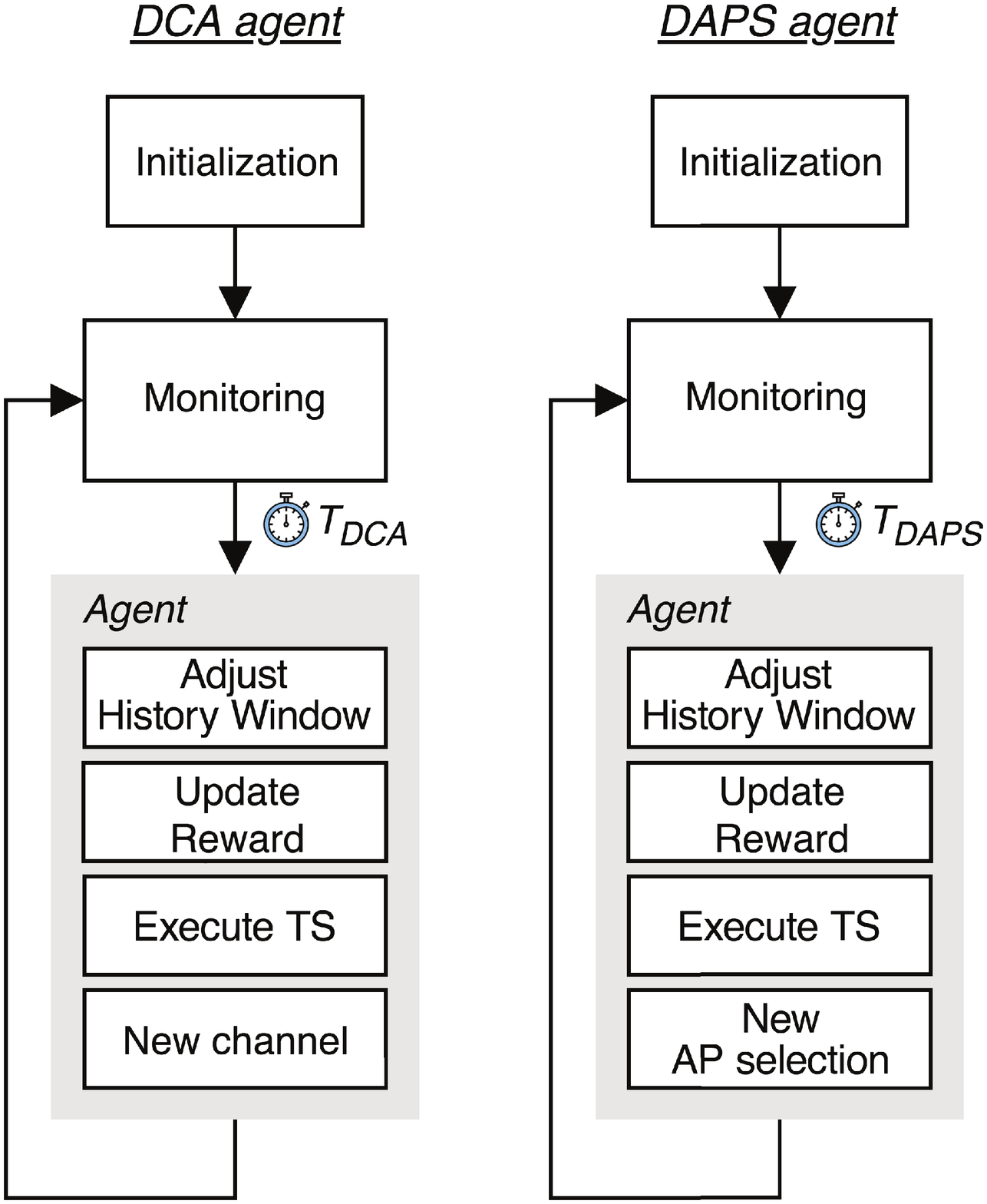}
     \caption{Flow diagram}
     \label{fig:Diagram}
 \end{subfigure}
  \begin{subfigure}[]{\textwidth}
  \centering
     \includegraphics[ width=0.9\textwidth]{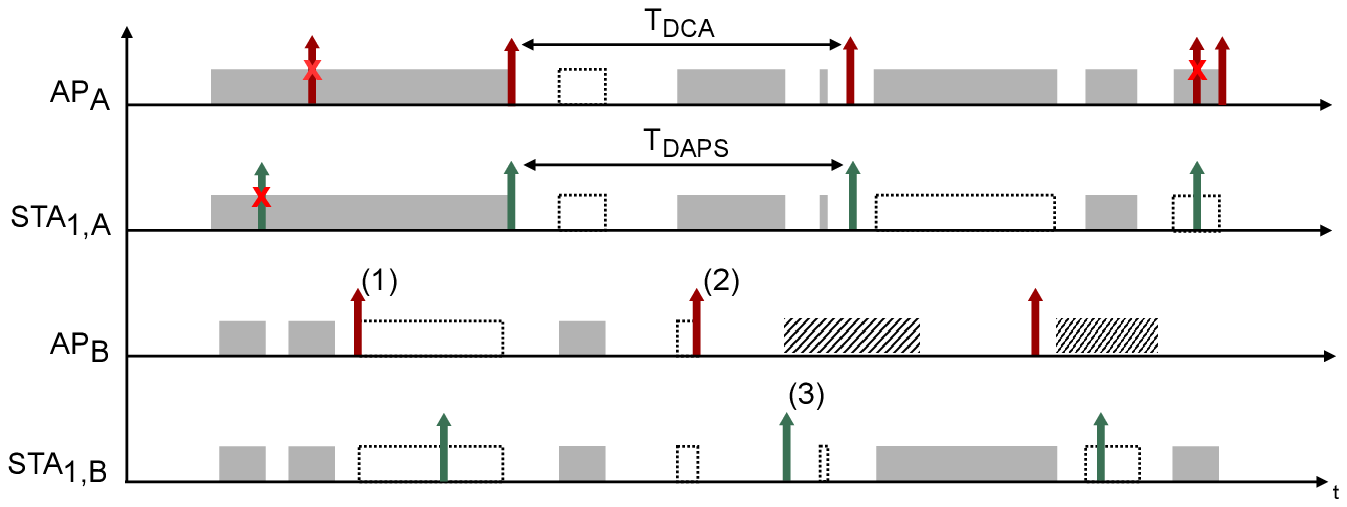}
     \caption{Time line operation of DCA/DAPS agents. Colored boxes represent active flow transfers from APs to stations, whereas doted boxes represent flow transfers sensed over the medium. Finally, boxes represented with a diagonal pattern are addressed to stations not represented in the time line. Red and green arrows represent that a DCA or a DAPS agent has been triggered, respectively.}
     \label{fig:TimeLine}
 \end{subfigure}
 \caption{Representation of the DCA and DAPS agent operation}
 \label{fig:DCA_DAPS}
\end{figure}

To further explain how agents perform, let us refer to Figure~\ref{fig:Topology}, in which a simple WLAN scenario is represented. Here, we consider that APs have selected different channels, whereas the stations have been considered to be attached following the SSF mechanism. Besides, APs and stations are considered to have the DCA (coloured in red) and DAPS (coloured in green) agents, respectively. For instance, focusing on the AP$_{B}$, it may experience a bad channel reward if it suddenly changes to the channel used by its neighbor. However, this player is also conditioned by the fact that negative rewards can be produced either if its neighbor changes to its operating channel, or if all the stations select it as the serving AP. Therefore, under an adversarial setting, players' decisions are highly conditioned to the ones made by the other contestants. This can be also confirmed from a station's point of view, as it may experience low satisfaction values if it select an overloaded AP, if its serving AP changes to a congested channel, or if many other stations switch to its serving AP.

In order to learn, we consider APs and stations to be equipped with an agent, which purpose is to select the most appropriate channel and AP, respectively, based on previously gathered experience. We define the concept of agent as a learner that performs background tasks, such as data-driven decisions, on behalf of an AP or an station. In Figure~\ref{fig:Diagram}, it is shown the agents' operation cycle, which comprises two phases. After the node initialization, we find the network \textit{monitoring} phase, in which the agent is intended to remain silent, observing and collecting the performance measurements for the last selected action. Although this phase does not involve the agent explicitly, it is considered part of its operation as it needs to observe the environment in order to acquire information. Then, once the timers, T$_{\text{DCA}}$ and T$_{\text{DAPS}}$, are expired, the second phase starts. We call it new action selection phase, and it comprises the performance evaluation of the last selected action, which process is subdivided in four steps.

\begin{enumerate}
    \item Adjust history window. Since agents keep track of all the data collected, they must get the past entries of the current action that fall inside the time boundaries specified by the sliding window size. By doing this procedure, agents average the performance observed during the \textit{monitoring} phase among past records. In further Sections~\ref{subsub:Activation_timers} and~\ref{SLW_AOI}, the sliding window concept is explained and analyzed.
    \item Update reward. Once agents obtain the average data, they must compute the parameters $\hat{\mu}_{i}(t)$ and $\sigma_{i}^2(t)$ for the current action, in order to update the probability distribution with the last performance data observed.
    \item Execute TS. After updating the estimated parameters of the distribution, the TS algorithm is executed by drawing a sample $\theta_{i}(t)$ from $\mathcal{N}(\hat{\mu}_{i}(t), \sigma_{i}^2(t))$.
    \item New channel/AP selection. The action returning the largest value of $\theta_{i}(t)$ will be selected as the new one. Then, agents will inform to APs and stations to update their configuration accordingly.
\end{enumerate}

Finally, a representative illustration of the agents' time line operation is depicted in Figure~\ref{fig:TimeLine}. For the sake of visualization we have only represented two of the different stations represented in Figure~\ref{fig:Topology}. In the point marked as (1), we find that after a proper monitorization of the environment, AP$_{\rm{B}}$'s DCA agent is triggered, motivating a channel switch for AP$_{\rm{B}}$. Immediately after this event, AP$_{\rm{B}}$ finds the medium busy as an ongoing data transfer is being carried by AP$_{\rm{A}}$. It is interesting to note that during that transfer, the DCA agent of AP$_{\rm{A}}$, as well as the DAPS agent of STA$_{\rm{1,A}}$, should have been triggered. However, since the ongoing transfer was still active, their timers have been postponed until its ending. In addition, asynchronous operations of the agents are naturally supported, which implies that during the \textit{monitoring} period (i.e., two consecutive inter DCA/DAPs agent activation epochs), we may observe other agents changing the configuration of their respective APs or stations. This effect is reflected in (2), in which AP$_{\rm{B}}$, again, changes its channel. Finally, at (3) STA$_{\rm{1,B}}$'s DAPS agent decides to change to AP$_{\rm{A}}$ in order to meet its traffic demands, decision that was caused by the unbalanced situation between APs.

\subsubsection{Activation Timers and Sliding Window}\label{subsub:Activation_timers}

Before getting into the performance analysis of the system under evaluation, we first assess the implications of the parameters used by the agents. So, we need to decide the rate at which agents will become active and choose a new action. From the point of view of learning agents, the more frequent the learning algorithm is executed, the faster will the learning process be. Thus, we need to choose the time interval between two consecutive agent activation epochs as short as possible to reduce the convergence time, while ensuring that network reconfiguration overheads can be assumed negligible. As we rely on IEEE 802.11r amendment, which reduces the roaming time between APs nearly to 50 ms, we consider that 3 minutes (180 s) is large enough and so, we set it as the time between two consecutive agent activation epochs for both DCA and DAPS.

On the other hand, the channel switch in WiFi networks is specified in the IEEE 802.11h amendment. This mechanism enables APs to announce a channel switch using channel switch announcement (CSA) frames before effectively moving to that channel~\cite{IEEE80211h}. When enabled, APs advertise through CSA frames the new channel, helping clients to switch to the target channel, saving scanning time. Therefore, clients, who support CSA, can perform the transition to the new channel with minimal downtime, instead of having to scan and discover the new channel in which their AP has switched. However, this process is not performed immediately as the AP sends a variable -\textit{and normally vendor-dependant}- number of frames, which contain the CSA announcement. Thus, the delay of the channel switch depends on the number of CSA frames being broadcasted. In this work, we have considered that this process may last up to 100 ms. Therefore, it allows us to set the time between two consecutive agent activation epochs also to 3 minutes (180 s), as the impact of the channel switching frames can be considered negligible.

Another consideration when using learning algorithms is the age of information. It is important to identify the existing trade-off between still valid and outdated information. In very dynamic scenarios, keeping track of old observations can lead agents to take decisions based on information that is outdated. However, not considering enough past data will reduce the ability to select a proper new action, as agents may lose useful information. To tackle this trade-off, we use the concept of sliding window, which is intended to filter the useful information from the outdated one. We will further discuss the impact of the window size in Section \ref{SLW_AOI}. Unless otherwise stated, we set by default the window size to 9 minutes (540 s), which corresponds to three agent activation periods.

%%%%%%%%%%%%%%%%%%%%%%%%%%%%%%%%%%%%%%%%%%%%%%%%%%%%%%%%%%%%%%%%%%%%%%%%%%%%%%%%%%

%%%%%%%%%%%%%%%%%%%%%%%%%% Section 5: Performance evaluation %%%%%%%%%%%%%%%%%%%%%

\section{Performance evaluation}\label{sec:system_eval}

In this section, we test the DCA and DAPS under different density conditions and throughput requirements to evaluate the performance of the learning MABs. To perform the evaluation, we have implemented from scratch our own simulator in C++ using the COST simulation libraries~\cite{COST}, which works as presented in Section~\ref{sec:sys_model}. The simulation platform developed is called Neko, and it can be found on GitHub\footnote{\url{https://github.com/wn-upf/Neko}}. The main reason that has led us to develop our own simulation tool is the need to simulate large scale networks for long periods of time.\footnote{As an example, it takes 6 minutes to simulate a medium-large scale network (100 APs and 1000 stations) in the Neko platform for a simulation time of 1~day, over an average quad-core Intel i5 3.8~GHz processor. On the contrary, in simulation platforms such as network simulator (ns) 3, it takes roughly 3~h to run a 1 minute simulation for a moderate scenario with 20 APs and 200 stations.} To achieve that goal, Neko considers the CSMA/CA abstraction described in Section~\ref{subsec:CSMA_abs}.

\begin{table}[t]
   \small % text size of table content
   \centering % center the table
   \caption{Simulation parameters}
   %\resizebox{\linewidth}{!}{%
       \begin{tabular}{ccc} % alignment of each column data
       \toprule
       \textbf{Parameter} & \textbf{Description} & \textbf{Value}\\ 
       \midrule
       $f_{c}$ & Central frequency band & \makecell{Depends on \\ channel selection}\\
       W & Channel bandwidth & 20 MHz\\
       P$_{tx}$ & Transmission power & 15 dBm\\
       G$_{tx}$ & Antenna transmission gain & 0 dB\\
       G$_{tx}$ & Antenna reception gain & 0 dB\\
       PL (d) & Path loss & Refer to \ref{appendix:A}\\
       CCA & CCA threshold & -80 dBm\\
       RSSI$_{\rm{th}}$ & RSSI threshold & -75 dBm\\
       N$_{\text{ss}}$ & Number of spatial streams & 2\\
       $L_{\rm{d}}$ & Data packet size & 12000 bits\\
       T$_{\text{on}}$ & Avg. connection duration & 1 s\\
       T$_{\text{off}}$ & Avg. connection interarrival time & 3 s\\
       CW$_{\text{min}}$ & Min. contention window & 16\\
       $p_{e}$ & Packet error rate & 0.1\\
       T$_{\text{sim}}$ & Simulation time & 86400 s\\
       T$_{\text{DCA}}$& \makecell{Time to trigger AP activation agent} & 180 s\\
       T$_{\text{DAPS}}$ & \makecell{Time to trigger station activation agent} & 180 s\\
       T$_{\text{sw}}$ & Sliding window interval & 540 s\\
       P$_{\text{th}}$ & Performance threshold & 85\%\\
       \bottomrule
       \end{tabular}
       %}
   \label{tab:params}
\end{table}

\subsection{Simulation set-up}\label{subsec:simul_params}

All considered scenarios consist of multiple IEEE 802.11ax-capable APs and stations. All of them include capabilities of the IEEE 802.11k and 802.11r amendments, as mentioned in Section~\ref{sec:sys_model}. In all scenarios the transmission power of APs and stations is set to 15~dBm. In all cases, we guarantee that stations detect at least one AP (i.e., the received power is higher than the CCA threshold), as otherwise, the station is re-located. The expected duration of the downlink traffic flows, and the expected time between two consecutive flows is given by T$_{\text{on}}$ and T$_{\text{off}}$, respectively. Transmission rates between stations and their serving APs are determined by the RSSI of the link. We have selected the IEEE 802.11ax path-loss model for enterprise scenarios \cite{merlin2015tgax}, which can be found in the \ref{appendix:A}. The main reason to select this path-loss model is that it takes into account the effect of walls, as the 5~GHz band is very sensitive to this parameter. Finally, we have provided APs with 3 different channels of the UNII-1 band. We have constrained the channel availability to numbers 36 (5.18~GHz), 40 (5.20~GHz) and 44 (5.22~GHz) in order to reduce the action set for the APs, as well as to get the most of the spectrum reuse. In Table~\ref{tab:params}, there are detailed the other parameters considered to obtain the results. Finally, physical and MAC layer parameters of the IEEE 802.11ax standard are shown in \ref{appendix:A}.

\subsection{Toy scenario}\label{subsubec:toy}

For this illustrative use case, we have designed a controlled environment to study the interactions and behavior of the network when applying DCA and DAPS independently. To this purpose, we have deployed three APs in a line with partial overlapping coverage areas, and 45 stations, which have been distributed uniformly in a 3D-space with dimensions 25 x 25 x 2~m (x, y and z axes, respectively). The throughput required by each station is randomly chosen in the range [1-5]~Mbps every time a new flow starts (i.e., when the station moves to the 'on' state) in order to tackle standard traffic demands. For instance, around 5~Mbps is the recommended bandwidth for high definition video quality. Regarding the action space for stations, it is important to remind that each station will construct its action set independently of the others, as it depends on the number of APs sensed over the RSSI$_{\rm{th}}$. Figure~\ref{fig:initialAssociation}, shows the considered deployment and how the different APs and stations are placed. Note that the same color scheme in the APs indicate that they have been configured with the same channel, whereas different colors will indicate different channels. For this controlled use case, as the number of APs is very low, we have limited the use of channels to numbers 36 and 40. Finally, the simulation parameters correspond to the ones presented in Table~\ref{subsec:simul_params}.

\begin{figure}[!b]
     \centering
     \begin{subfigure}[]{0.32\textwidth}
         \centering
         \includegraphics[ width=\textwidth]{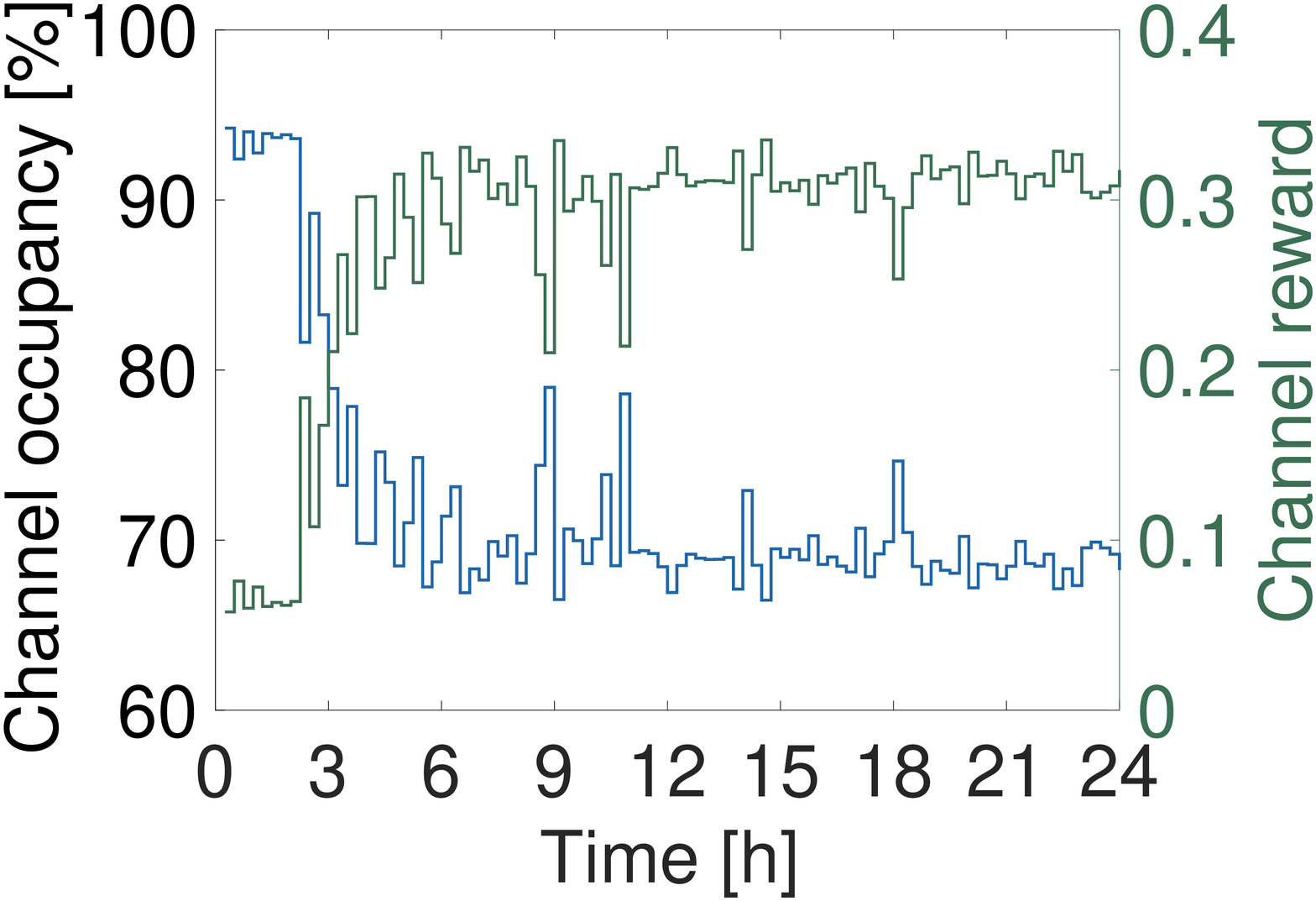}
         \caption{AP1}
         \label{fig:AP1CHRew}
     \end{subfigure}\hfill% 
     \begin{subfigure}[]{0.32\textwidth}
         \centering
         \includegraphics[ width=\textwidth]{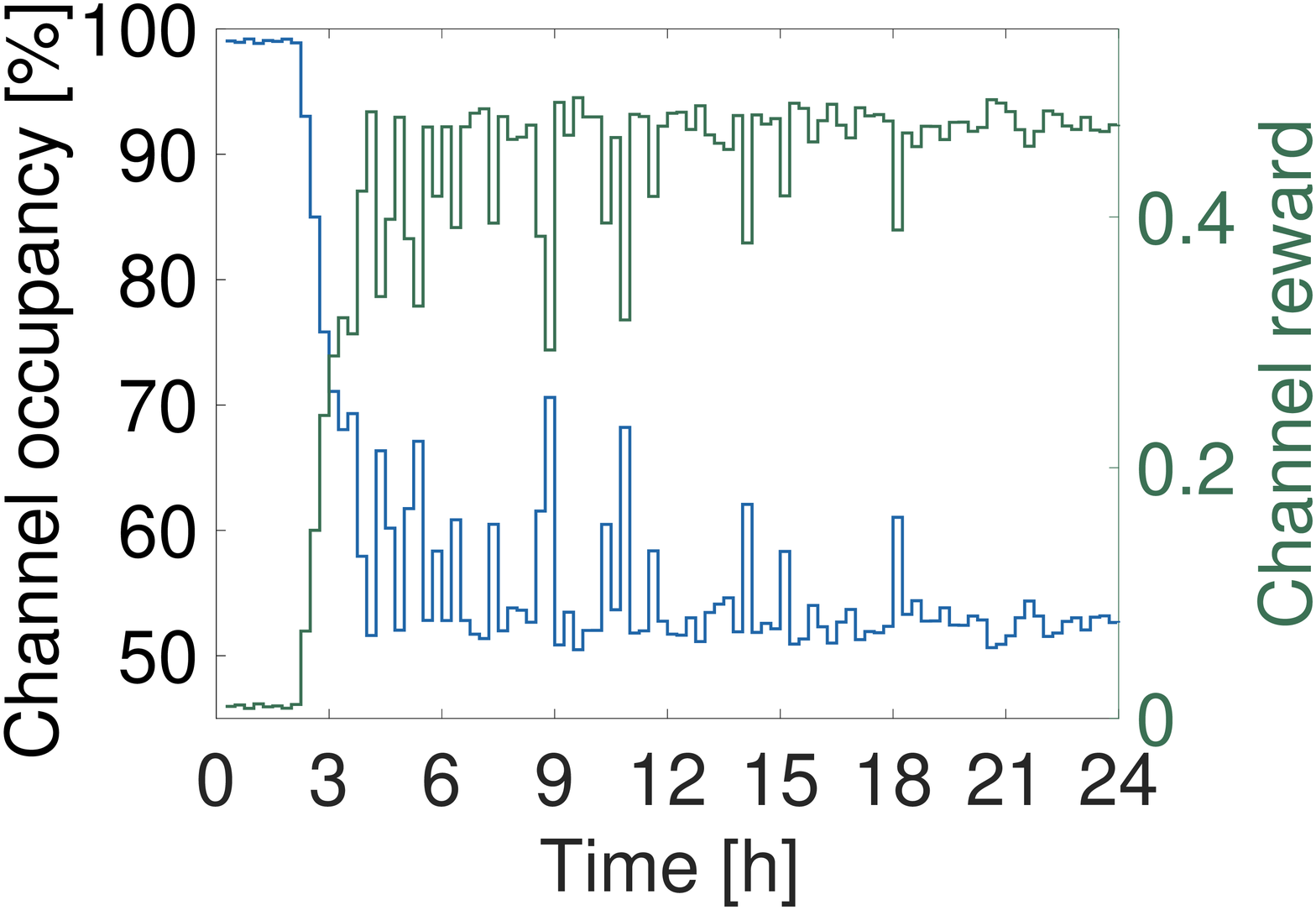}
         \caption{AP2}
         \label{fig:AP2CHRew}
     \end{subfigure}\hfill% 
     \begin{subfigure}[]{0.32\textwidth}
         \centering
         \includegraphics[ width=\textwidth]{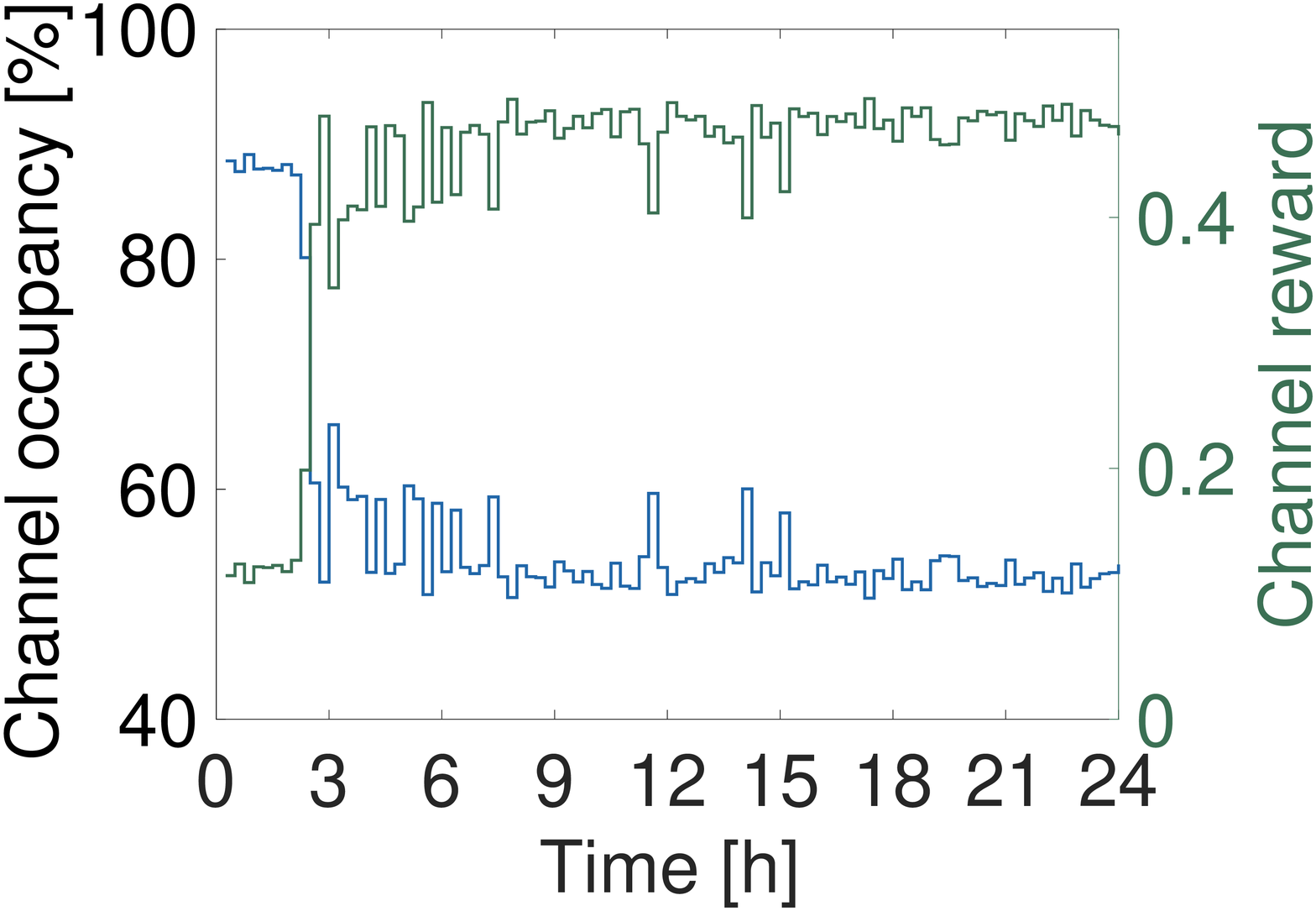}
         \caption{AP3}
         \label{fig:AP3CHRew}
     \end{subfigure}
     \caption{\small{Evolution of the APs' performance. In blue is represented the channel occupancy, whereas in green is represented the channel reward}}
     \label{fig:CHR_CHO}
\end{figure}

\begin{figure}[!b]
     \centering
     \begin{subfigure}[]{0.32\textwidth}
         \centering
         \includegraphics[ width=\textwidth]{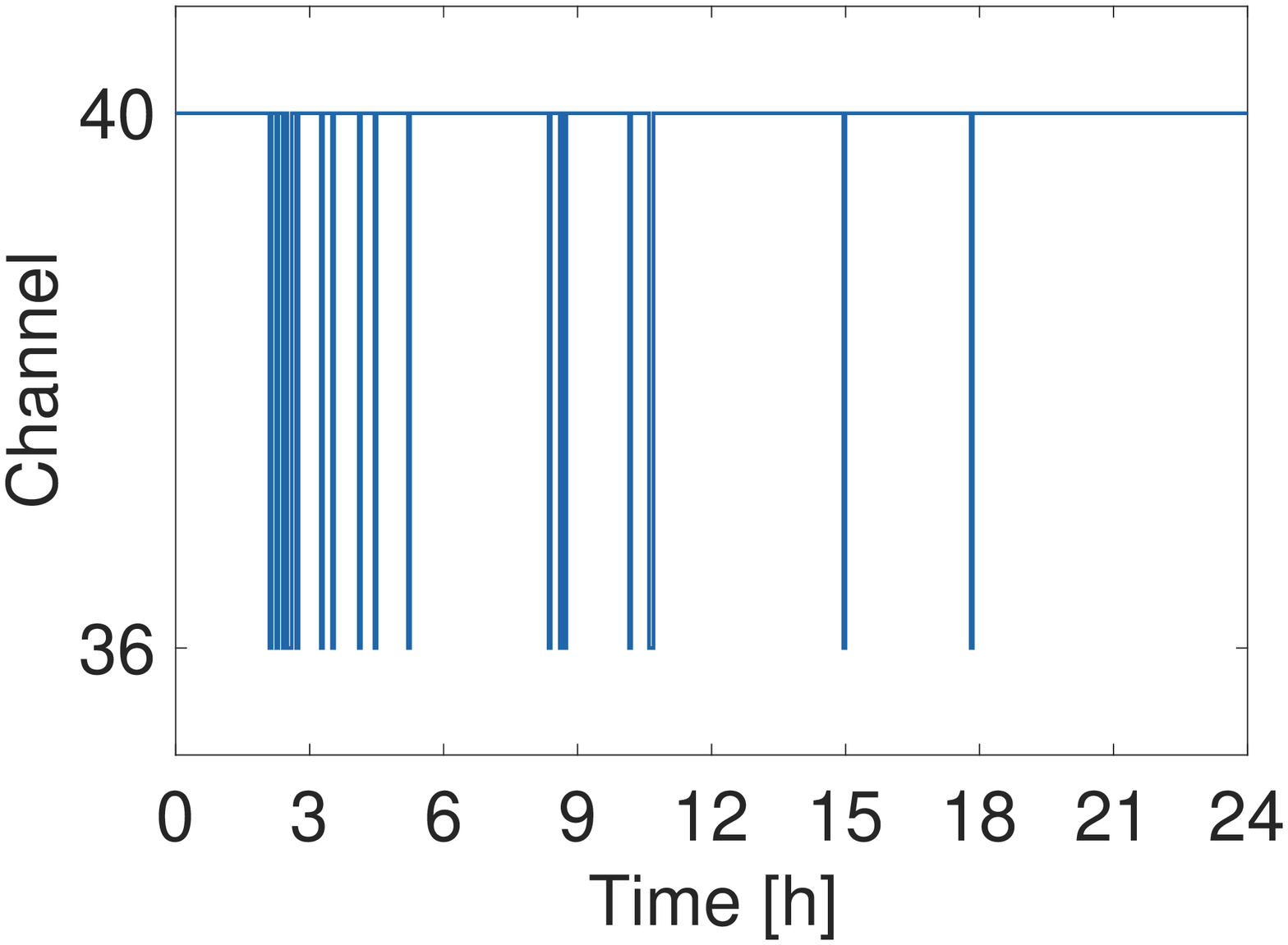}
         \caption{AP1}
         \label{fig:AP1CHAct}
     \end{subfigure}\hfill% 
     \begin{subfigure}[]{0.32\textwidth}
         \centering
         \includegraphics[ width=\textwidth]{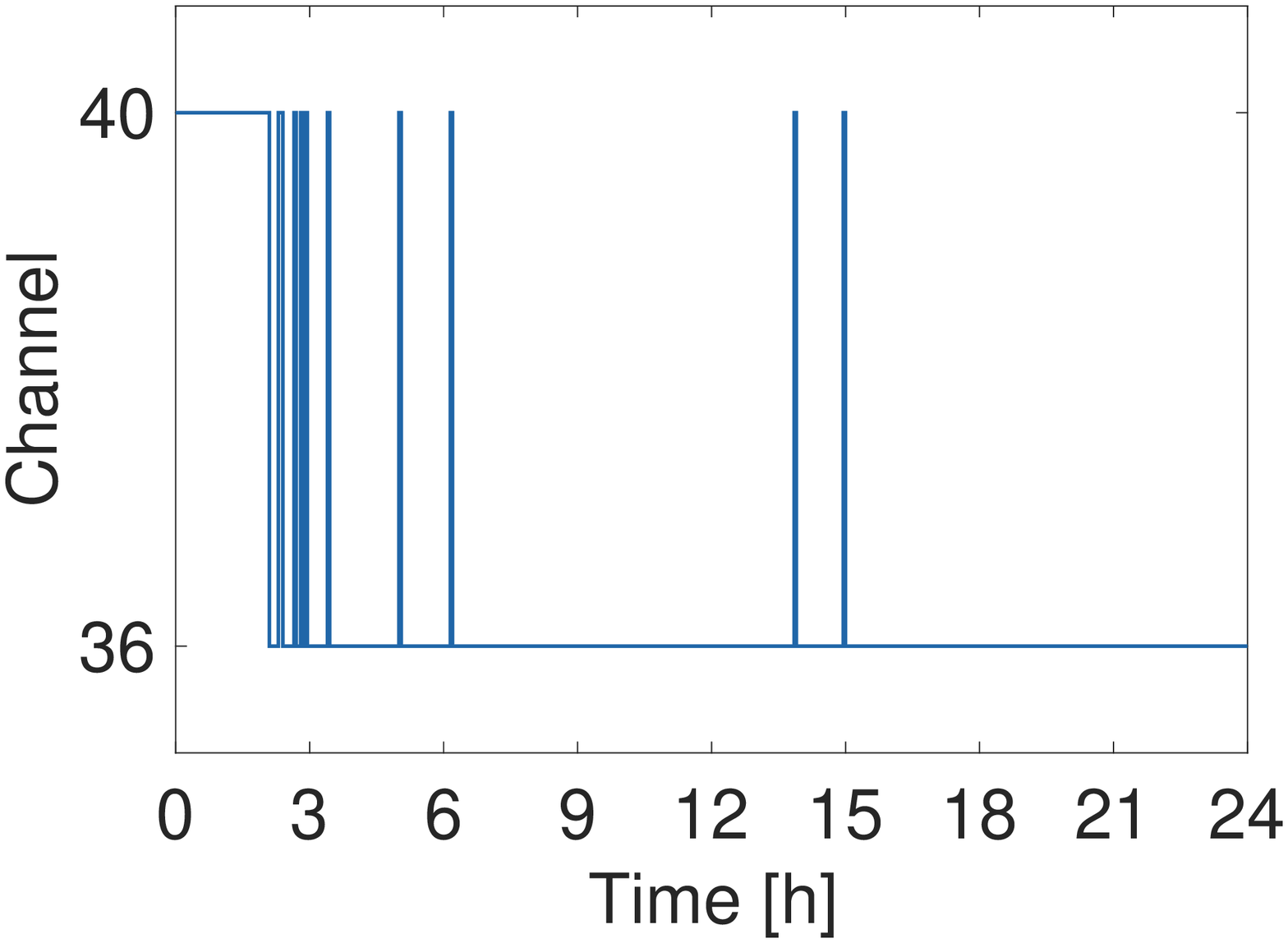}
         \caption{AP2}
         \label{fig:AP2CHAct}
     \end{subfigure}\hfill% 
     \begin{subfigure}[]{0.32\textwidth}
         \centering
         \includegraphics[ width=\textwidth]{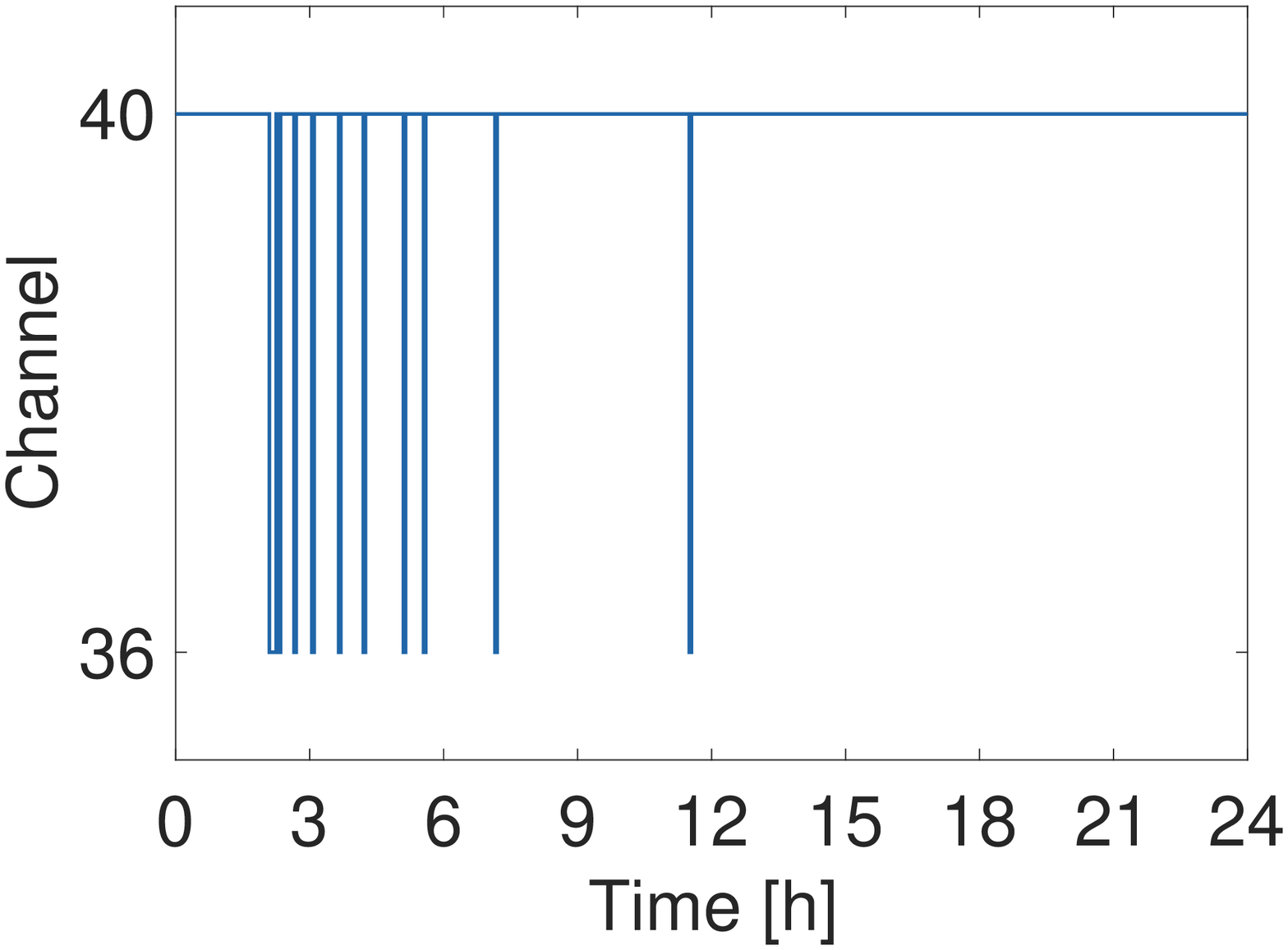}
         \caption{AP3}
         \label{fig:AP3CHAct}
     \end{subfigure}
     \caption{\small{Action evolution for the APs}}
     \label{fig:CHAction}
\end{figure}

\subsubsection{DCA evaluation}

Firstly, we are going to evaluate the effect of applying the DCA mechanism in the network performance. To do so, we configure the three APs with the same radio channel, enabling the DCA agent on them. On the contrary, all the stations do not have their DAPS agent enabled, so we can analyze the implications of employing the DCA alone.

Figure~\ref{fig:CHR_CHO} shows the occupancy of each AP. From $t = 0$~h and $t = 2$~h, we observe the first static stage in which APs remain with the initial configuration (i.e., all APs are configured with the same channel). Then, from $t = 2$~h and $t = 8$~h APs' DCA agents start the learning stage until they converge into a solution. From Figures~\ref{fig:AP1CHRew} and~\ref{fig:AP2CHRew}, for both AP$_{1}$ and AP$_{2}$, we can see two peaks at $t = 9$h that indicate a bad exploration from one of the two APs. However, this effect is not reproduced in AP$_{3}$, as seen in Figure~\ref{fig:AP3CHRew}, due to AP$_{1}$ being out of range from the CCA threshold of AP$_{3}$. To analyze the consequence behind these peaks, we have represented the action evolution for all three APs. In Figure~\ref{fig:CHAction}, it can be observed that AP$_{1}$ is the cause of the aforementioned peaks, since its channel switch from 40 to 36 downgrades not just its own reward, but AP$_{2}$'s reward too. Moreover, we can see that during the learning stage, the TS algorithm constructs its probabilistic model as the time advances by exploring different actions. Consequently, at the end, exploring is less frequent and exploitation is almost exclusively performed. It is worth noticing that channel reward is inverse to the AP occupancy as stated in Section~\ref{subsubsec:ch_reward}.

\subsubsection{DAPS evaluation}\label{subsub:DAPS}

Now, we evaluate the effect of the DAPS mechanism. To do so, again, we configure the three APs with the same radio channel, but deactivating their agents. Now, all the stations have their DAPS agent activated. The initial association at the beginning of the simulation is done through the SSF mechanism. Before running the simulation, we can observe from this configuration that AP$_{2}$ will suffer from starvation, and AP$_{1}$ and AP$_{3}$ will be able to allocate most of the available airtime to their stations. Therefore, stations attached to AP$_{2}$ will be encouraged to re-associate as they will receive a poor satisfaction value. 

\begin{figure}[!b]
     \centering
     \begin{subfigure}[]{0.32\textwidth}
         \centering
         \includegraphics[ width=\textwidth]{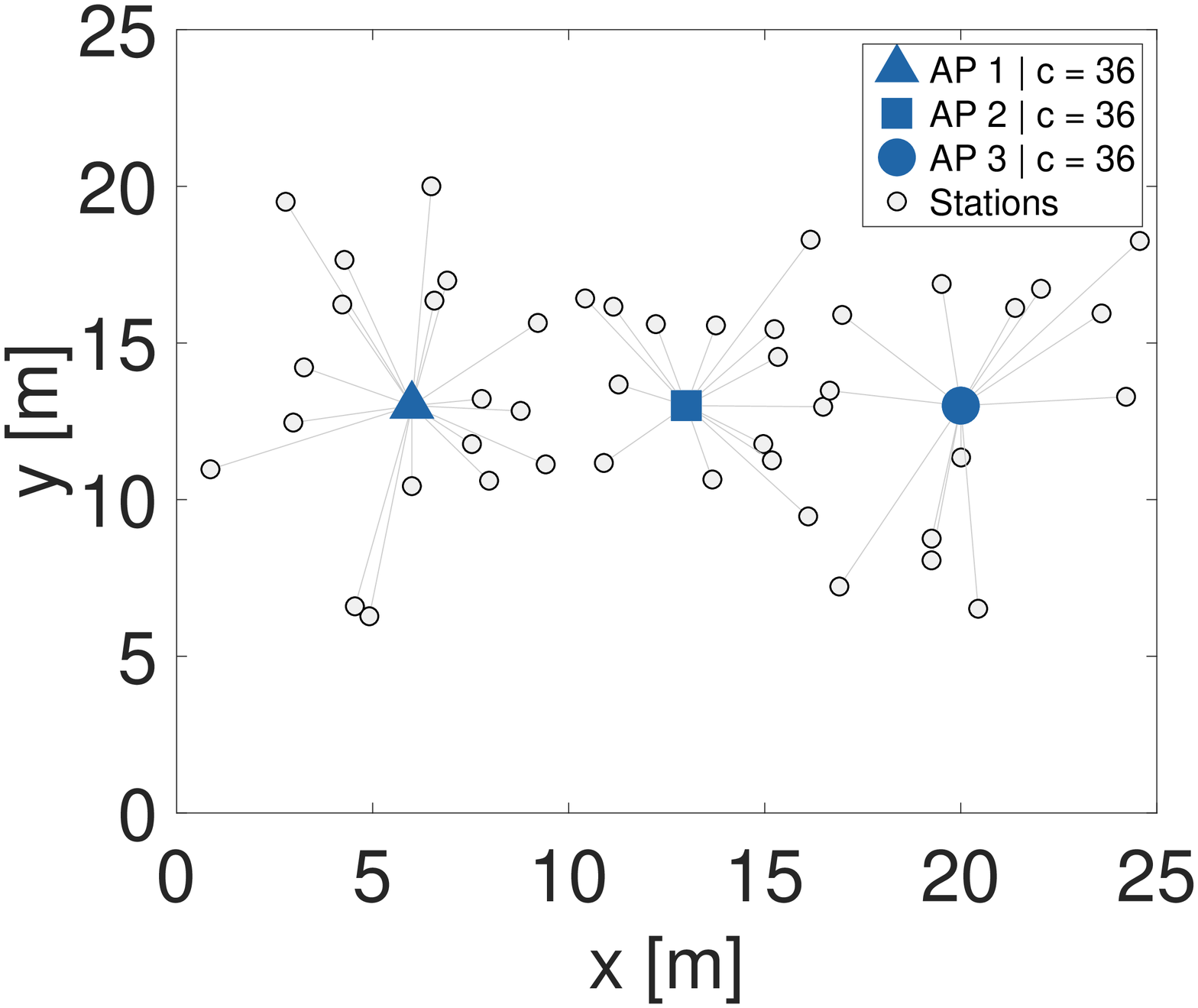}
         \caption{Initial association}
         \label{fig:initialAssociation}
     \end{subfigure}
     \hfill
     \begin{subfigure}[]{0.32\textwidth}
         \centering
         \includegraphics[ width=\textwidth]{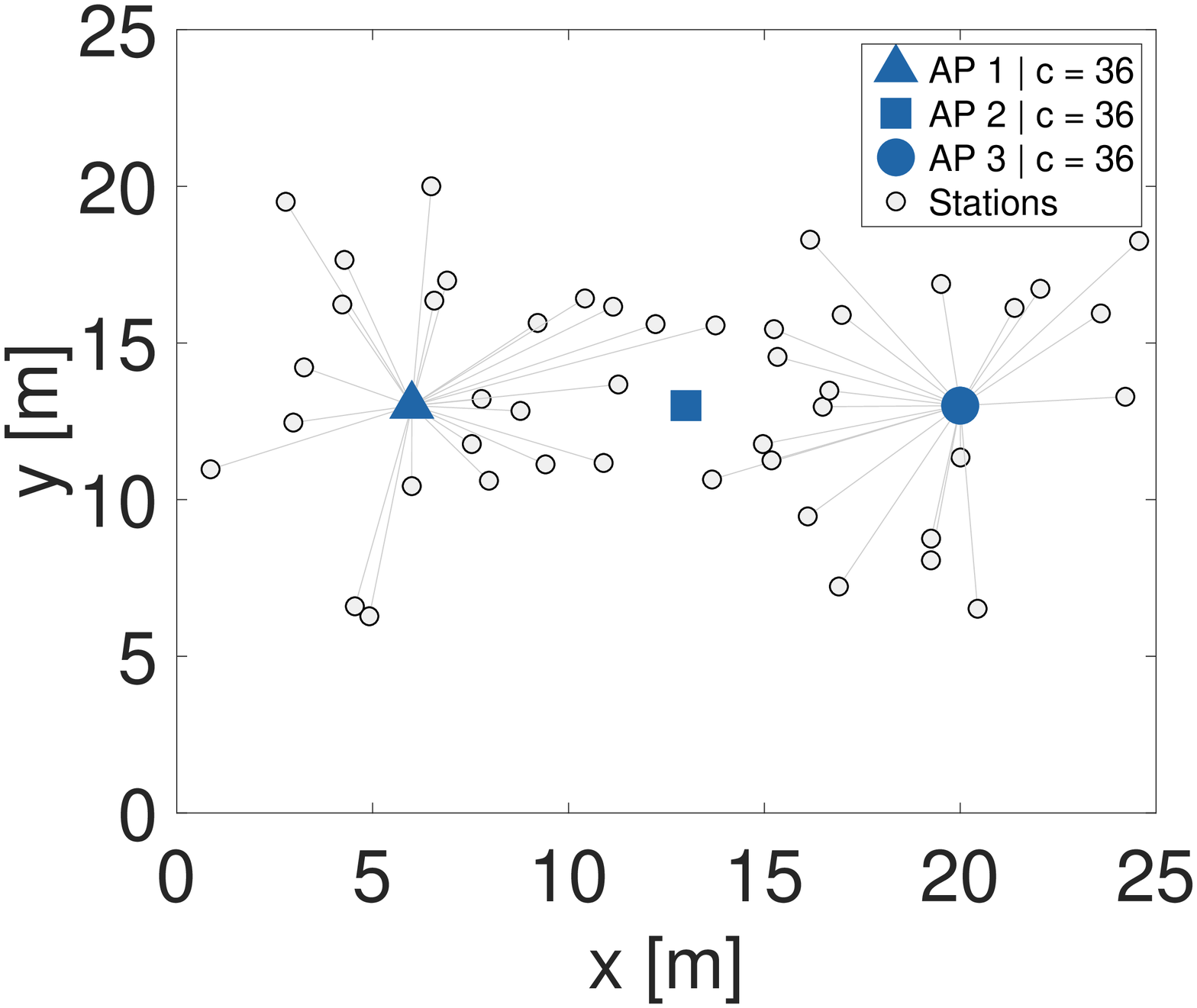}
         \caption{Final association}
         \label{fig:finalAssociation}
     \end{subfigure}
     \hfill
     \begin{subfigure}[]{0.32\textwidth}
         \centering
         \includegraphics[ width=\textwidth]{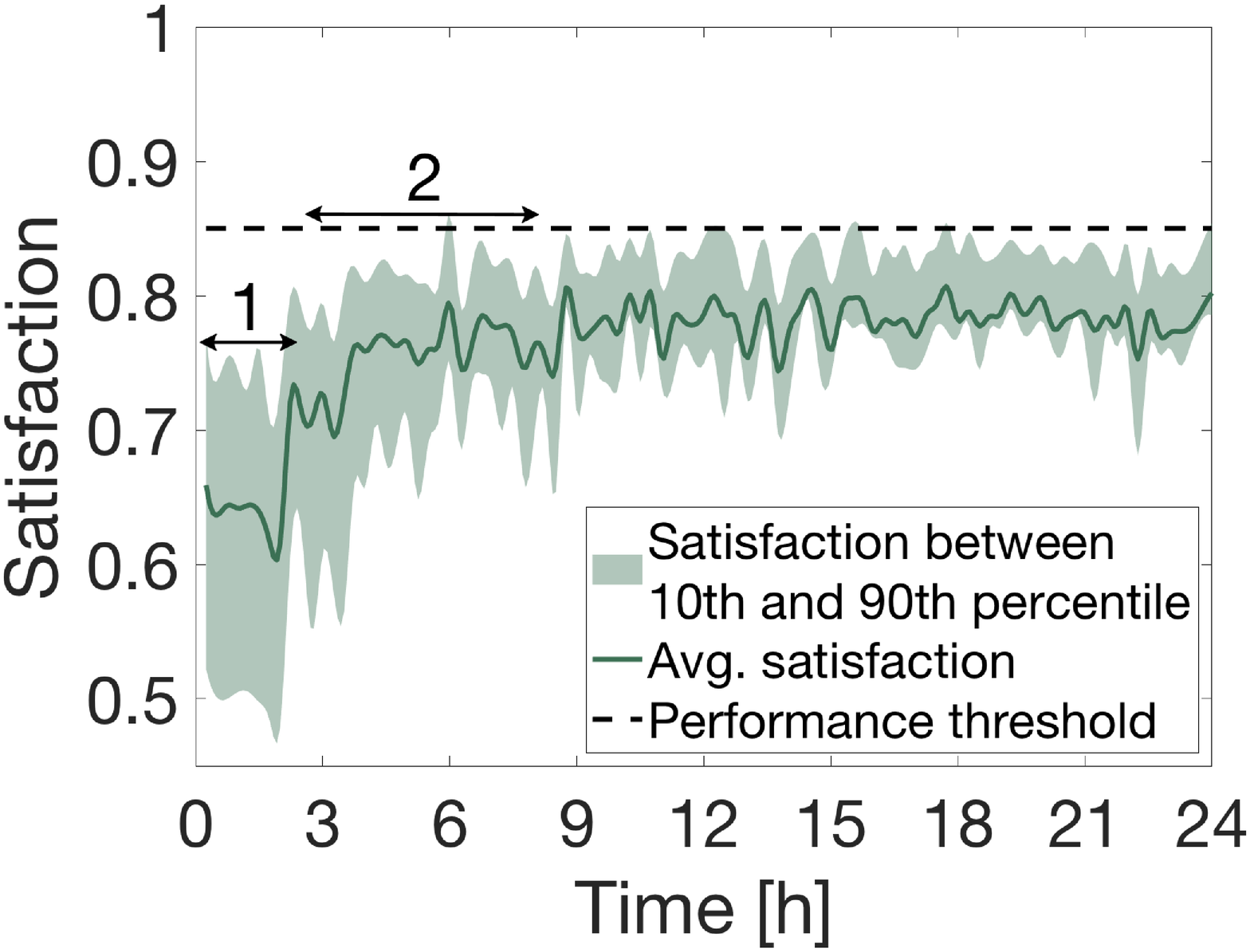}
         \caption{Satisfaction evolution}
         \label{fig:toy_percentile}
     \end{subfigure}
     \caption{\small{Performance of DAPS in the toy scenario}}
     \label{fig:AUSA}
\end{figure}

Figure~\ref{fig:finalAssociation} shows the final association scheme of the scenario, and as expected, the stations that have perceived low satisfaction values leave their serving AP, in order to associate to a different AP that ensures a higher airtime allocation, even if that means using lower transmission rates. Figure~\ref{fig:toy_percentile} shows the average satisfaction of all the stations. It can be observed that during the beginning of the simulation (stage marked as 1) the traditional association performed badly. However, at $t = 2$~h, the DAPS agent is activated, and stations are allowed to explore different APs during the learning stage (marked as 2). After some time, stations converge into a solution, even though it is below the performance threshold set (Table~\ref{tab:params}). Therefore, we consider that in this case, using only the DAPS agent at the stations, the network is not able to reach a feasible solution. This effect is related to the fact that APs sharing the same radio channel prevent any feasible re-association option. Then, we can conclude that the performance of the DAPS mechanism is severely conditioned to the efficiency of the DCA mechanism to properly allocate orthogonal channels to overlapping BSSs. However, we find that the use of DAPS can be useful to overcome an unbalanced distribution of the stations.

\subsubsection{Concurrent DCA and DAPS}

Finally, we consider the case in which both agents operate concurrently in the toy scenario. Again, the initial channel allocation is the same as for all APs, whereas stations are associated following the SSF mechanism. By evaluating the concurrent operation of DCA and DAPs, we expect to see the effects over the network of both DCA and DAPS agents, so we can observe the potential advantages of running them at the same time.

\begin{figure}[!b]
     \centering
     \begin{subfigure}[]{0.3\textwidth}
         \centering
         \includegraphics[ width=\textwidth]{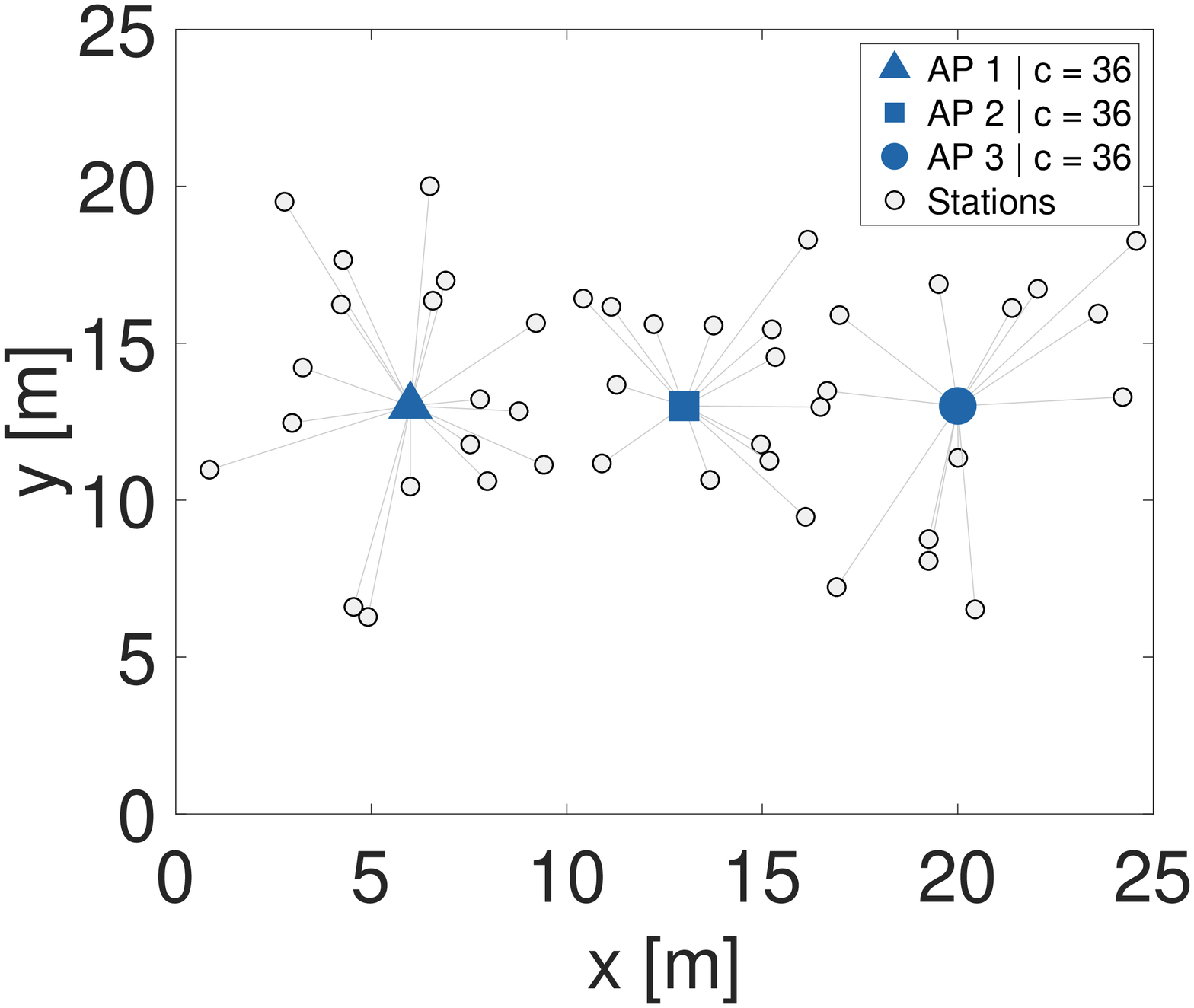}
         \caption{Initial scheme}
         \label{fig:initialAssociationBoth}
     \end{subfigure}
     \hfill
     \begin{subfigure}[]{0.3\textwidth}
         \centering
         \includegraphics[ width=\textwidth]{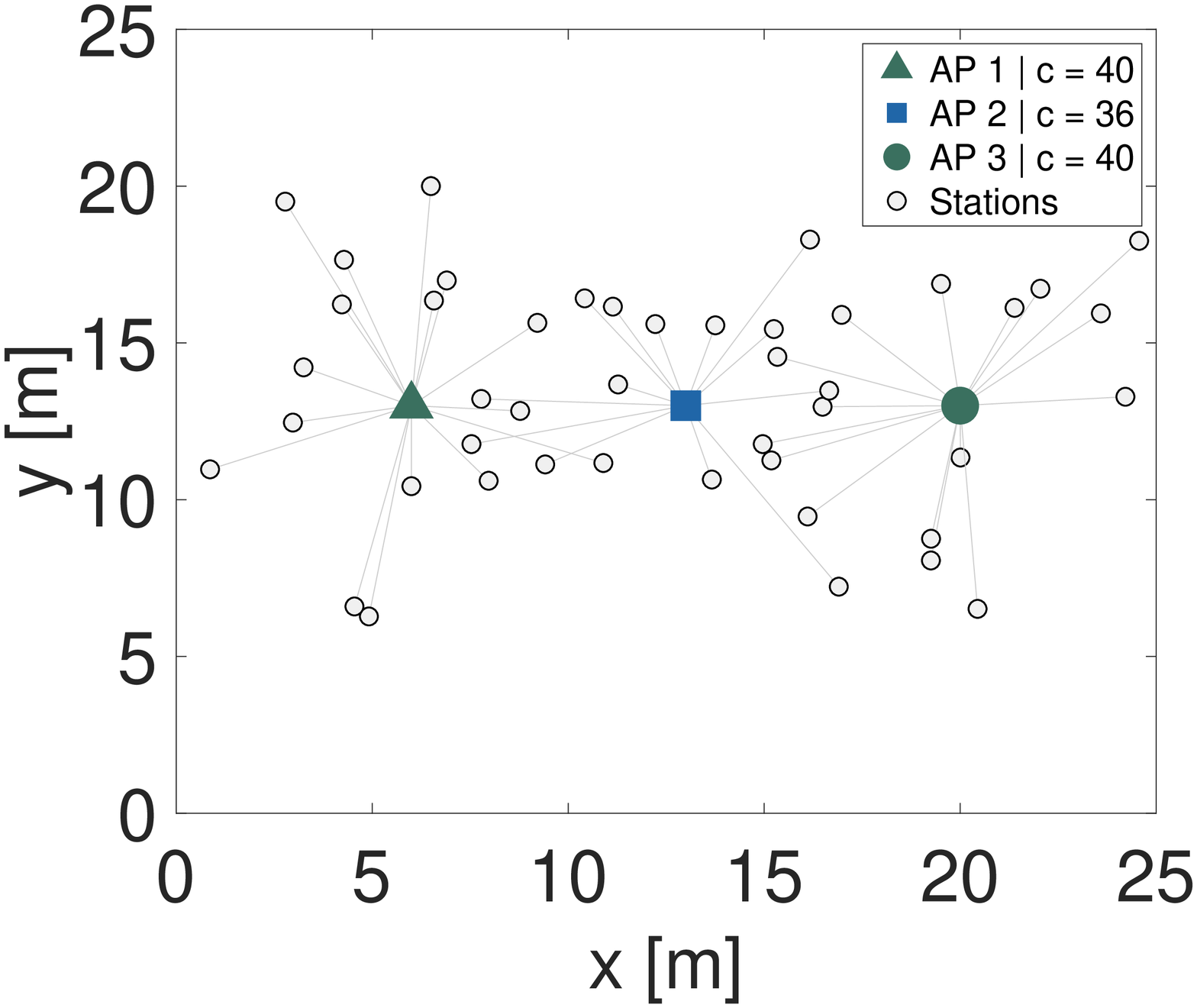}
         \caption{Final scheme}
         \label{fig:finalAssociationBoth}
     \end{subfigure}
     \hfill
     \begin{subfigure}[]{0.33\textwidth}
         \centering
         \includegraphics[ width=\textwidth]{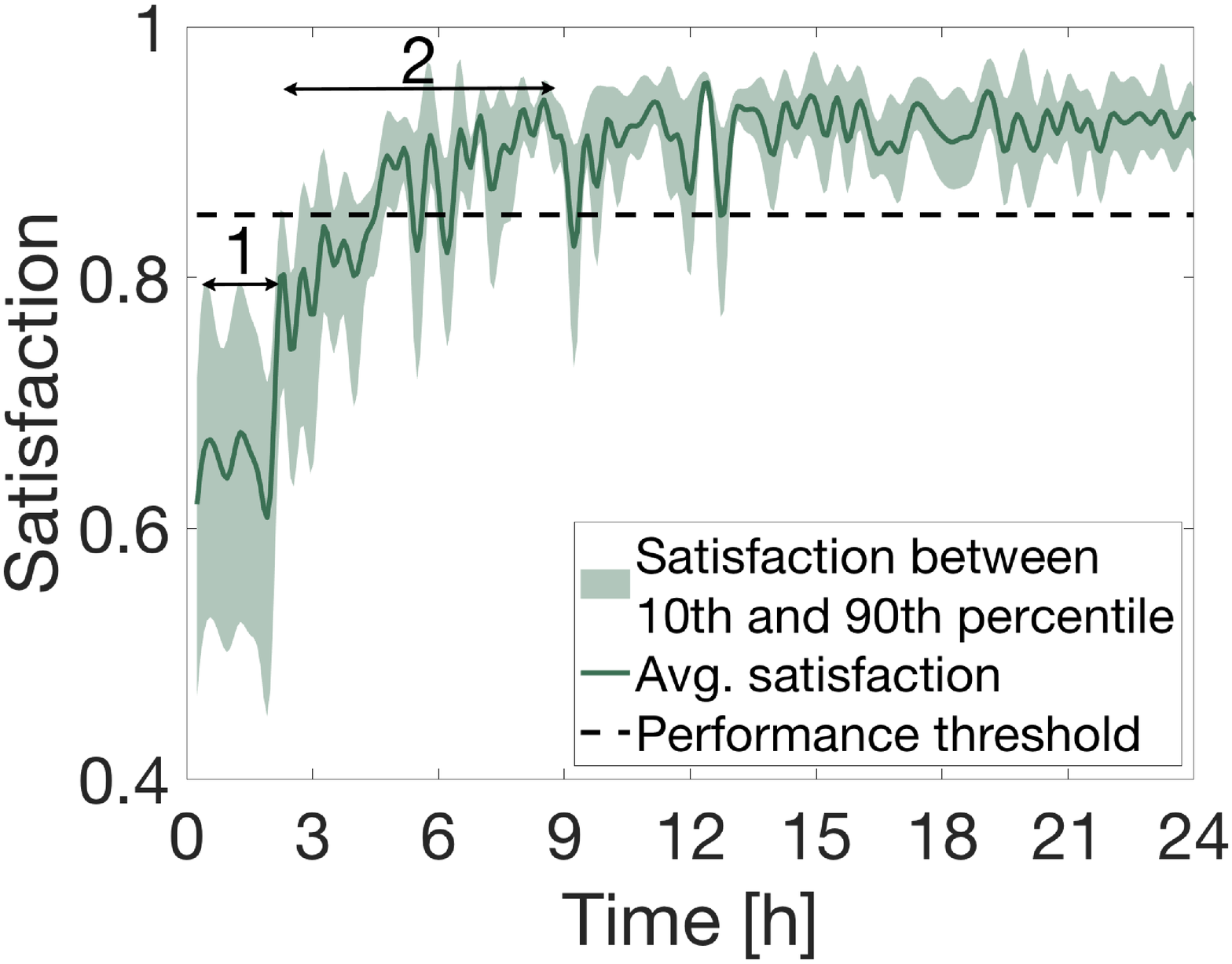}
         \caption{Satisfaction evolution}
         \label{fig:prcentileBoth}
     \end{subfigure}
     \caption{\small{Joint performance of DCA and DAPS}}
     \label{fig:DAPS_DCA}
\end{figure}

Figure~\ref{fig:finalAssociationBoth} shows the final result. We can observe that the concurrent execution of both DCA and DAPS have accomplished the expectations. We observe that APs have been reconfigured into a feasible solution. AP$_{1}$ and AP$_{3}$ have been allocated with the same channel, as they are out of their CCA range, whereas AP$_{2}$ has been allocated with a different one. Therefore, DCA agents overcome the flow starvation effect that, initially, AP$_{2}$ was suffering. On the other hand, the effect of the DAPS agents can be observed over the new distribution of the stations. However, the relevance of the DAPS in this scenario is significantly lower, than the DCA, due to the fact that only few stations have been reallocated. Figure~\ref{fig:prcentileBoth} shows the satisfaction evolution for all the stations, in which it is represented the mean value, as well as the 90th and 10th percentile of the measurements that are represented by the upper and lower bounds of the shaded area. From the figure, we observe that now the average satisfaction surpasses the performance threshold, which confirms that the efficiency of the solution remains mostly in the DCA mechanism to properly allocate orthogonal channels to overlapping BSSs.

Although some random bad performances around $t = 9$~h and $t = 12$~h can be observed in Figure~\ref{fig:prcentileBoth}, the network converges into a solution approximately at time $t = 5$~h. This fact is quite relevant as we can see that by applying both mechanisms we can get a considerable gain over the static approach (i.e., performance observed during the stage marked in point 1).

\begin{figure}[!t]
     \centering
     \begin{subfigure}[]{0.3\textwidth}
         \centering
         \includegraphics[ width=\textwidth]{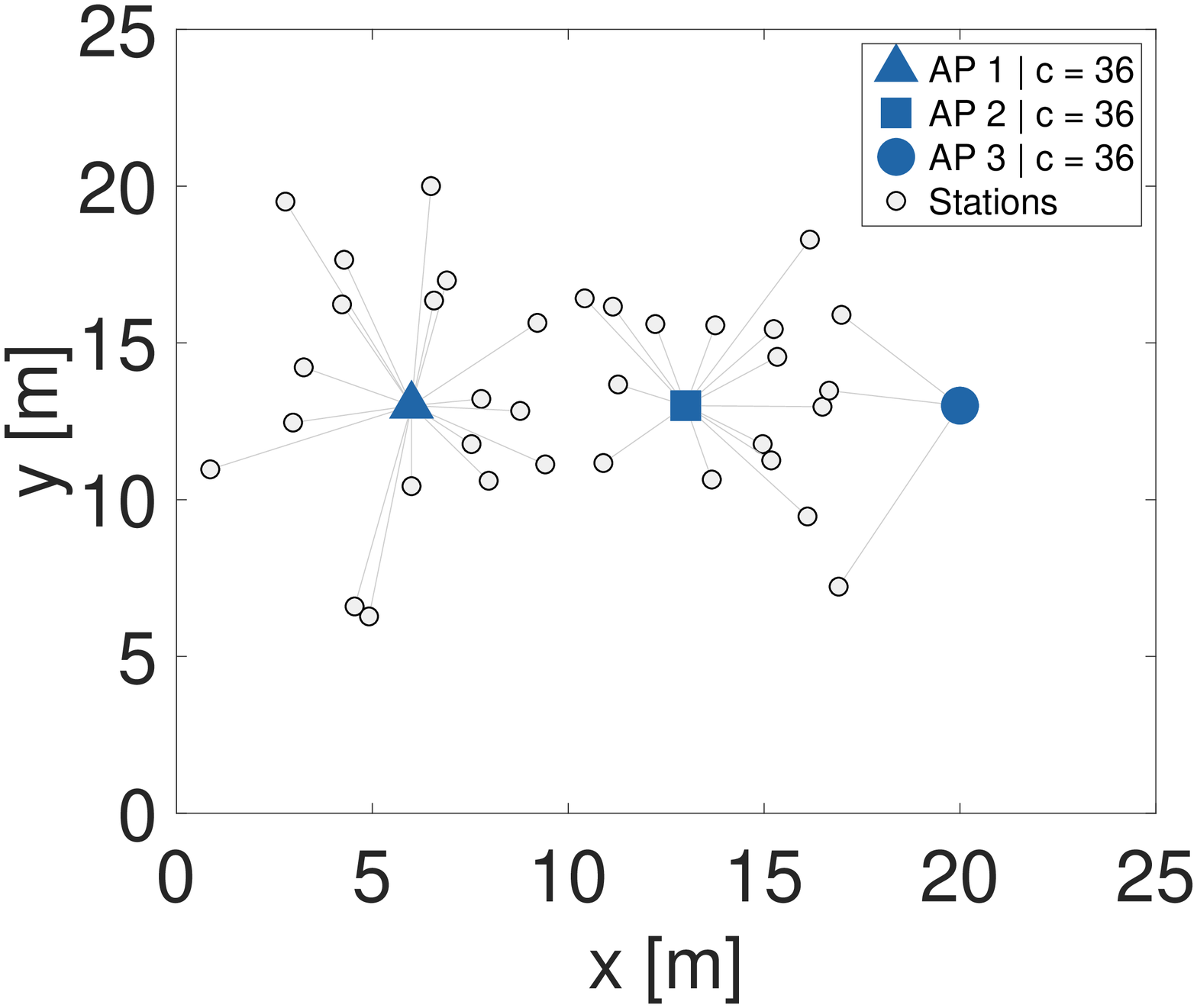}
         \caption{Initial scheme}
         \label{fig:initialAssociationBothUnbalanced}
     \end{subfigure}
     \hfill
     \begin{subfigure}[]{0.3\textwidth}
         \centering
         \includegraphics[ width=\textwidth]{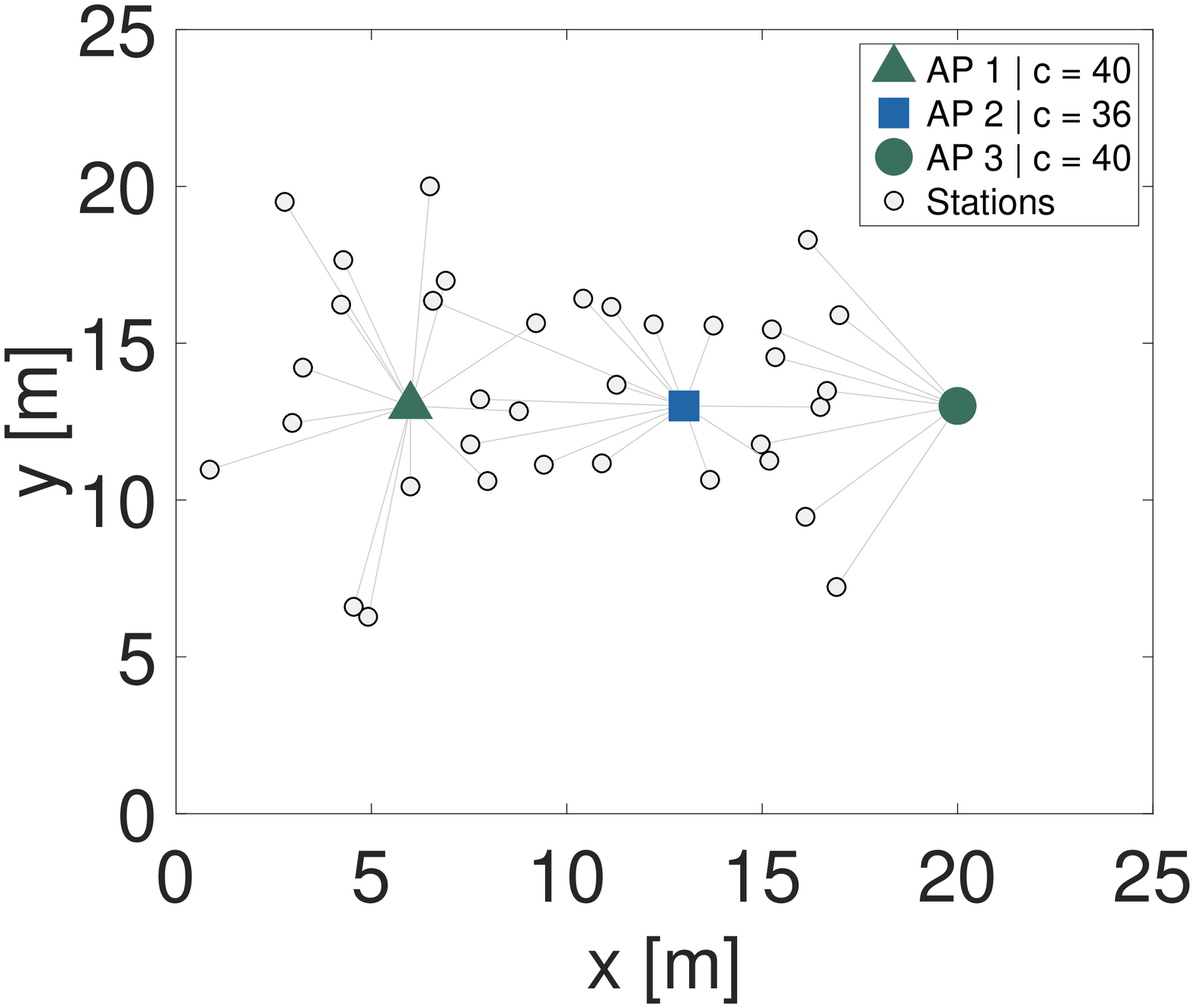}
         \caption{Final scheme}
         \label{fig:finalAssociationBothUnbalanced}
     \end{subfigure}
     \hfill
     \begin{subfigure}[]{0.33\textwidth}
         \centering
         \includegraphics[ width=\textwidth]{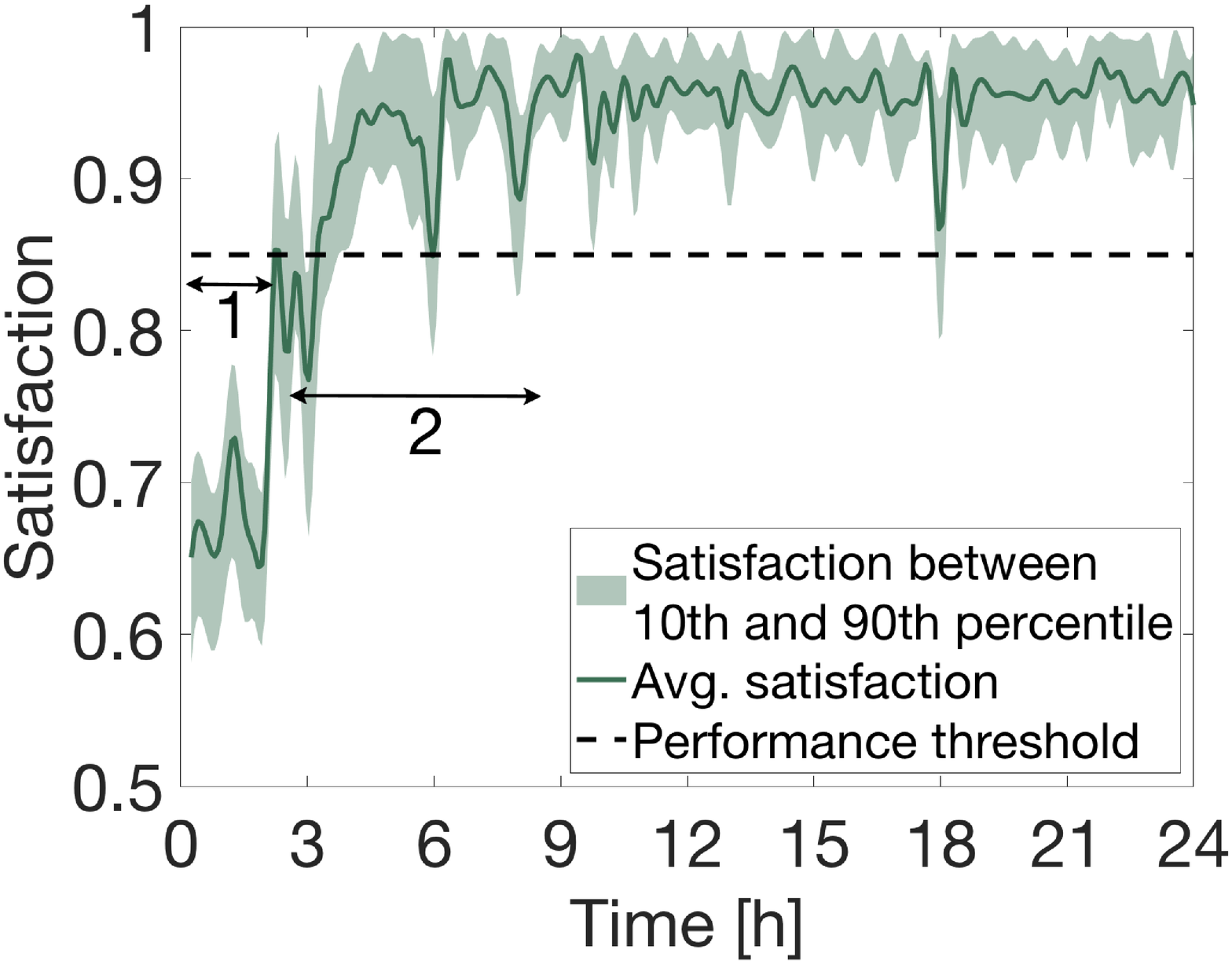}
         \caption{Satisfaction evolution}
         \label{fig:prcentileBothUnbalanced}
     \end{subfigure}
     \caption{\small{Joint performance of DCA and DAPS in an unbalanced scenario}}
     \label{fig:DAPS_DCA2}
\end{figure}

Finally, we assess the concurrent operation over an unbalanced scenario to properly evaluate the effects of the DAPS in such conditions. Results are shown in both Figure~\ref{fig:initialAssociationBothUnbalanced} and Figure~\ref{fig:prcentileBothUnbalanced}. Again, we observe that DCA agents have accomplished their tasks. However, now, we are more interested to evaluate the performance of the DAPS agents. From the figure, we observe that the unbalanced situation has been correctly mitigated, as the station distribution is fairer among the different APs. Specifically, we can observe that AP$_{3}$ has doubled the number of stations attached, which from AP$_{2}$ to AP$_{3}$. At the same time, such effect, has caused that stations from AP$_{1}$ migrate to AP$_{2}$ to further balance the number of stations associated to every AP. Here, it is presented a clear example in which actions from a set of players may condition decisions taken by others.

\subsection{Large random scenarios}

Once we have seen the implementation and benefits of applying the DCA and DAPS mechanisms in a controlled scenario, the objective now is to evaluate their capabilities in multiple scenarios randomly generated. Such evaluation will prove whether the DCA and DAPs mechanisms can help to improve the overall network performance. We aim to assess how DCA and DAPS react for different throughput requirements, number of APs and number of stations.

To proceed with the evaluation, we increment the 3D area considered previously to 30~x 30 x 2~m. We have simulated 100 different scenarios, either for the static approach (i.e., using always the initial configuration), and when DCA and DAPS are applied concurrently. All simulations represent 1 day of virtual time (i.e., 86400 seconds). It is worth mention that, at the start of each simulation, APs select their initial channel in a random fashion, and stations are attached to APs based in the SSF criteria. Besides, no channel restriction is placed, so APs can select channel numbers 36, 40 and 44. Both APs and stations are agent-enabled. The rest of the simulation parameters remain the same as the presented in Section~\ref{subsec:simul_params}.

In terms of performance metrics, we compute the average satisfaction achieved for all the stations during the simulation time for each one of the 100 scenarios. Then, we analyze the distribution obtained from the 100 average satisfaction values obtained. The representation of the obtained average results is done through box plots, in which the central mark indicates the median, and the bottom and top edges of the box indicate the 25th and 75th percentiles. Whiskers extend to the most extreme data points that are not considered outliers, whereas the outliers are plotted using the ‘o’ symbol.

\subsubsection{Throughput requirements}

\begin{figure*}[!b]
    \centering
    \begin{subfigure}[]{.32\textwidth}
         \centering
         \includegraphics[width =\textwidth]{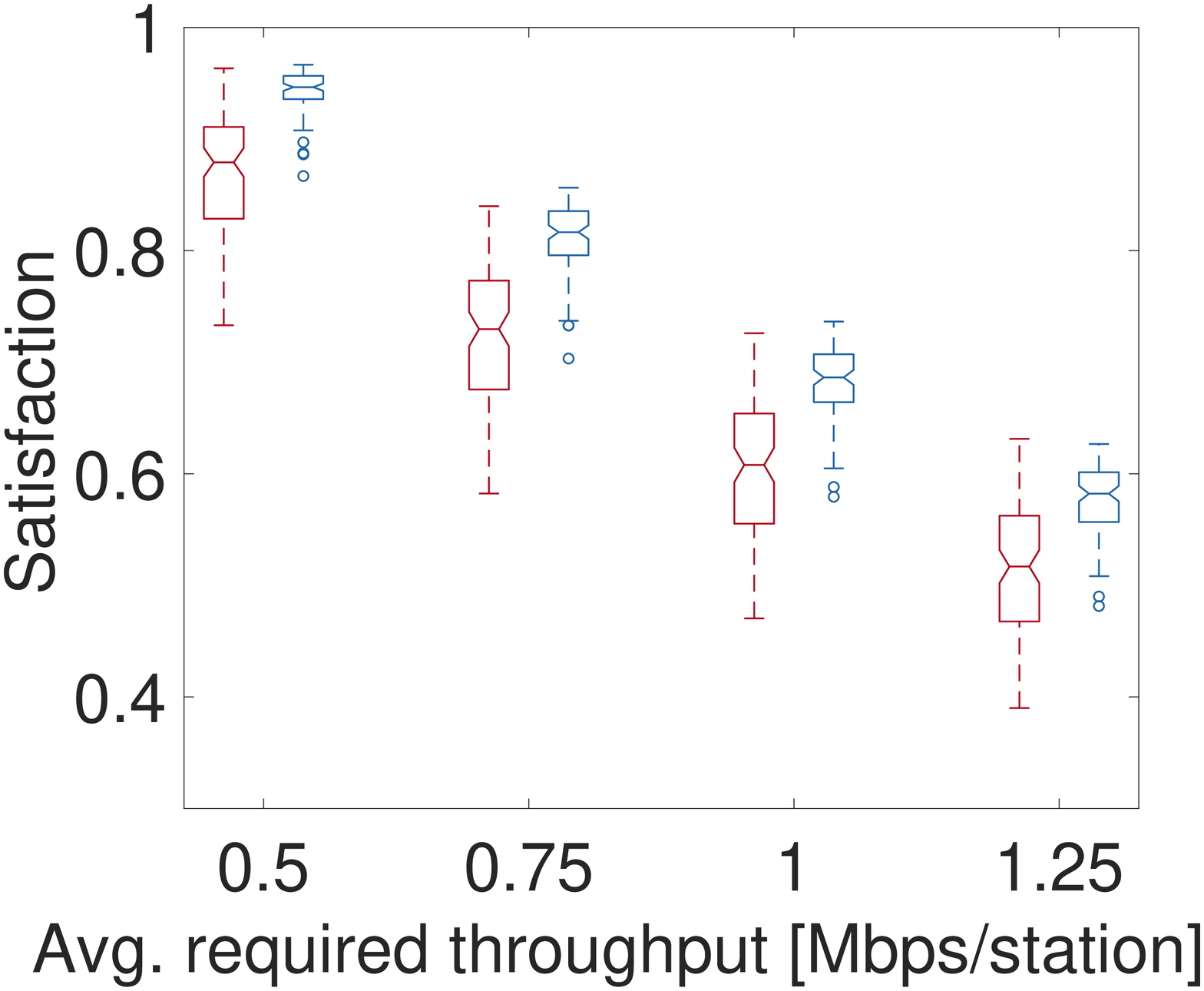}
         \caption{}
         \label{fig:Sat_AvgTh}
     \end{subfigure}
     \hfill
     \begin{subfigure}[]{.32\textwidth}
         \centering
         \includegraphics[width =\textwidth]{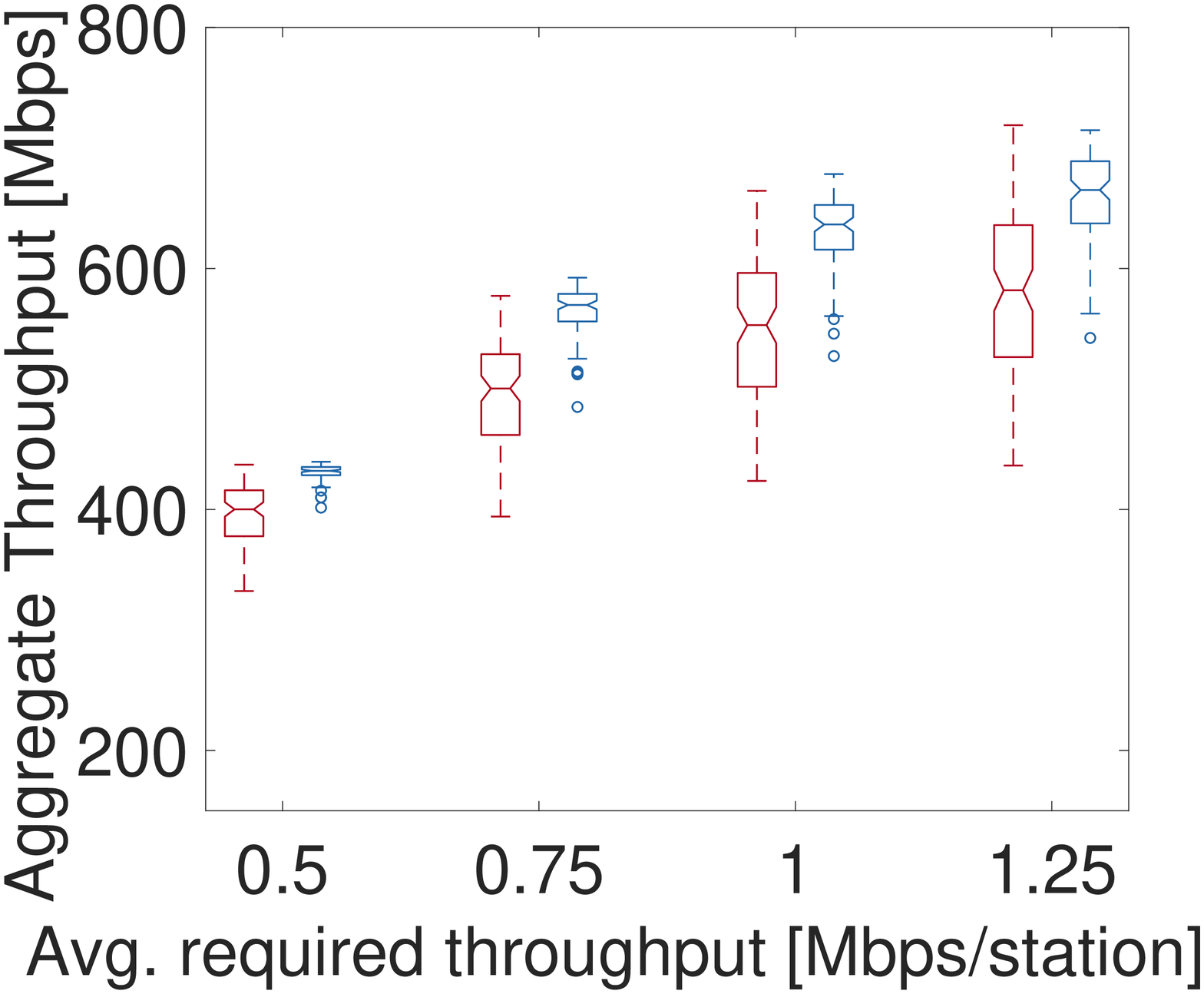}
         \caption{}
         \label{fig:Th_AvgTh}
    \end{subfigure}
    \hfill
    \begin{subfigure}[]{.32\textwidth}
         \centering
         \includegraphics[width=\textwidth]{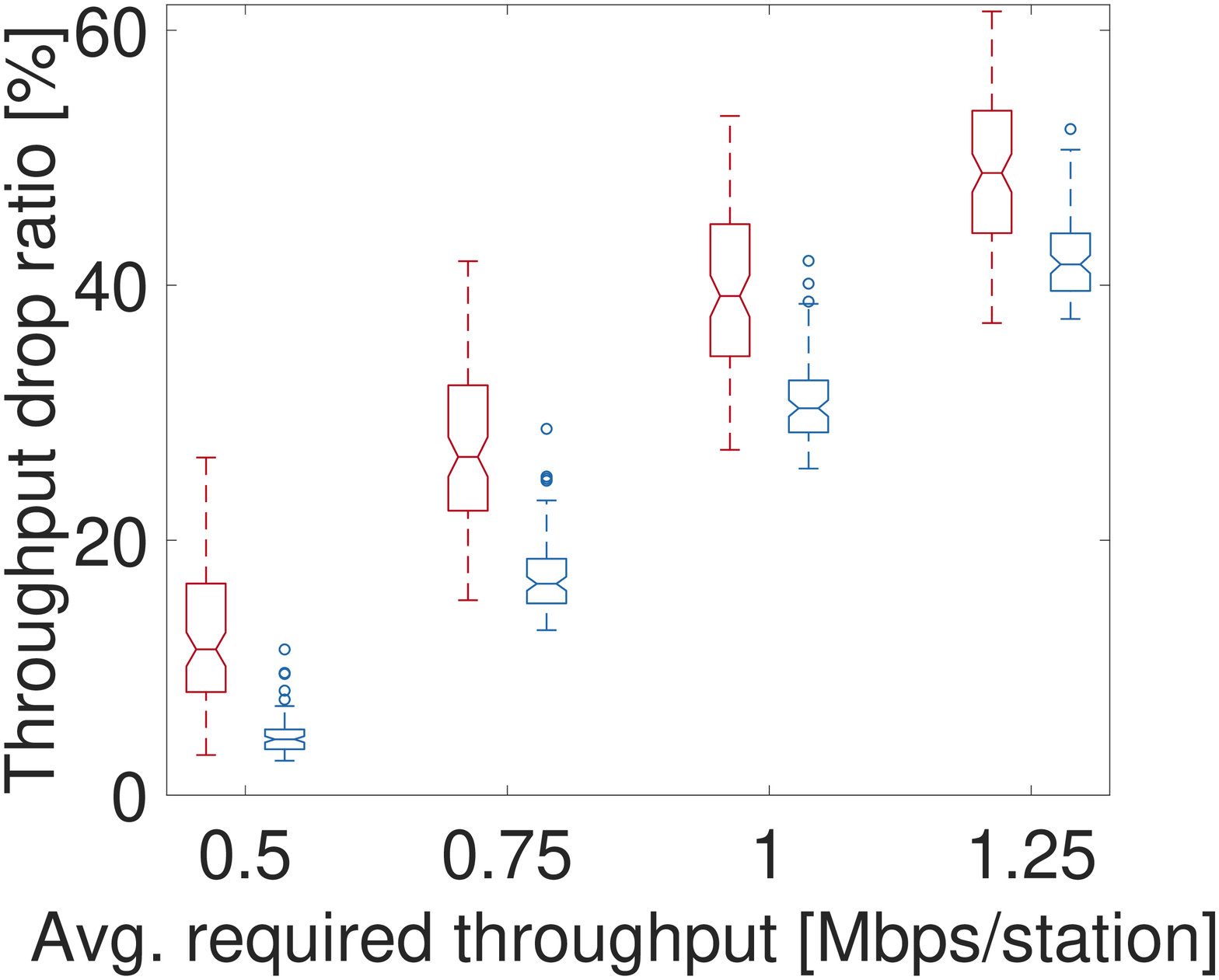}
         \caption{}
         \label{fig:Drop_AvgTh}
     \end{subfigure}
     \caption{\small{Increasing throughput requirements case. From right to left, we find the satisfaction, aggregate throughput and the throughput drop ratio, respectively. In red is shown the joint SSF and static channel allocation schemes, whereas in blue are represented the results of applying DCA and DAPS mechanisms.}}
     \label{fig:rand_VarBW}
\end{figure*}

We first study the performance of both techniques by considering different stations' downlink throughput requirements. To do so, we fix the network density to 15 APs and 225 stations, and consider four different throughput ranges: [1-3]~Mbps, [1-5]~Mbps, [1-7]~Mbps and [1-9]~Mbps. Figure~\ref{fig:rand_VarBW} shows the obtained results for the satisfaction, aggregate throughput and the throughput drop ratio (i.e., the percentage of traffic that cannot be served) metrics. Note that in Figure~\ref{fig:rand_VarBW}, the x-axis represents the average required throughput per station, i.e.,  
\begin{equation}\label{eqn:AvgBW}
\centering
\begin{split}
    \overline{B} = \E[B]\bigg(\frac{T_{\text{on}}}{T_{\text{on}} + T_{\text{off}}}\bigg), \nonumber
\end{split}
\end{equation}
where $\E[B]$ is the average value of the chosen range.

Comparing the adaptive MABs approach with the static one, we can observe that both schemes are very sensitive to the stations' required throughput. For instance, regarding the satisfaction achieved, we find that for both cases it becomes lower when the throughput range increases. However, taking the median value as a reference, we see that the DCA and DAPS perform 10\% better regardless the stations' required throughput. In addition, we can see a clear tendency in the throughput values when assessing the static and traditional approach. It can be appreciated that from the 0.75~Mbps/station to the 1.25~Mbps/station cases, the whiskers of the box encompass a larger range of values, clearly indicating a high variability in the data, as results are very sensitive to the specific topology of each scenario. Comparing the DCA and DAPs performance against the static scheme, we find that this dependency on the scenario's topology, and so the variance in the results, is highly reduced using the adaptive MABs approach. For instance, we observe a variance reduction of 60\% between the 25th and 75th percentile for the 1~Mbps/station case. In fact, we want to make special mention to the good performance of our mechanisms regardless the throughput demands, in which the MABs strategy outperform the static scheme reducing the variability, as all the obtained values are closer to the median value. Moreover, results show that in high traffic conditions, the only solution remains in densify the network with more APs, since the throughput drop ratio values indicate that almost half of the airtime required can not be allocated properly. 

\begin{figure*}[t!!]
    \centering
    \begin{subfigure}[]{.32\textwidth}
         \centering
         \includegraphics[width =\textwidth]{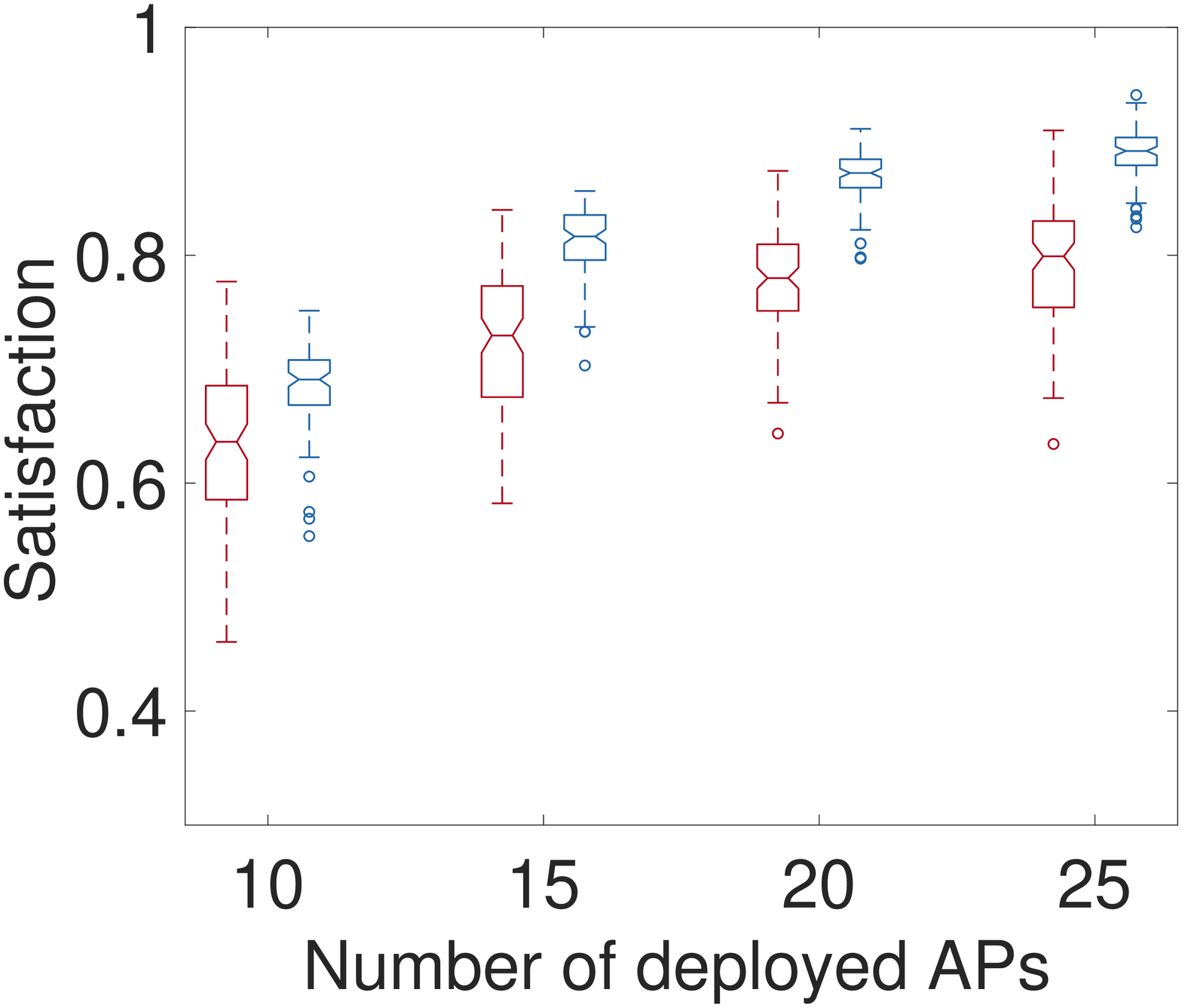}
         \caption{}
         \label{fig:Sat_ApDens}
     \end{subfigure}
     \hfill
     \begin{subfigure}[]{.32\textwidth}
         \centering
         \includegraphics[width =\textwidth]{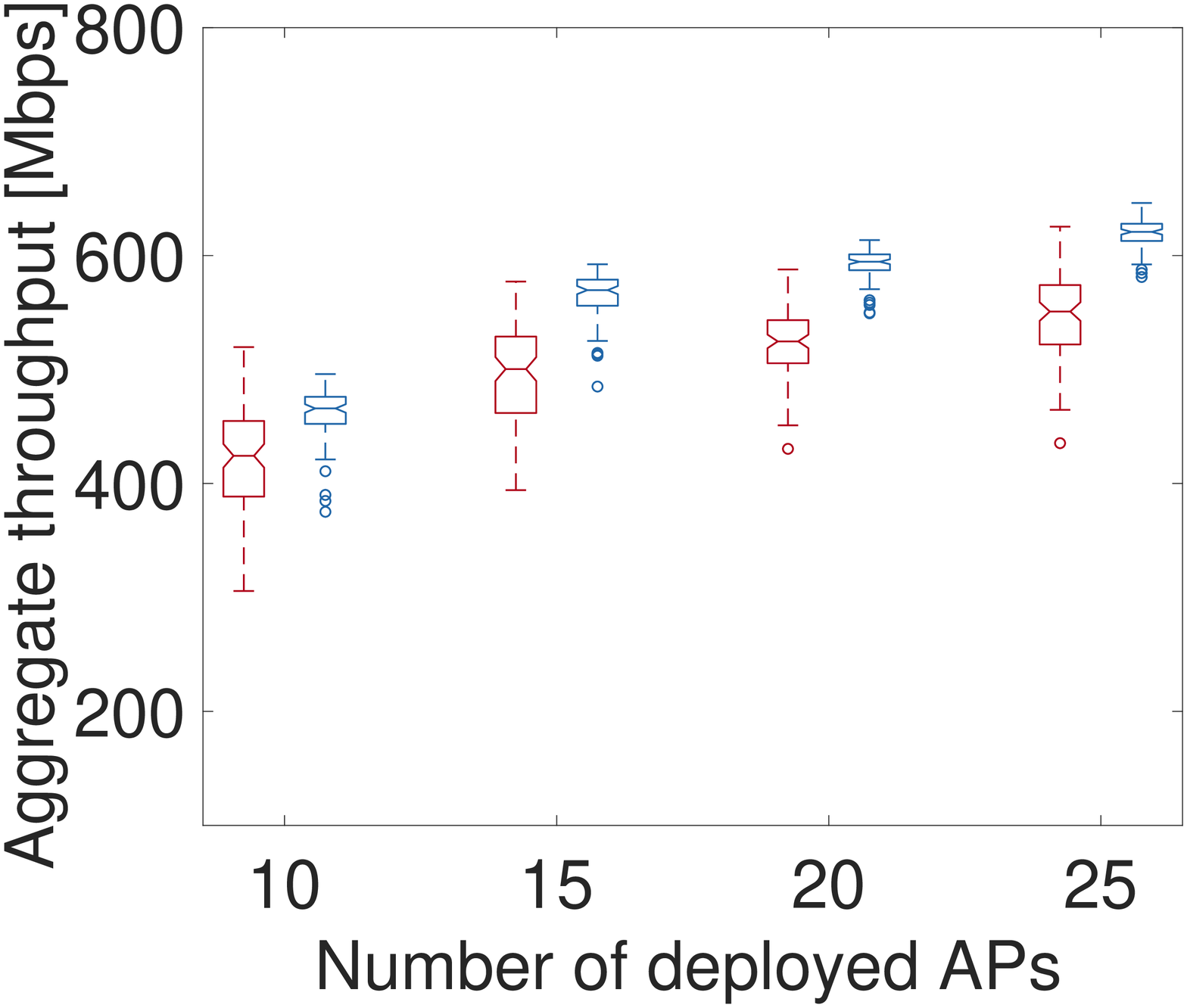}
         \caption{}
         \label{fig:Th_ApDens}
    \end{subfigure}
    \hfill
    \begin{subfigure}[]{.32\textwidth}
         \centering
         \includegraphics[width=\textwidth]{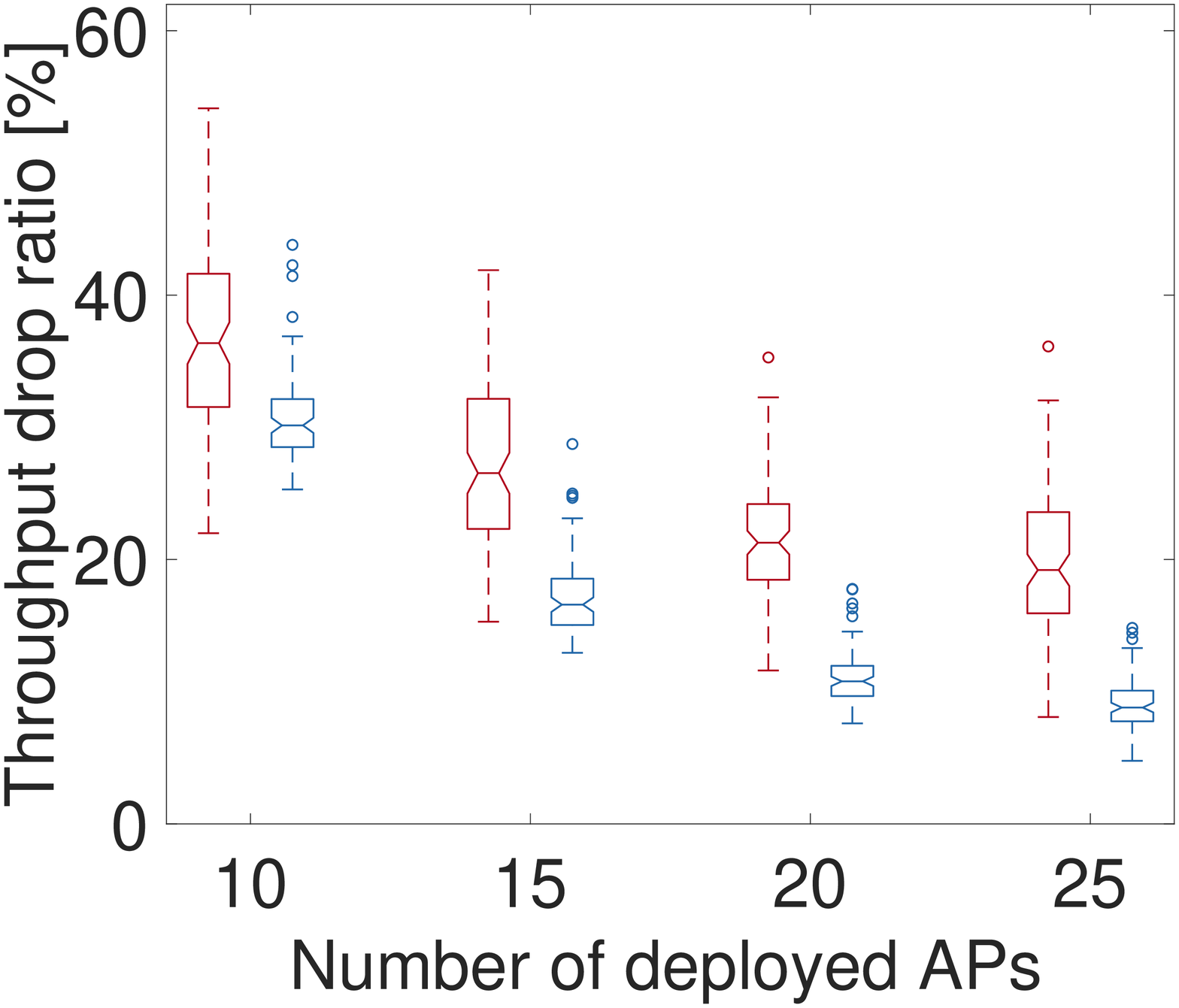}
         \caption{}
         \label{fig:Drop_ApDens}
    \end{subfigure}
    \caption{\small{AP densification case. From right to left, we find the satisfaction, aggregate throughput and the throughput drop ratio, respectively. In red is shown the joint SSF and static channel allocation schemes, whereas in blue are represented the results of applying DCA and DAPS mechanisms.}}
    \label{fig:rand_APDense}
\end{figure*}

\subsubsection{APs densification}

As we have seen, high traffic demands can prevent the network to deliver the required service. In this section, we investigate if adding more APs to the network may contribute to improve the network performance. To do so, we keep the same number of stations (225 stations), while traffic demands from stations are pick from the range [1-5]~Mbps. Figure~\ref{fig:rand_APDense} show the obtained results. We can observe how adding more APs improve the network performance, as the uptrend evolution of the satisfaction and throughput values indicate. Also, notice in Figure~\ref{fig:Sat_ApDens} that the gains of adding new APs when applying the DCA and DAPS strategies get lower at each step, in special from the 20 APs case to the 25 APs, which is about 1\%. This effect is an indicator that the network will barely improve even though more APs will be placed. In order to overcome this effect, APs should be strategically placed in order to be detected above the RSSI$_{\rm{th}}$, so the DAPS mechanism can be triggered, and an effective user re-association produced. Although network densification can be a good solution to tackle network congestion, Figure~\ref{fig:Drop_ApDens} shows that it does not improve the performance by itself. Analyzing the throughput drop ratio in the case of 25 APs, we notice that the static approach still performs badly since the 75th percentile of the measurements surpasses a value of 20\%. This effect is associated to the topology dependency mentioned before, as overlapping APs with the same channel configuration are more likely to happen. In this type of scenarios, the benefits of using the DCA mechanism are very relevant, since the network is able to better manage the spectrum resources, achieving a better performance, and very low variance between different scenarios. 

\begin{figure*}[b]
    \centering
    \begin{subfigure}[]{.32\textwidth}
         \centering
         \includegraphics[width =\textwidth]{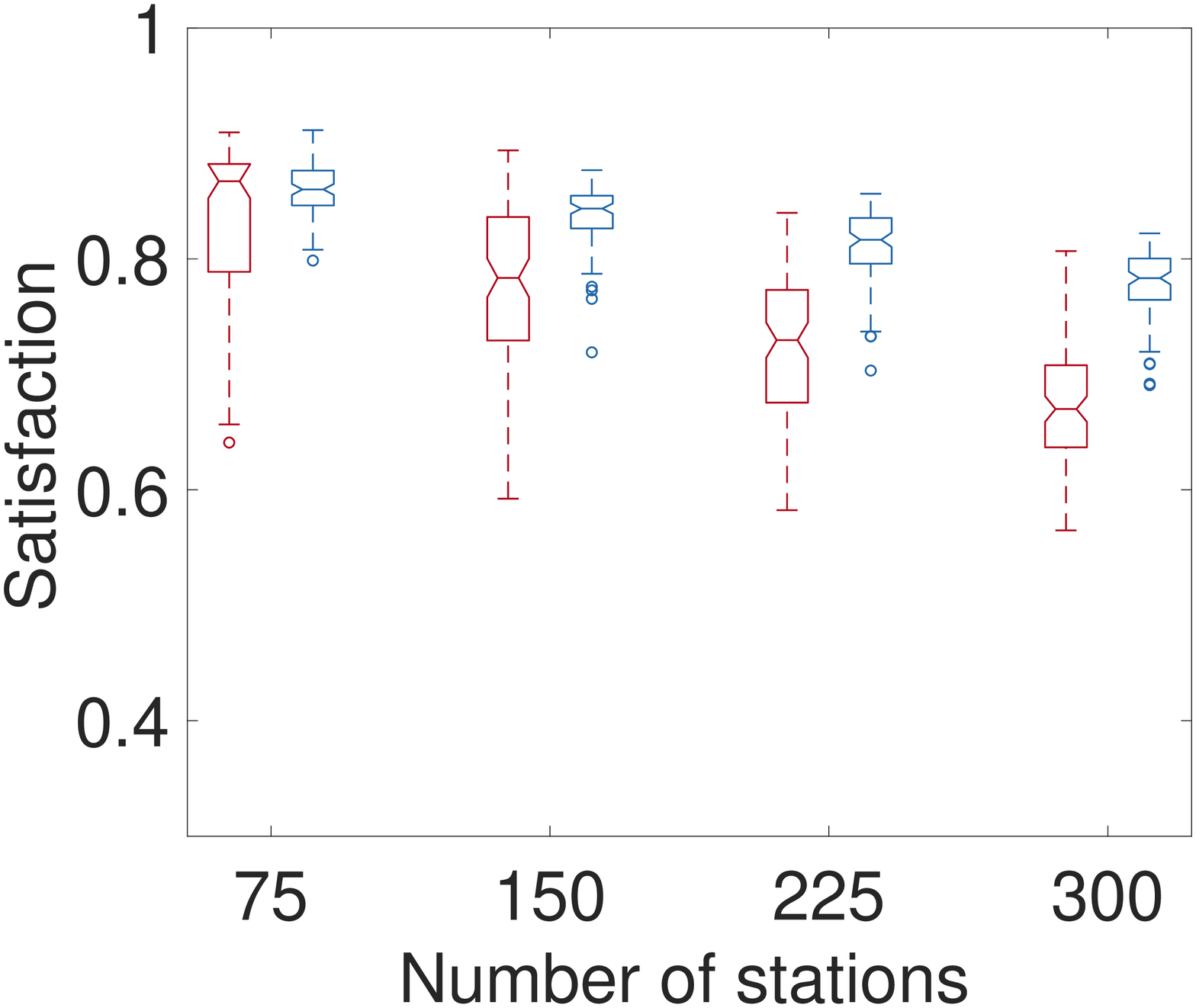}
         \caption{}
         \label{fig:Sat_NetDens}
     \end{subfigure}
     \hfill
     \begin{subfigure}[]{.32\textwidth}
         \centering
         \includegraphics[width =\textwidth]{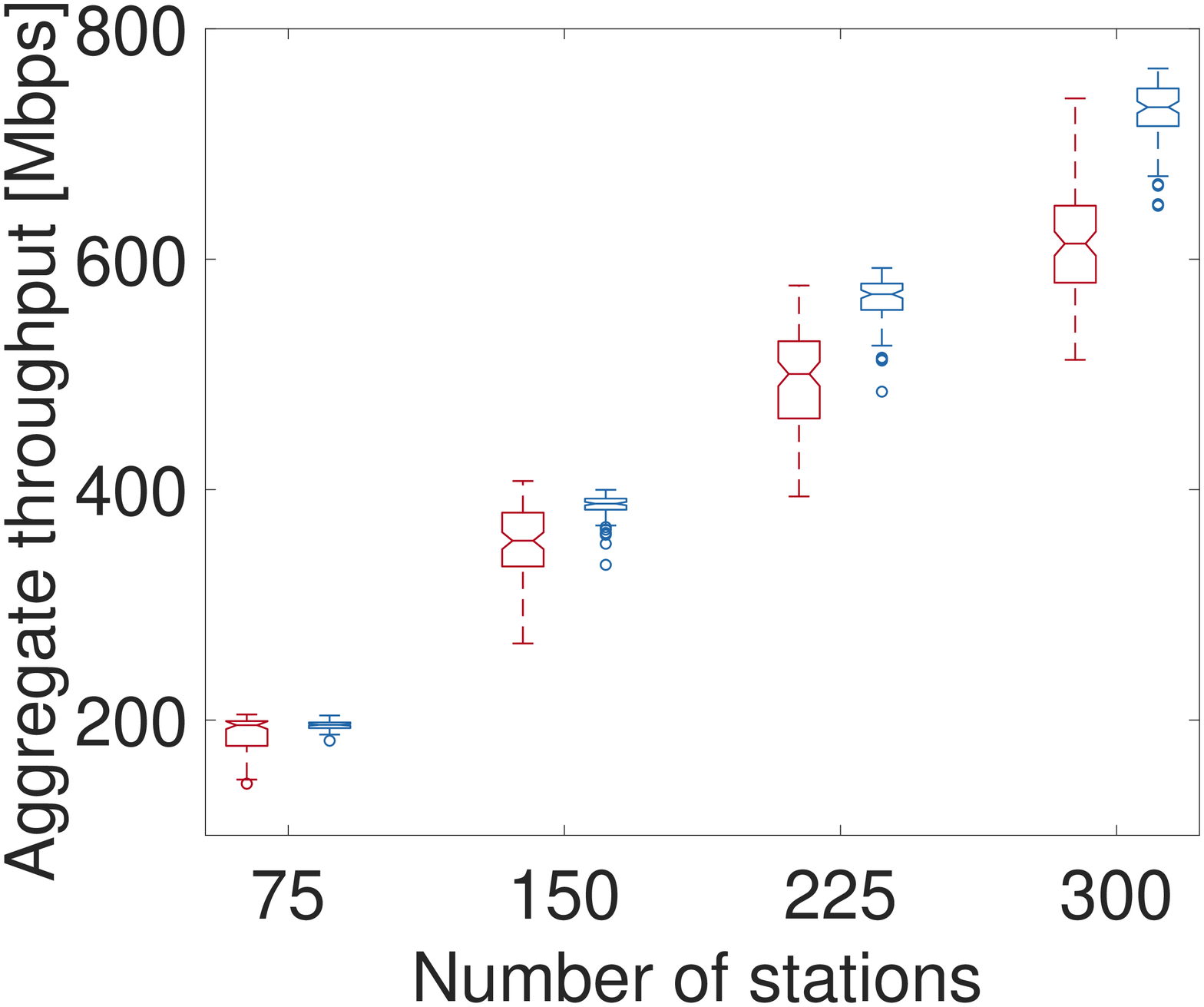}
         \caption{}
         \label{fig:Th_NetDens}
    \end{subfigure}
    \hfill
    \begin{subfigure}[]{.32\textwidth}
         \centering
         \includegraphics[width=\textwidth]{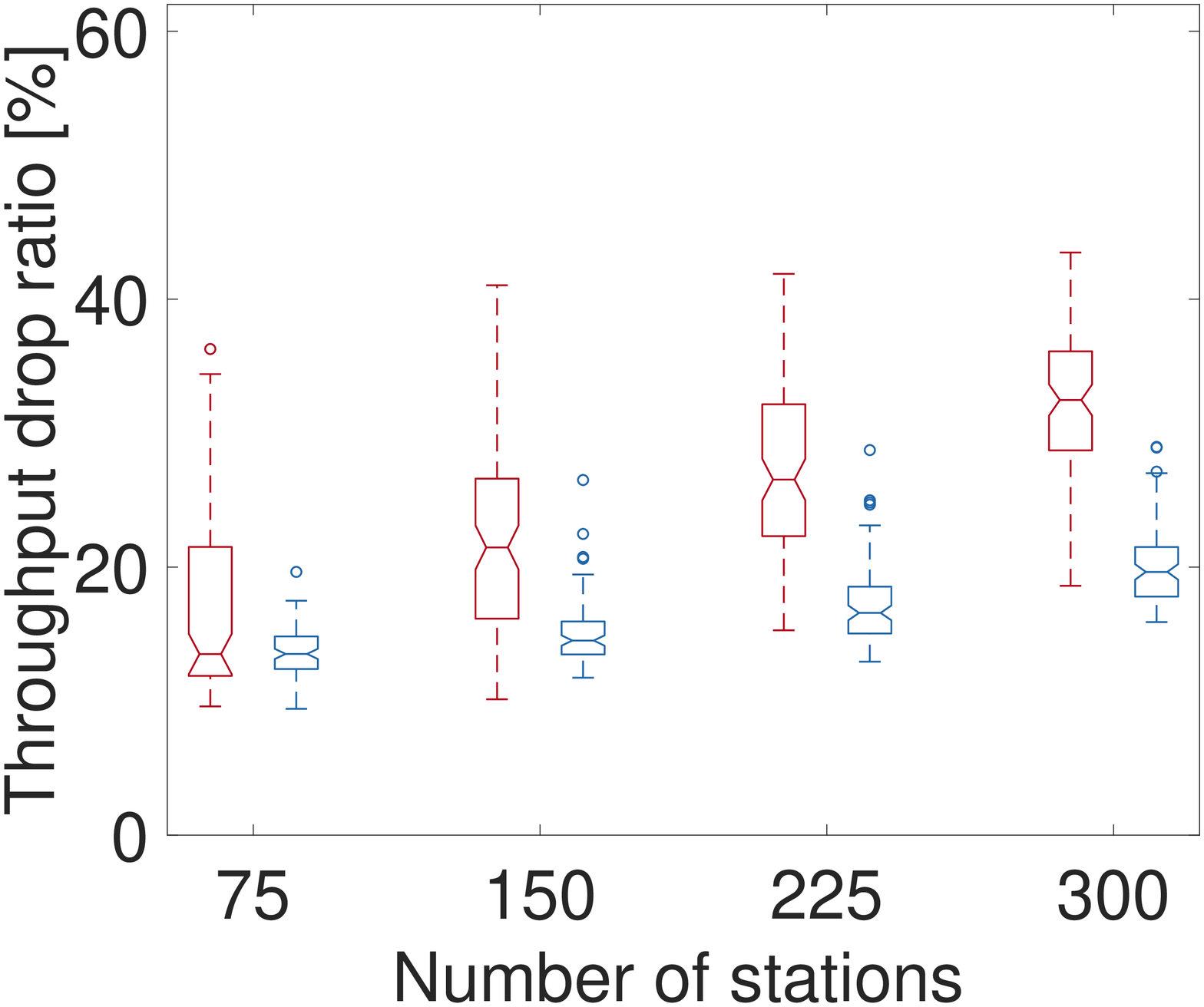}
         \caption{}
         \label{fig:Drop_NetDens}
    \end{subfigure}
     \caption{\small{Variable AP and user density case. From right to left, we find the satisfaction, aggregate throughput and the throughput drop ratio, respectively. In red is shown the joint SSF and static channel allocation schemes, whereas in blue are represented the results of applying DCA and DAPS mechanisms.}}
     \label{fig:rand_NetDense}
\end{figure*}

\subsubsection{Station density}

In the following, we study the performance of the DAC and DAPS with different number of stations. We have evaluated scenarios with 75, 150, 225 and 300 stations, keeping the station/AP ratio to 15. Thus, for each density we will have 5, 10, 15 and 20 APs, respectively. Traffic demands per station remain in the range [1-5]~Mbps. Figure~\ref{fig:rand_NetDense} shows the results obtained. As it can be seen, the learning approach remains as the best solution in terms of performance. Despite the downtrend of the boxes, the difference between the static and the adaptive MABs approach gets higher, as the network gets denser. For instance, for 150 stations the gain is around 7\%, whereas for the 225 and 300 stations the gain gets up to 11\% and 16\%, respectively. This effect shows us that the adaptive MABs approach is capable to support a larger number of users before downgrading significantly the performance. Again, the low variance results observed in all figures show that in a wide diversity of scenarios the output values are very constant in front of the static approach, which presents a high variability even in scenarios with few APs as indicated by the larger whiskers range. Finally, it can be observed that, when applying the DCA and DAPS, agents are able to successfully learn even in complex and challenging scenarios, such as the case of having 20 APs and 300 stations.

\begin{figure}[t]
\centering
    \begin{minipage}[b]{0.45\linewidth}
        \includegraphics[width=\textwidth]{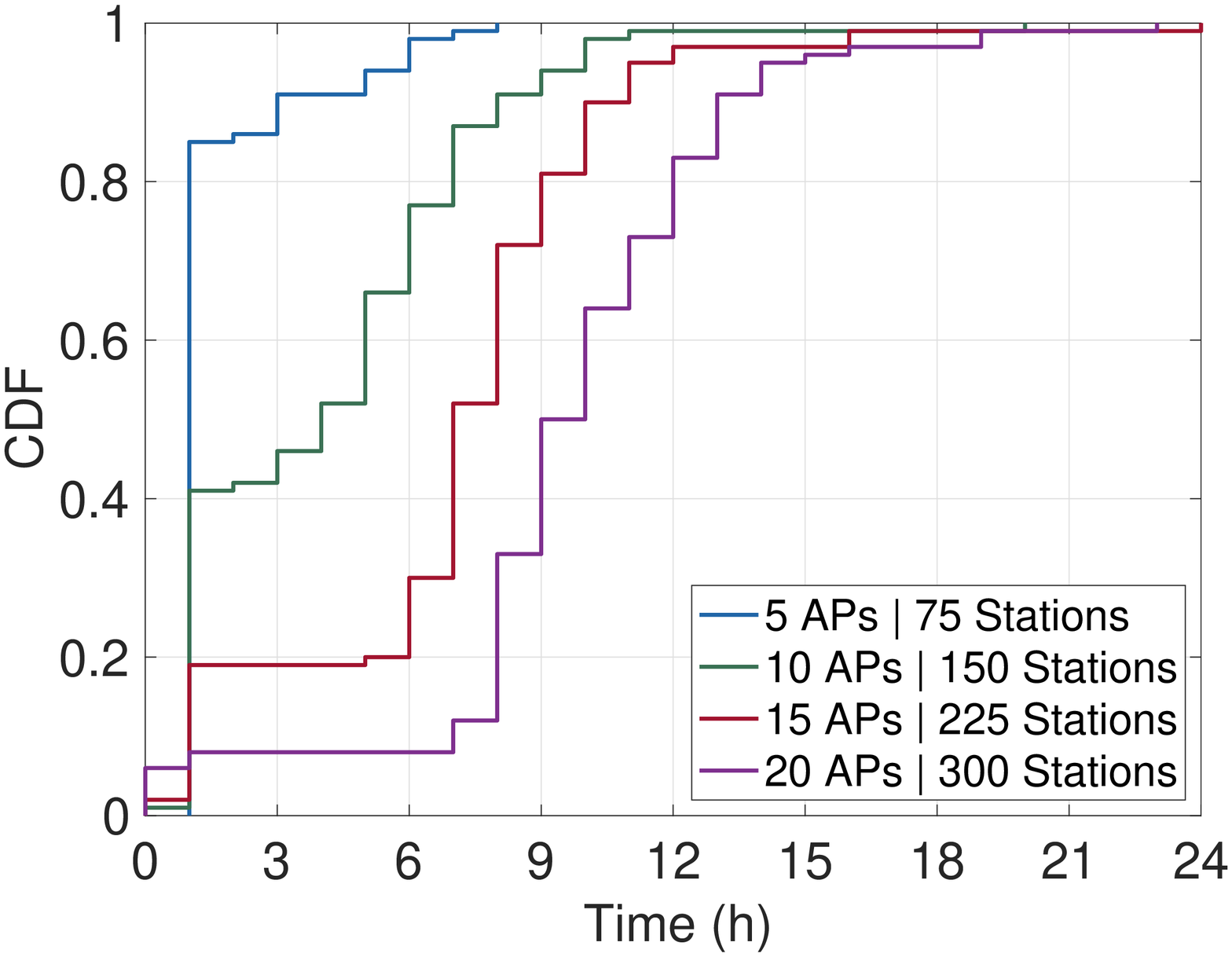}
        \caption{Empirical CDF for different network station density cases}
        \label{fig:CDF_Conv}
    \end{minipage}
    \hfill
    \begin{minipage}[b]{0.45\linewidth}
        \includegraphics[width=\textwidth]{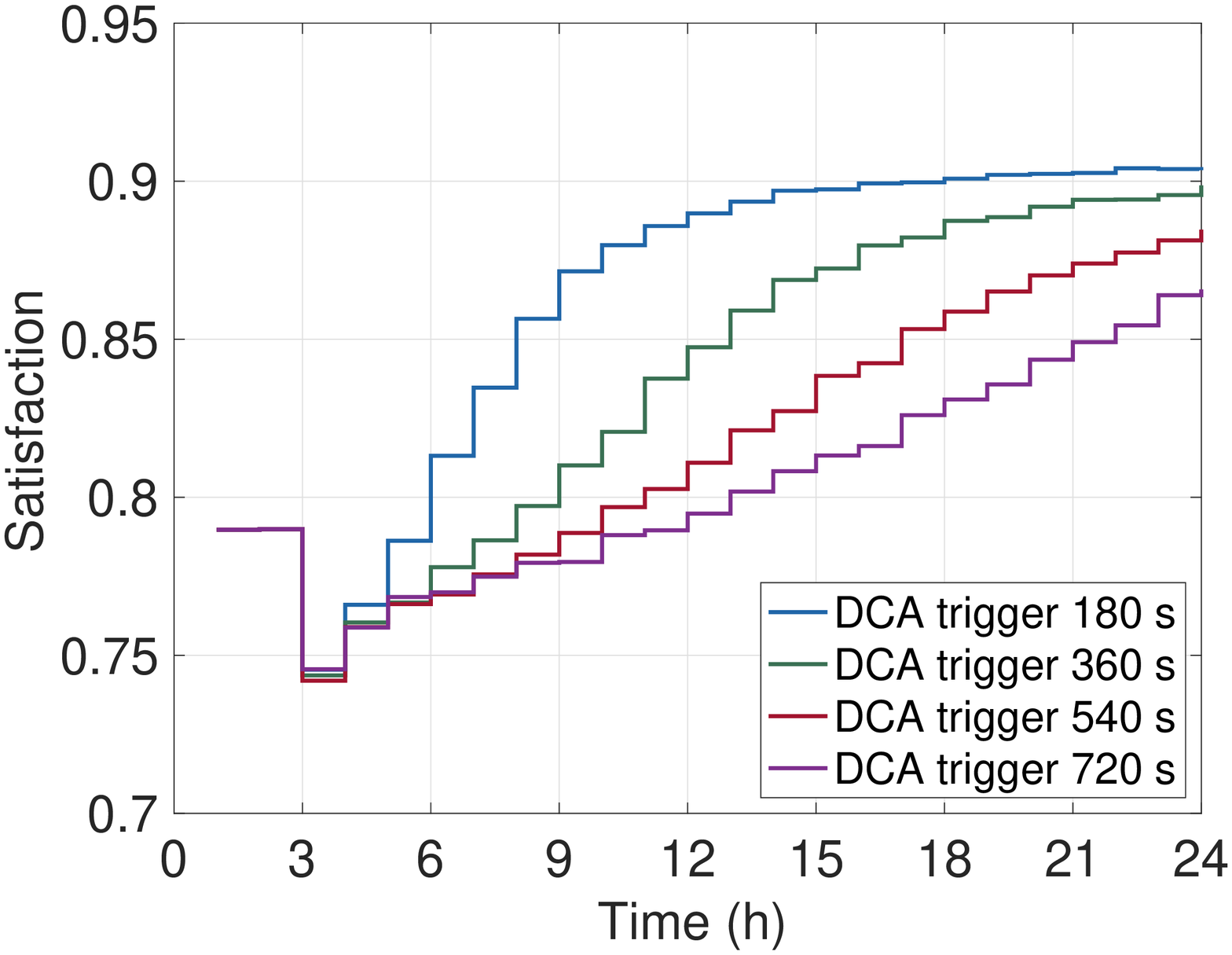}
        \caption{Evolution of the satisfaction for different DCA timer values}
        \label{fig:MABVar}
    \end{minipage}
\end{figure}

Finally, Figure~\ref{fig:CDF_Conv} shows the empirical cumulative distribution function (CDF) of the convergence time for the different considered scenarios. We define the convergence time in each individual scenario as the instant of time in which the median value of the satisfaction gets above the performance threshold, P$_{th}$. As it is shown, the results display that there exist a temporal dependence between convergence and network density. Thus, for 75, 150, 225 and 300 stations, considering the time at which the 80 \% of the individual scenarios have converged, we find out that their convergence times are 1, 7, 9, and 13 hours, respectively. Therefore, the denser the network, the more time the network will need to reach a solution. As we consider scenarios in which mobility is not significant, such as office buildings, where stations remain quiet during long periods of time, we consider that the obtained times are low enough to be acceptable in practice. 

Nevertheless, it is unrealistic to think that a network will remain static for very long periods of time (i.e., more than 10h). In order to overcome such a requirement, by identifying the periods of maximum traffic load (i.e., the busy hour), and enabling the learning agents only during these periods of time through multiple days, we could expect to obtain similar results as shown here.

\subsubsection{Agent Action-Selection timers}\label{sec:Interaction}

We finally asses the implications of varying both T$_{\text{DCA}}$ and T$_{\text{DAPS}}$ timers in the network response, in order to study the behavior of the system when the agents are executed more and less frequently. First, we tackle the AP selection case alone, fixing T$_{\text{DCA}}$ to 180~s, and testing different values for T$_{\text{DAPS}}$. Again, simulations comprise 100 random scenarios, in which the network consists of 15 APs and 225 stations. The traffic load of each flow are kept in the range of [1-5]~Mbps. As in previous sections, the simulation time is set to one day (84600~s). The agents are activated at $t=3~h$. From the results obtained, although not shown here, we observe that the DAPS timer does not have a significant impact on the performance of the stations as we only experience differences of 2 \% between different timer configurations. 

In the same conditions as in previous case, now, we vary the T$_{\text{DCA}}$, while keeping T$_{\text{DAPS}}$ to 180~s. In this case, the results are shown in Figure~\ref{fig:MABVar}, where the average satisfaction evolution is presented for different T$_{\text{DAPS}}$ values. We tested values of 360~s, 540~s and 720~s, which correspond to 2, 3 and 4 times the default T$_{\text{DAPS}}$ value of 180~s. We observe that for small values of T$_{\text{DAPS}}$ convergence is reached in much less time that if timers are set to higher values. Therefore, we observe that there is no gain by increasing the value of the timers. Then, we suggest to set those timers to the minimum possible value given the reconfiguration overheads are kept negligible.

% ---------------------
% ---------------------

\subsection{Dynamic environment}\label{subsec:nonstationary_scenario}

\subsubsection{A sudden channel change}\label{subsubsec:nonStationary}

So far, network conditions remained static, meaning that no variation or anomaly was introduced in the network. However, as a final use case, we want to analyze the behavior of the learning agents in a changing environment. For the sake of practicality, we evaluated this scenario in the controlled deployment of the toy scenario used in Section~\ref{subsubec:toy}. In this case, we consider that the whole set of stations are agent-enabled, whereas AP$_{1}$ and AP$_{3}$ are agent-enabled but AP$_{2}$ is not. As in the toy scenario, we use only channels 36 and 40. 

\begin{figure*}[t]
\centering
    \begin{subfigure}[]{.35\textwidth}
         \centering
         \includegraphics[width =\textwidth]{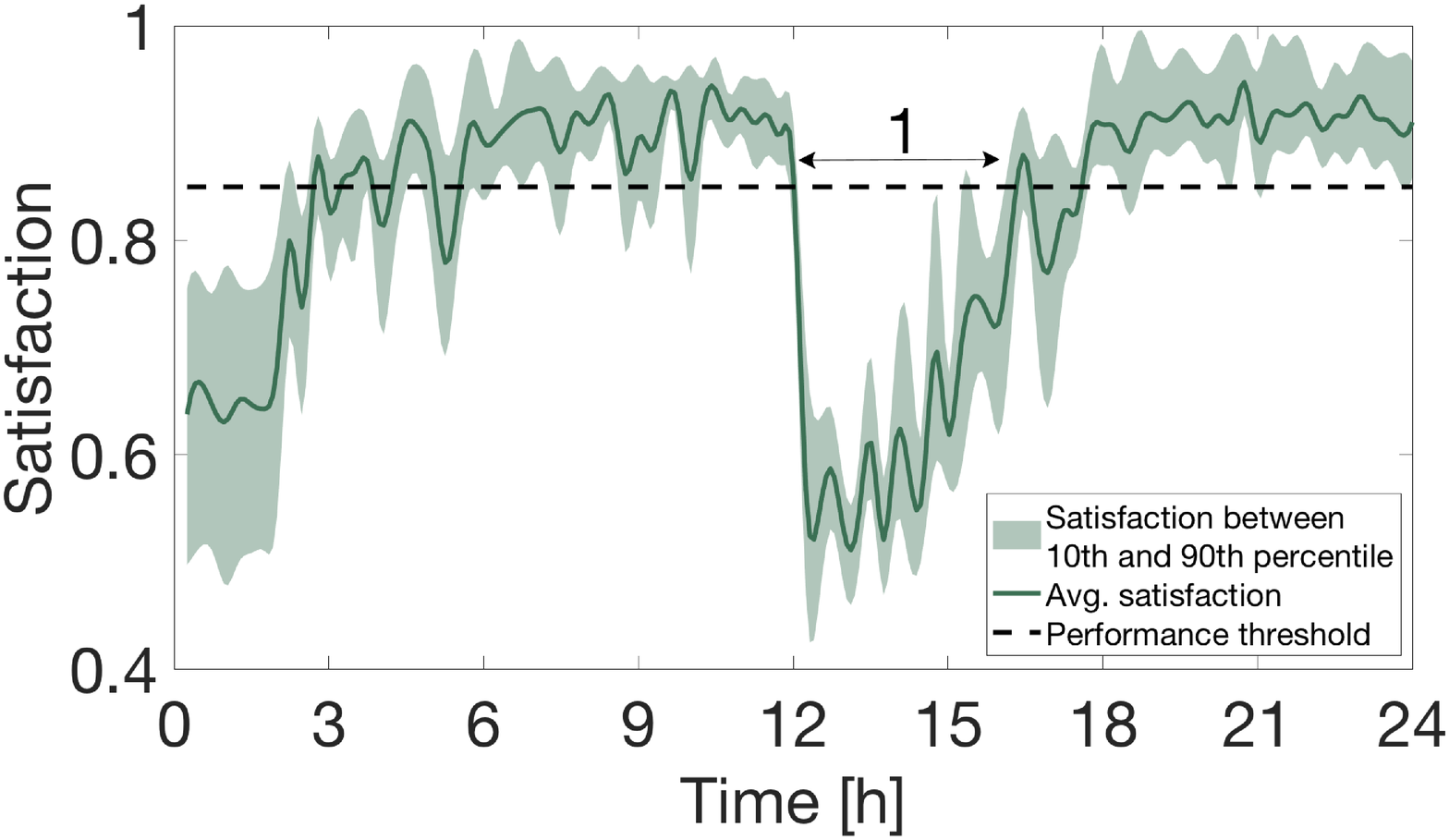}
         \caption{Stations satisfaction}
         \label{fig:SatNon}
     \end{subfigure}
     \hfill
     \begin{subfigure}[]{.27\textwidth}
         \centering
         \includegraphics[width =\textwidth]{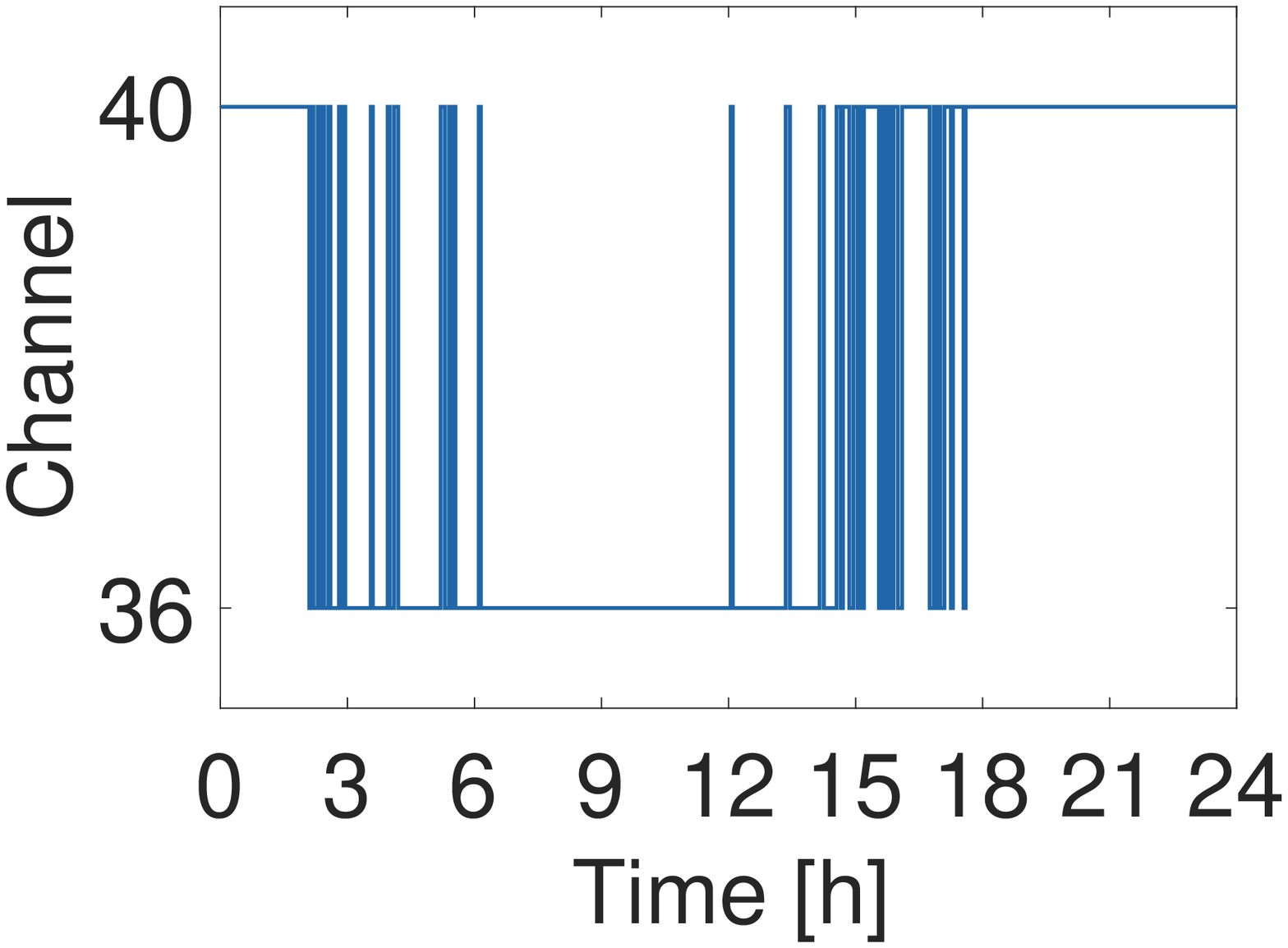}
         \caption{AP$_{3}$ channel selection}
         \label{fig:ActionAP}
    \end{subfigure}
    \hfill
    \begin{subfigure}[]{.27\textwidth}
         \centering
         \includegraphics[width=\textwidth]{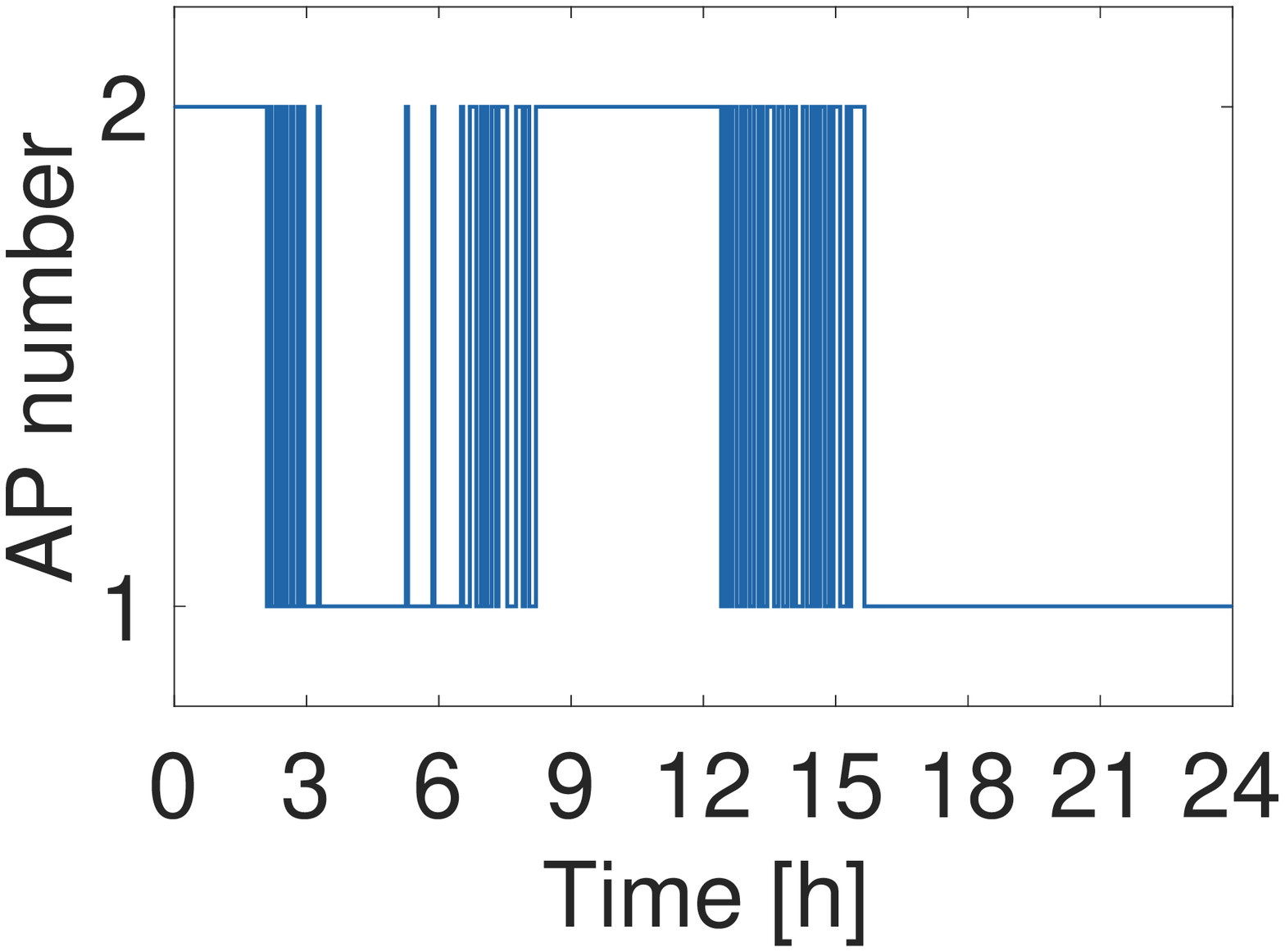}
         \caption{STA$_{2}$ AP selection}
         \label{fig:ActionSTA}
     \end{subfigure}
     \hfill
     \begin{subfigure}[]{.4\textwidth}
         \centering
         \includegraphics[width =\textwidth]{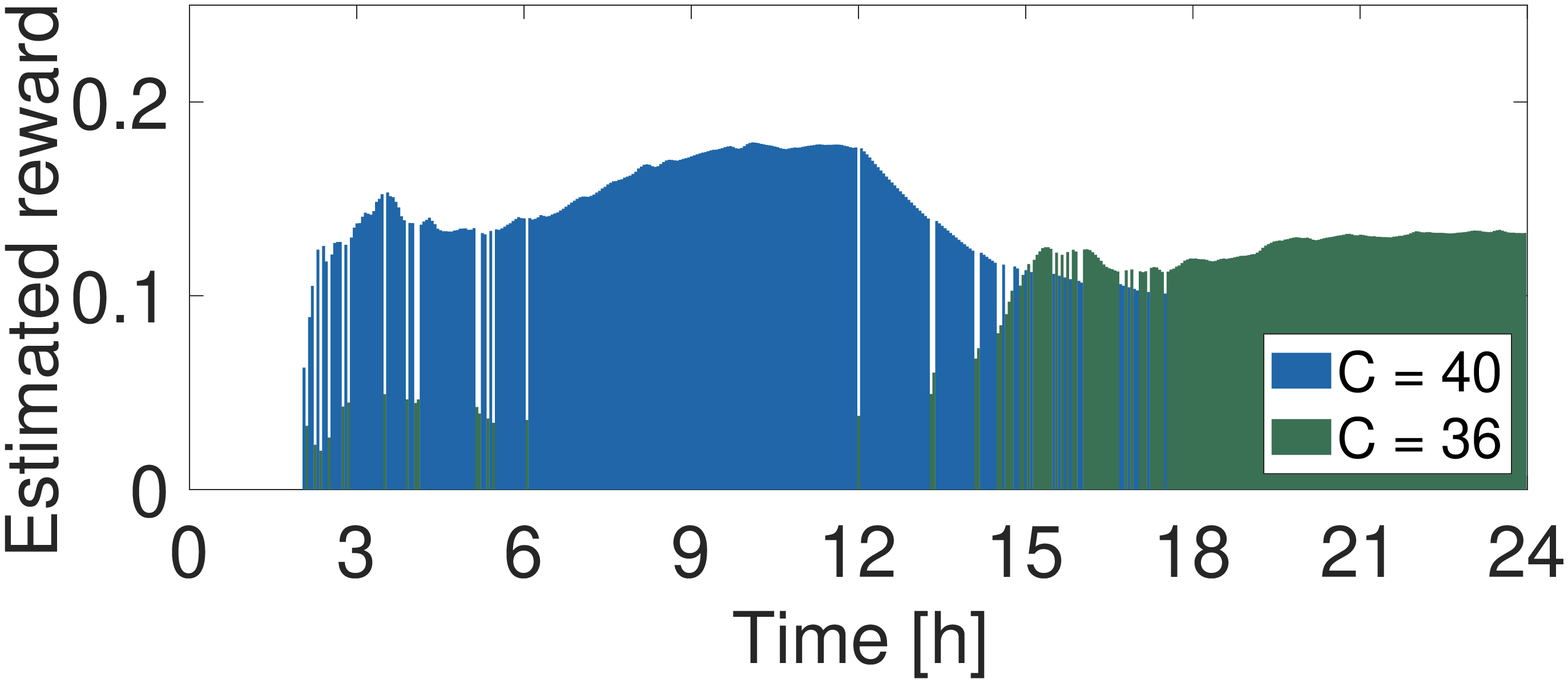}
         \caption{AP$_{3}$ estimated reward}
         \label{fig:EvoAP}
    \end{subfigure}
    \quad
    \begin{subfigure}[]{.4\textwidth}
         \centering
         \includegraphics[width=\textwidth]{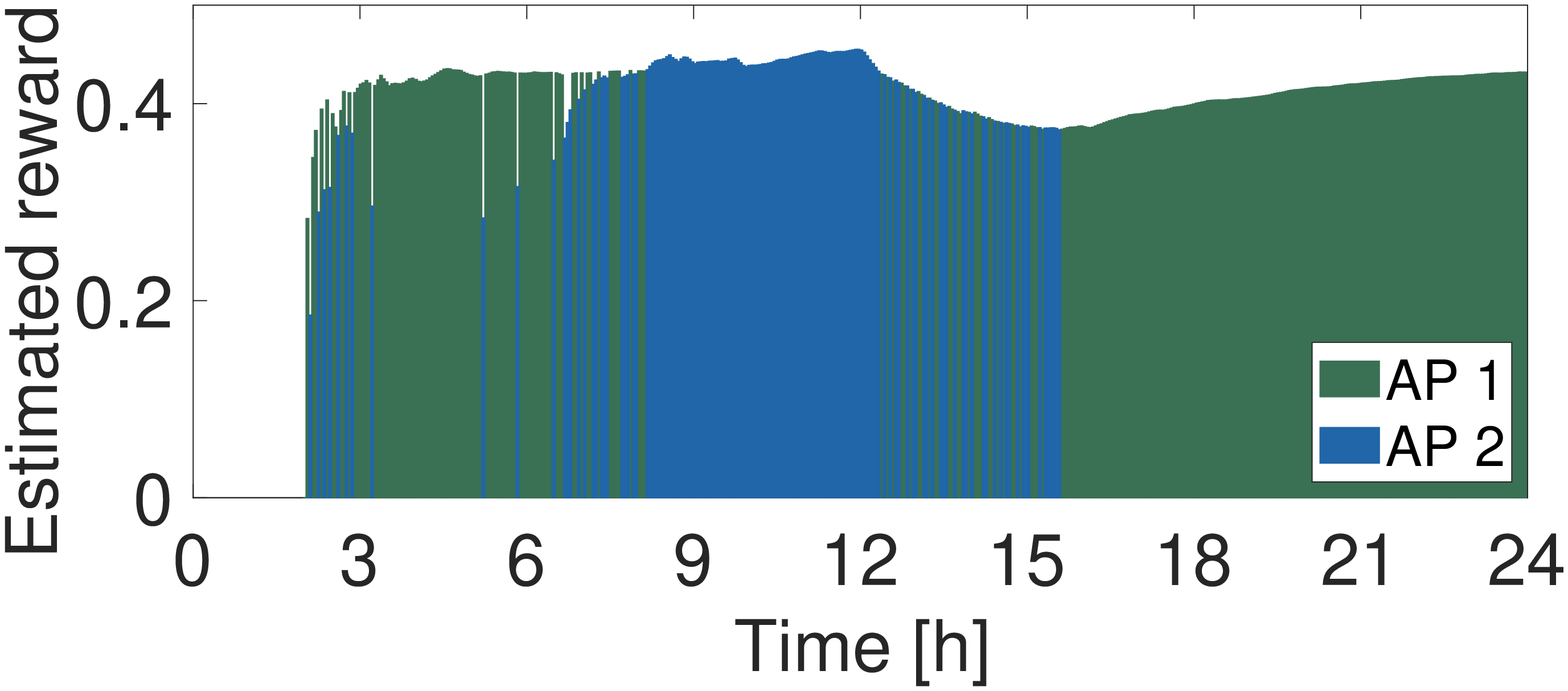}
         \caption{STA$_{2}$ estimated reward}
         \label{fig:EvoSTA}
     \end{subfigure}
     \caption{Satisfaction, rewards, and channel and AP selected by AP 3 and station 2, respectively, when assessing changing conditions.}
\end{figure*}

In order to assess changing conditions, we initially configure the three APs in the same radio channel, expecting that AP$_{1}$ and AP$_{3}$ will learn to reconfigure themselves as the simulation goes by. At $t = 12$~h, we observe that AP$_{1}$ and AP$_{3}$ have chosen the channel not used by AP$_2$. Then, we trigger a channel reconfiguration on AP$_{2}$, resulting in AP$_{1}$ and AP$_{3}$ to start exploring again the other channel as the reward of the currently selected channel decreases. Figure~\ref{fig:SatNon} shows the average satisfaction experienced by the stations. At $t = 2$~h the learning procedure starts. We can observe how rapidly the network converges to a feasible solution. However, at $t = 12$~h, we trigger the AP$_{2}$ reconfiguration from channel 40 to channel 36, remaining in this channel until the end of the simulation. As a result, both agent-enabled APs, AP$_1$ and AP$_3$, and stations start to perceive low rewards, and so they begin to explore the different available actions again. Therefore, channel 40, which was discarded at the early stages of the simulation, is  chosen as the preferable one now. We can see that in the stage marked as (1) the second learning phase happens, and the decision-making process evolves satisfactory. After a while, convergence is achieved and the satisfaction is stabilized over the performance threshold. 

For a more specific overview of the process, refer to Figures~\ref{fig:EvoAP} and~\ref{fig:EvoSTA}, where we have represented the estimated reward evolution experienced by AP$_{3}$ and STA$_{2}$, respectively. We have selected these two nodes as they can give us better insights on how the learning process is performed once the AP$_{2}$ changes its channel. Considering first Figure~\ref{fig:EvoAP}, we observe that AP$_{3}$ starts to explore the available channels, reaching a solution by selecting channel 36. However, at the half of simulation, due to the overlap caused by AP$_{2}$ being reconfigured to channel 36 too, AP$_{3}$'s reward starts to decay. Consequently, AP$_{3}$ explores again, and the action discarded at the beginning starts now to receive higher rewards, and so become more likely. Therefore, AP$_{3}$ switches to channel 40, moving to the new expected solution. 

On the other hand, Figure~\ref{fig:EvoSTA} depicts the behavior of STA$_{2}$. We observe that, when the learning stage finishes, the AP selected through the SSF criteria (AP$_{2}$) is discarded in order to be attached to AP$_{1}$. However, we can observe that even before $t = 12$~h, this station starts a learning phase again. This particular effect is caused due to other stations lasting more time to finish their learning process, which lastly may cause AP$_{1}$ to be saturated. As a result, STA$_{2}$ learns and attaches itself again to AP$_{2}$, since it provides better rewards due to the fact that at $t = 9$~h it is configured with a different channel than its neighbors AP$_{1}$ and AP$_{3}$. Finally, at $t = 12$~h, and since the AP$_{2}$ is forced to change the channel, STA$_{2}$ explores again, selecting AP$_{1}$ as final AP.

\subsubsection{Age of data and sliding window}\label{SLW_AOI}

Through this paper, we have considered a reinforcement learning approach in which agents' decisions are based on the previous selected actions and their performance. In wireless environments, such a procedure requires to carefully keep track of past records as network conditions may change at a fast pace. Differentiate between valid and obsolete data is necessary in order take actions according to the current state of the network. So, old observations can lead to perform bad decisions, but considering only recent data may result in losing useful information. Then, the age of data becomes a variable that must be included into the decision making process. 

\begin{figure}[t]
    \centering
    \includegraphics[width = .4\linewidth]{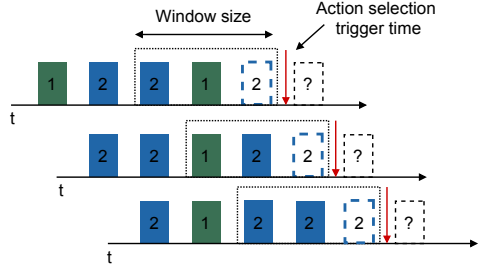}
    \caption{Sliding window time line representation.}
    \label{fig:SLWTime}
\end{figure}

To tackle this trade-off, we applied the concept of sliding window, which is intended to filter useful information from the outdated one. Particularly, the sliding window operation consists in a time interval that moves along with the simulation time, so performance records outside the window are not considered when evaluating the performance of an action. Using it, we allow agents to filter data, as well as to control and prevent bad decisions due to sporadic actions from other nodes. Figure~\ref{fig:SLWTime} shows the sliding window feature. Once the trigger time (i.e., T$_{DCA}$ or T$_{DAPS}$) is finished, agents will update, for the last action taken (depicted with a dashed line in the figure), the parameters $\hat{\mu}_{x}(t), \sigma_{x}^2(t)$, so the TS updates the distribution of the action. Only past records inside the window boundaries are used to update such parameters. A new action will be selected by drawing new values from  $\mathcal{N}(\hat{\mu}_{x}(t), \sigma_{x}^2(t))$ and selecting the arm returning the higher value.

\begin{figure*}[t]
\centering
    \begin{subfigure}[]{.45\textwidth}
         \centering
         \includegraphics[width =\textwidth]{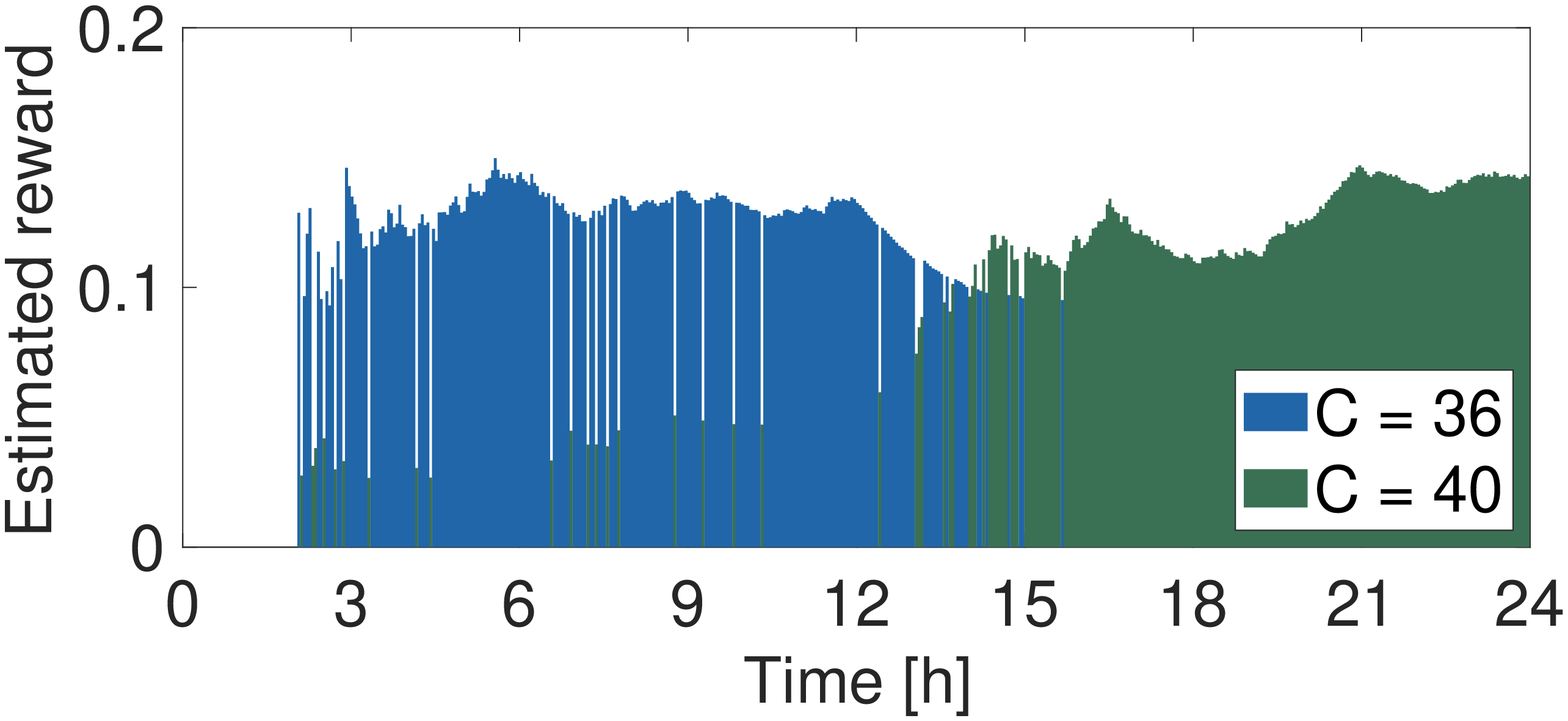}
         \caption{Sliding window interval 60~s}
         \label{fig:SLW60}
     \end{subfigure}
     \quad
     \begin{subfigure}[]{.45\textwidth}
         \centering
         \includegraphics[width =\textwidth]{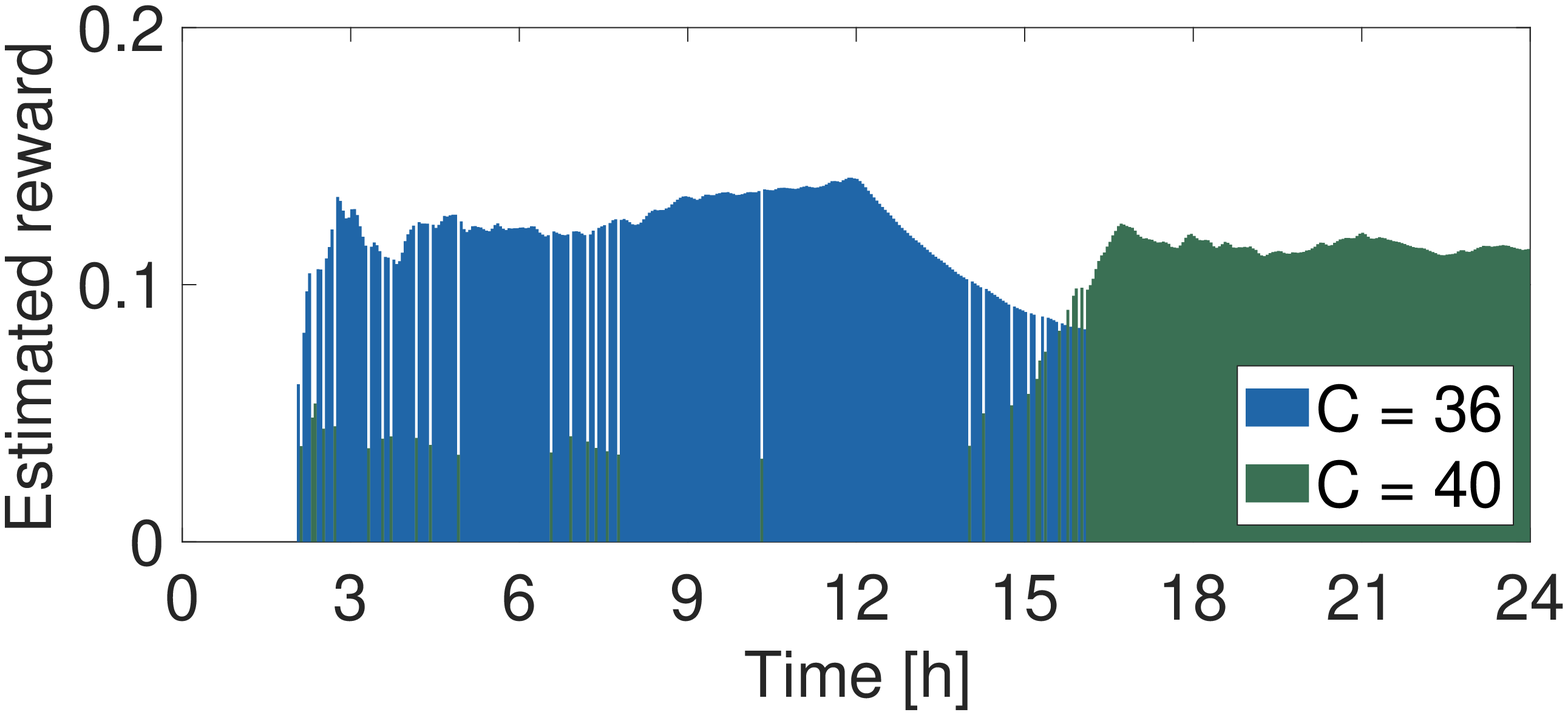}
         \caption{Sliding window interval 540~s}
         \label{fig:SLW540}
    \end{subfigure}
    \par\medskip
    \begin{subfigure}[]{.45\textwidth}
         \centering
         \includegraphics[width=\textwidth]{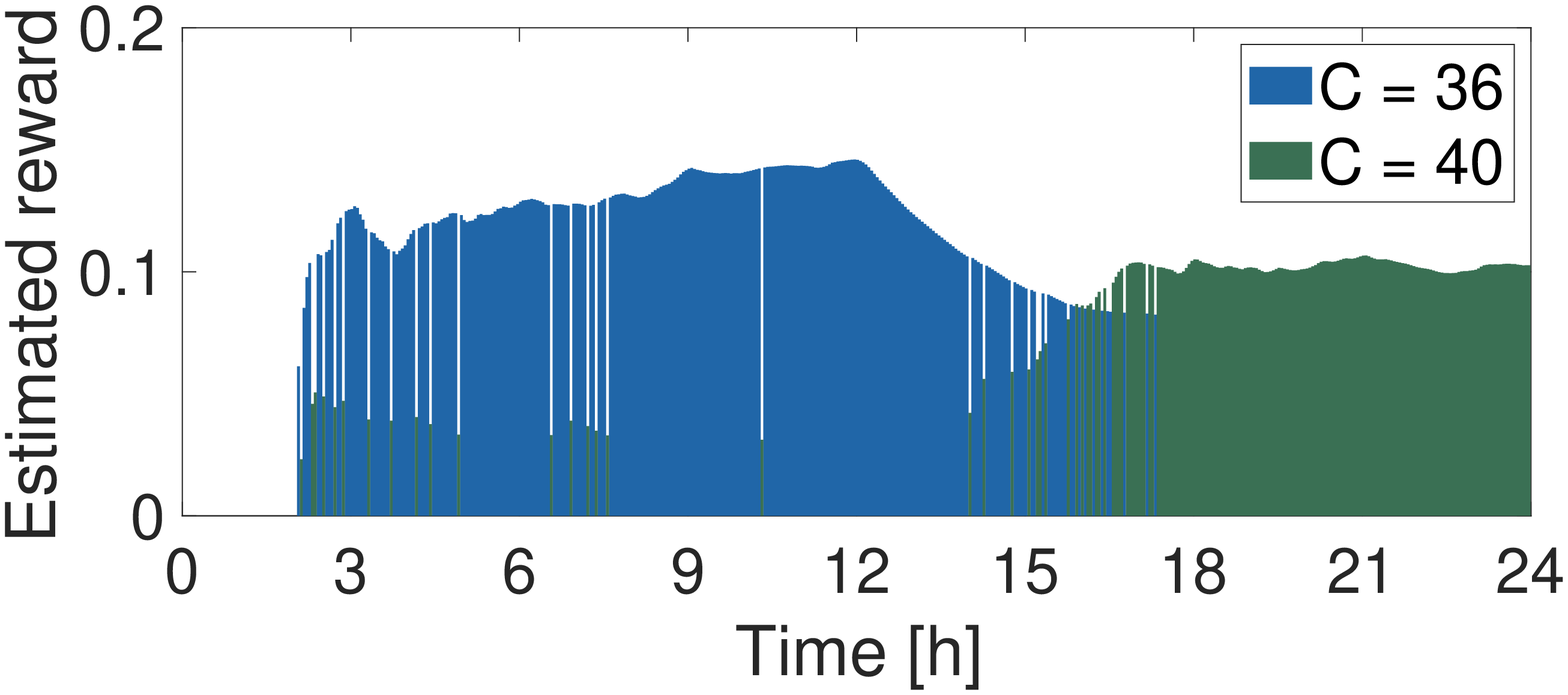}
         \caption{Sliding window interval 900~s}
         \label{fig:SLW900}
     \end{subfigure}
     \quad
     \begin{subfigure}[]{.45\textwidth}
         \centering
         \includegraphics[width=\textwidth]{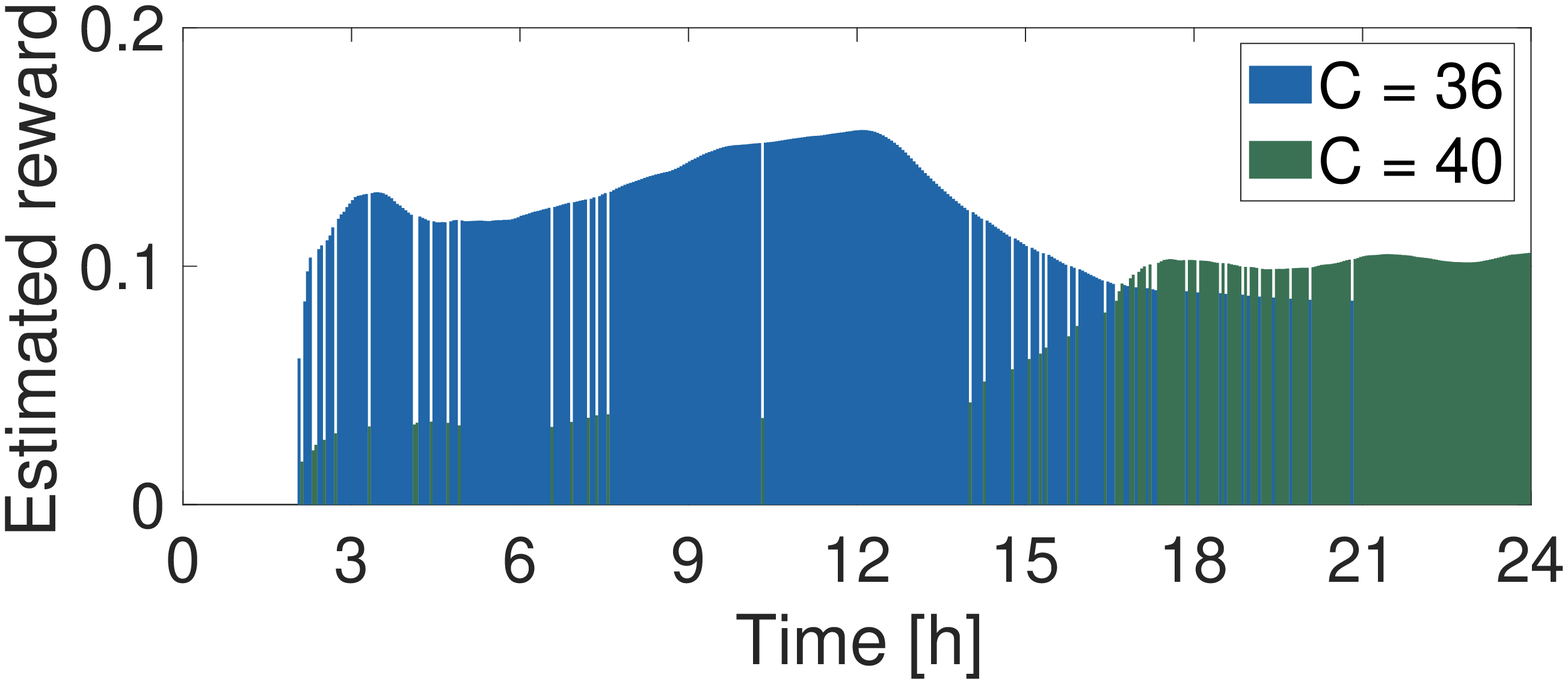}
         \caption{Sliding window interval 3600~s}
         \label{fig:SLW3600}
     \end{subfigure}
     \caption{Estimated reward evolution for AP$_{1}$ when considering different time windows.}
     \label{fig:SLW_EVO}
\end{figure*}

We have studied the implications of the sliding window size in the decision making process. We carried out this study under the changing environment conditions described in Section~\ref{subsubsec:nonStationary}, as it includes both stationary and non-stationary changes in the environment. Figure~\ref{fig:SLW_EVO} shows the estimated reward evolution of AP$_{1}$ for different sliding window sizes.

From the figures presented, we can see different behaviors as the sliding window interval increases. First, notice the sharper shape caused by having small size windows. This effect is related to the fact that the agents are reacting fast to changes from other players, since the reward evaluation is only averaged over a small set of reward entries. Then, it makes agents to become more vulnerable to others' decisions, and any random exploration by an agent may lead to an action change in all the others. Indeed, this issue can be corroborated on Figure~\ref{fig:SLW60} as frequent explorations are performed even after reaching a convergence state at $t = 6$~h. On the contrary, larger windows help to control and minimize the impact of the other agents on its own behavior, as it can be seen in Figure~\ref{fig:SLW3600}, in which smoother transitions reveal that agents are more robust against sporadic changes.

Although having a larger window size helps out agents to overcome the case of intermittent bad performances, it costs agent reaction time. We refer to the agents' ability to detect and avoid an action that has been repeatedly performing bad. Then, setting a conservative approach in order to provide robustness to agents may lead to unfeasible large reconfiguration times. This issue is shown in Figure~\ref{fig:SLW900} and Figure~\ref{fig:SLW3600}, in which agents require much more time to change to a better action. For instance, from $t = 12$~h, we find that for the 900~s and 3600~s cases, changing an action requires up to 6~h and 12~h, respectively. Therefore, we observe that there exists a clear trade-off between robustness and effective learning.

From all the simulations done, we observe that a window size of 540~s works well in the scenarios considered in this paper. As it can be seen in Figure~\ref{fig:SLW540}, exploration stages are barely performed after reaching convergence, and a quick reaction time (i.e., 3 h after $t = 12$~h) is registered.

%%%%%%%%%%%%%%%%%%%%%%%%%%%%%%%%%%%%%%%%%%%%%%%%%%%%%%%%%%%%%%%%%%%%%%%

%%%%%%%%%%%%%%%%%%%%%%%%% Section 7: Conclusions %%%%%%%%%%%%%%%%%%%%%%%%%
\section{Conclusions}\label{sec:conclus}

In this work, we have evaluated the implications of introducing learning algorithms for dynamic network adaptation. By means of the multi-armed bandits framework, we asses the concurrent decentralized channel and AP selection by enabling agent-empowered APs and stations, which learn by interacting with the environment. Through simulations, we provide insights on how MABs perform in high dense deployments, and their potential use in large enterprise WLANs. In this context, we have seen that DCA and DAPS improve the channel utilization and fairness. In addition, we have found that the DCA has a high but coarse impact on the network performance, whereas the DAPS mechanisms allow stations to eradicate unbalanced situations due to SSF association criteria, so fine tuning the network performance. Then, we can state that an effective AP selection scheme must consider and evaluate, not only link quality, but load metrics in order to ensure a good network performance. In addition, since channel interference is a determinant issue, a proper channel allocation mechanism must be executed along with the AP selection, as the impact in the performance highly depends on it. Additionally, obtained results have shown that the application of learning agents highly improves and minimizes the topology dependency that static strategies suffer, as the dynamic agents adapt the network configuration to the observed conditions.

We conclude that MAB-enabled agents work well under the presented conditions, showing the potential of ML mechanisms to significantly improve the network operation. Moreover, we have shown that an agent-enabled network using ML is capable of solving anomalies and adapting itself in base of the metrics presented. As future work, we will asses the potential implications of machine learning for non-static scenarios, as well as the use of learning algorithms in a central network controller, where a global view of the network state can be used to further enhance the system performance. In this regard, we are interested in the study of the trade-off between centralized and decentralized operations, as well as on the limitations of learning algorithms for optimization purposes.

%%%%%%%%%%%%%%%%%%%%%%%%%%%%%%%%%%%%%%%%%%%%%%%%%%%%%%%%%%%%%%%%%%%%%%%

\newpage
\appendix
\section{}\label{appendix:A}

We assume that all stations and APs operate using the IEEE 802.11ax amendment. The PHY and MAC parameters considered during the simulations are presented in Table~\ref{tab:80211AXparams}. We compute the duration of both data and control packet transmissions as detailed below. It is worth mention that we do not consider packet aggregation, and only one spatial stream per station is employed.

\begin{table} [h]
   \small % text size of table content
   \centering % center the table
   \caption{Simulation parameters}
   %\resizebox{\linewidth}{!}{%
       \begin{tabular}{ccc} % alignment of each column data
       \toprule
       \textbf{Parameter} & \textbf{Description} & \textbf{Value}\\ 
       \midrule
        $t_e$ & Empty slot duration & 9 $\mu$s\\
        $t_{\rm{SIFS}}$ & SIFS duration & 16 $\mu$s\\
        $t_{\rm{DIFS}}$ & DIFS duration & 34 $\mu$s\\
        $t_{\rm{PHY}}$ & Legacy preamble & 20 $\mu$s\\
        $t_{\rm{PHY-HE_{su}}}$ & HE single-station preamble & 164 $\mu$s\\
        $\sigma_{\rm{leg}}$ & Legacy OFDM symbol duration & 4 $\mu$s\\
        $\sigma$ & OFDM symbol duration & 16 $\mu$s\\
        $L_{\rm{SF}}$ & Service field length & 16 bits\\
        $L_{\rm{RTS}}$ & RTS packet length & 160 bits\\
        $L_{\rm{CTS}}$ & CTS packet length & 112 bits\\
        $L_{\rm{D}}$ & Data packet size & 12000 bits\\
        $L_{\rm{MH}}$ & MAC Header length & 320 bits\\
        $L_{\rm{ACK}}$ & ACK packet length & 112 bits\\
        $L_{\rm{TB}}$ & Tail bits length & 18 bits\\
       \bottomrule
       \end{tabular}
       %}
   \label{tab:80211AXparams}
\end{table}

\begin{equation*}
\centering
\begin{split}
    t_{\rm{RTS}} = t_{\rm{PHY}} + \left\lceil{\frac{L_{\rm{SF}} + L_{\rm{RTS}} + L_{\rm{TB}}}{L_{\rm{DBPS}}(\gamma_{i,j})}}\right\rceil\sigma_{\rm{leg}} 
\end{split}
\end{equation*}

\begin{equation*}
\centering
\begin{split}
    t_{\rm{CTS}} = t_{\rm{PHY}} + \left\lceil{\frac{L_{\rm{SF}} + L_{\rm{CTS}} + L_{\rm{TB}}}{L_{\rm{DBPS}}(\gamma_{i,j})}}\right\rceil\sigma_{\rm{leg}} 
\end{split}
\end{equation*}

\begin{equation*}
\centering
\begin{split}
    t_{\rm{DATA}} = t_{\rm{PHY-HE_{su}}} + \left\lceil{\frac{L_{\rm{SF}} + L_{\rm{MH}} + L_{\rm{d}} + L_{\rm{TB}}}{L_{\rm{DBPS}}(\gamma_{i,j})}}\right\rceil\sigma
\end{split}
\end{equation*}

\begin{equation*}
\centering
\begin{split}
    t_{\rm{ACK}} = t_{\rm{PHY}} + \left\lceil{\frac{L_{\rm{SF}} + L_{\rm{ACK}} + L_{\rm{TB}}}{L_{\rm{DBPS}}(\gamma_{i,j})}}\right\rceil\sigma_{\rm{leg}}
\end{split}
\end{equation*}

\vspace{0.2cm}
\noindent where $L_{\rm{DBPS}}(\gamma_{i,j})$ is the number of bits in each OFDM symbol, which in fact depends on the MCS accordingly selected to the RSSI value received ($\gamma_{i,j}$) for the link pair of station $i$ and AP $j$. Thus, $L_{\rm{DBPS}}(\gamma_{i,j})= N_{sc}N_{m}N_{c}N_{ss}$ where $N_{sc}$ is the number of data sub-carriers, $N_{m}$ is the number of bits per modulation symbol, $N_{c}$ is the coding rate and $N_{ss}$ is the number of spatial streams. In addition, note that control frames are transmitted in legacy mode using the lowest rate at MCS 0, and therefore being $L_{\rm{DBPS}} = 24$ bits.

Regarding the path loss, we have selected the IEEE 802.11ax enterprise model described in \cite{merlin2015tgax}, since we are considering a single floor environment. The path loss between a station $i$ and AP $j$ is given by
\begin{equation*}\label{eqn:pathloss}
\centering
\begin{split}
    \text{PL} (d_{i,j}) &= 40.05 + 20\log_{10}\left(\frac{f_c}{2.4}\right) + 20\log_{10}(\text{min}(d_{i,j},d_{bp}))\\
    &+(d_{i,j}>d_{bp})\cdot35\log_{10}\left(\frac{d_{i,j}}{d_{bp}}\right) + 7W_{i,j}
\end{split}
\end{equation*}
where $f_c$ is the AP's central frequency in GHz, $d_{i,j}$ is the distance between station $i$ and AP $j$ in meters, $d_{bp}$ is the breaking point distance in meters, and $W_{i,j}$ are the number of traversed walls. We set the breaking point distance, $d_{bp}$, to 5 m and the number of traversed walls, $W_{i,j}$, to 4. Note that the resulting propagation losses are given in dB.

\section*{Acknowledgments}

This  work  has  been  partially  supported  by the  Spanish Ministry of Economy and Competitiveness under the Maria de Maeztu  Units of Excellence  Programme  (MDM-2015-0502), by the Spanish Government under grant WINDMAL PGC2018-099959-B-I00 (MCIU/AEI/FEDER,UE), by the Catalan Government under grant 2017-SGR-1188, and  by  a  Gift from the Cisco University Research Program (CG\#890107, Towards Deterministic Channel Access in High-Density WLANs) Fund, a corporate advised fund of Silicon Valley Community Foundation. 

\bibliographystyle{unsrt}
\bibliography{references}

\end{document}